%% file: main.tex
\documentclass[10pt,twocolumn,letterpaper]{article}

\usepackage{cvpr}              % To produce the CAMERA-READY version
\input{preamble}

\definecolor{cvprblue}{rgb}{0.21,0.49,0.74}
\usepackage[pagebackref,breaklinks,colorlinks,allcolors=cvprblue]{hyperref}

\def\paperID{AI4Space-35} % *** Enter the Paper ID here
\def\confName{CVPR}
\def\confYear{2026}

\title{AstroSplat: Physics-Based Gaussian Splatting for Rendering \\ and Reconstruction of Small Celestial Bodies}

\author{Jennifer Nolan\\ {\tt\small jnolan9@gatech.edu}\and Travis Driver \\{\tt\small travisdriver@gatech.edu}\\ \\
Georgia Institute of Technology\\
Atlanta, GA\\
\and John Christian\\ {\tt\small john.a.christian@gatech.edu}
}

\begin{document}
\maketitle

%% teaser figure
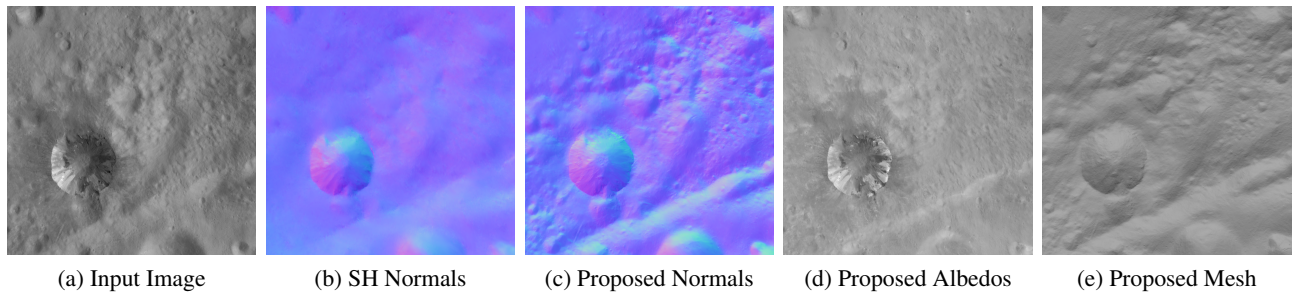
\begin{figure*}[t]
    \centering
    \input{fig/teaser.tex}

    \caption{\textbf{Our proposed AstroSplat framework compared to the traditional spherical harmonic (SH) parameterization.} The SH parametrization results in (b) smoothed normals maps, while the physics-based reflectance modeling of AstroSplat yields more detailed (c) surface normals, (d) albedos, and (e) meshes.}
    \label{fig:teaser}
\end{figure*}

\input{sec/0_abstract}    
\input{sec/1_intro}
\input{sec/2_related_work}
\input{sec/3_splatting}
\input{sec/4_photometry}
\input{sec/5_experiments}
\input{sec/6_results}
\input{sec/7_conclusion}

%%%%%%%%%%%%%%%%%%%%%%%%%%%%%%%%%%%%%%
%%%%%%%%%%%%%%%%%%%%%%%%%%%%%%%%%%%%%%
{
    \small
    \bibliographystyle{ieeenat_fullname}
    \bibliography{main}
}

%You must include your signed IEEE copyright release form when you submit your finished paper.
%We MUST have this form before your paper can be published in the proceedings.

%Please direct any questions to the production editor in charge of these proceedings at the IEEE Computer Society Press:
%\url{https://www.computer.org/about/contact}.

% WARNING: do not forget to delete the supplementary pages from your submission 
% \input{sec/X_suppl}

\end{document}

%% file: preamble.tex
\usepackage{adjustbox}
\usepackage{bm}

\usepackage{float}
\newcommand{\ra}[1]{\renewcommand{\arraystretch}{#1}} % command for increasing table row spacing

\newcommand{\vvec}[1]{\bm{\mathrm{#1}}}

\newcommand{\oB}{\mathrm{B}}
\newcommand{\oC}{\mathrm{C}}

\newcommand{\sang}{\iota}
\newcommand{\eang}{\varepsilon}

\usepackage{xcolor}
\usepackage{tikz}
\usetikzlibrary{shapes,arrows,calc,patterns,angles,quotes,fadings,decorations.pathreplacing}
\usepackage{tikz-3dplot}
\usepackage{tkz-graph}
\tikzstyle{vertex}=[circle, draw, inner sep=0pt, minimum size=8pt]

\tikzset{%
  >=latex, % option for nice arrows
  inner sep=0pt,%
  outer sep=1pt,%
  mark coordinate/.style={inner sep=0pt,outer sep=0pt,minimum size=3pt,
    fill=black,circle}%
}

%% file: fig/teaser.tex
\centering
\setlength{\tabcolsep}{2pt} % Default is 6pt
\begin{tabular}{cp{3.3cm}p{3.3cm}p{3.3cm}p{3.3cm}p{3.3cm}p{3.3cm}}
    % & \multicolumn{1}{c}{\small{Ground Truth}} & \multicolumn{1}{c}{\small{Render}} & \multicolumn{1}{c}{\small{Normals}} & \multicolumn{1}{c}{\small{Albedos}} \\
    & \multicolumn{1}{c}{} & \multicolumn{1}{c}{} & \multicolumn{1}{c}{} & \multicolumn{1}{c}{} & \multicolumn{1}{c}{} \\
    &
    \includegraphics[width=\linewidth]{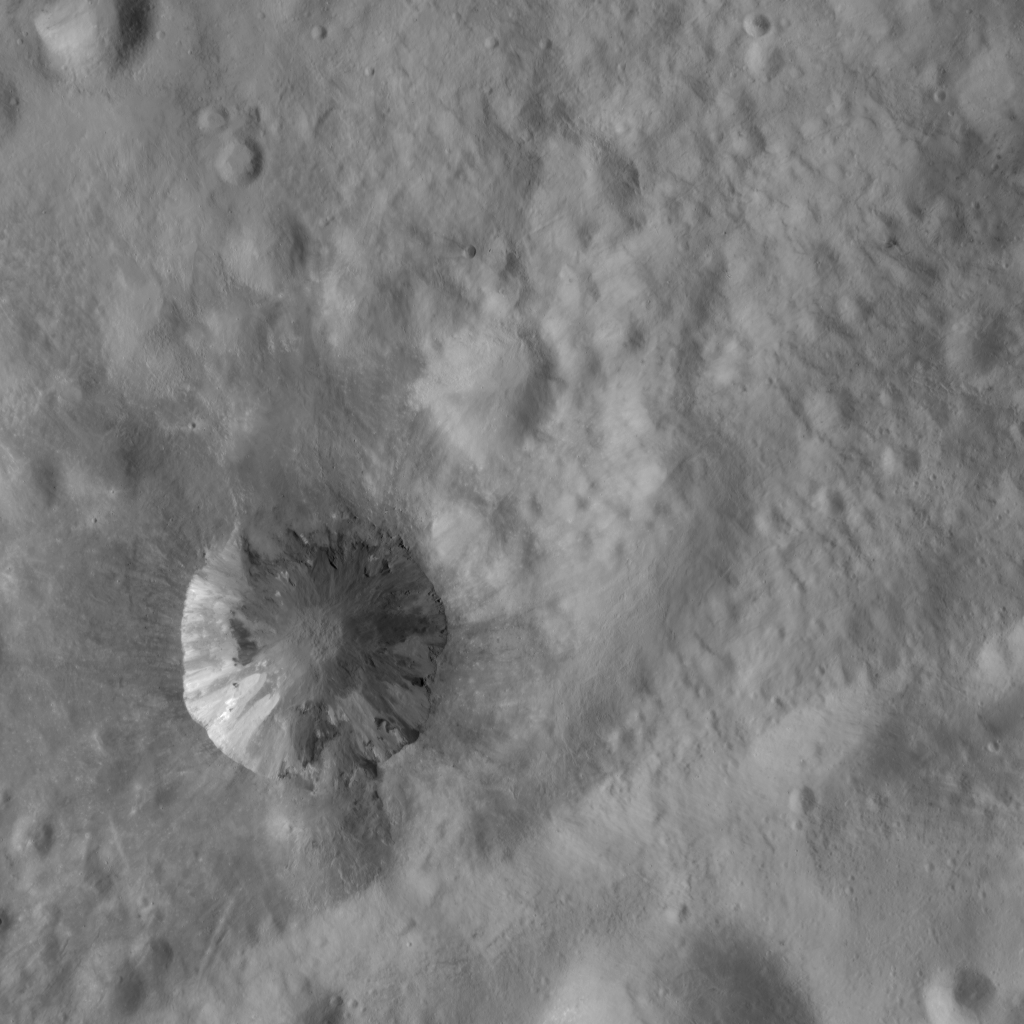} & %FC21B0008195_11275021648F1G.png
    \includegraphics[width=\linewidth]{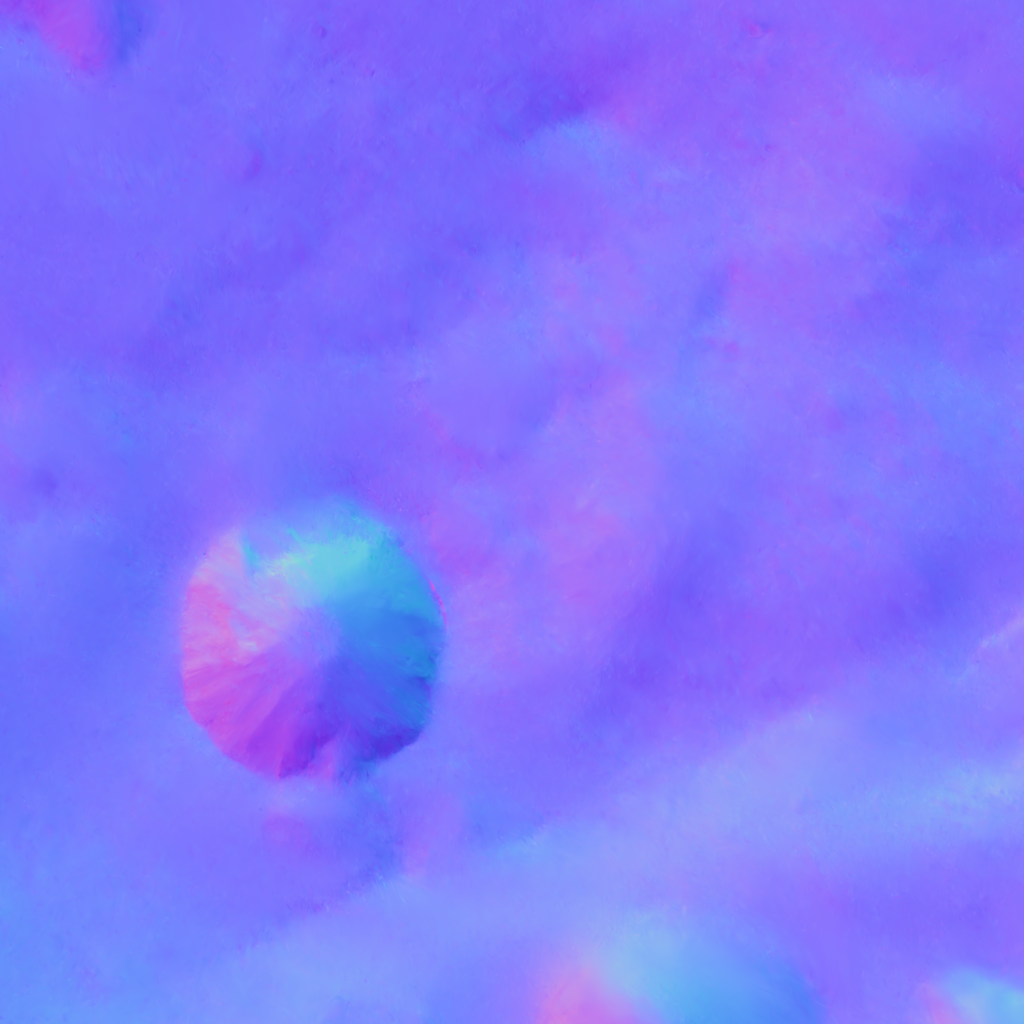} &
    \includegraphics[width=\linewidth]{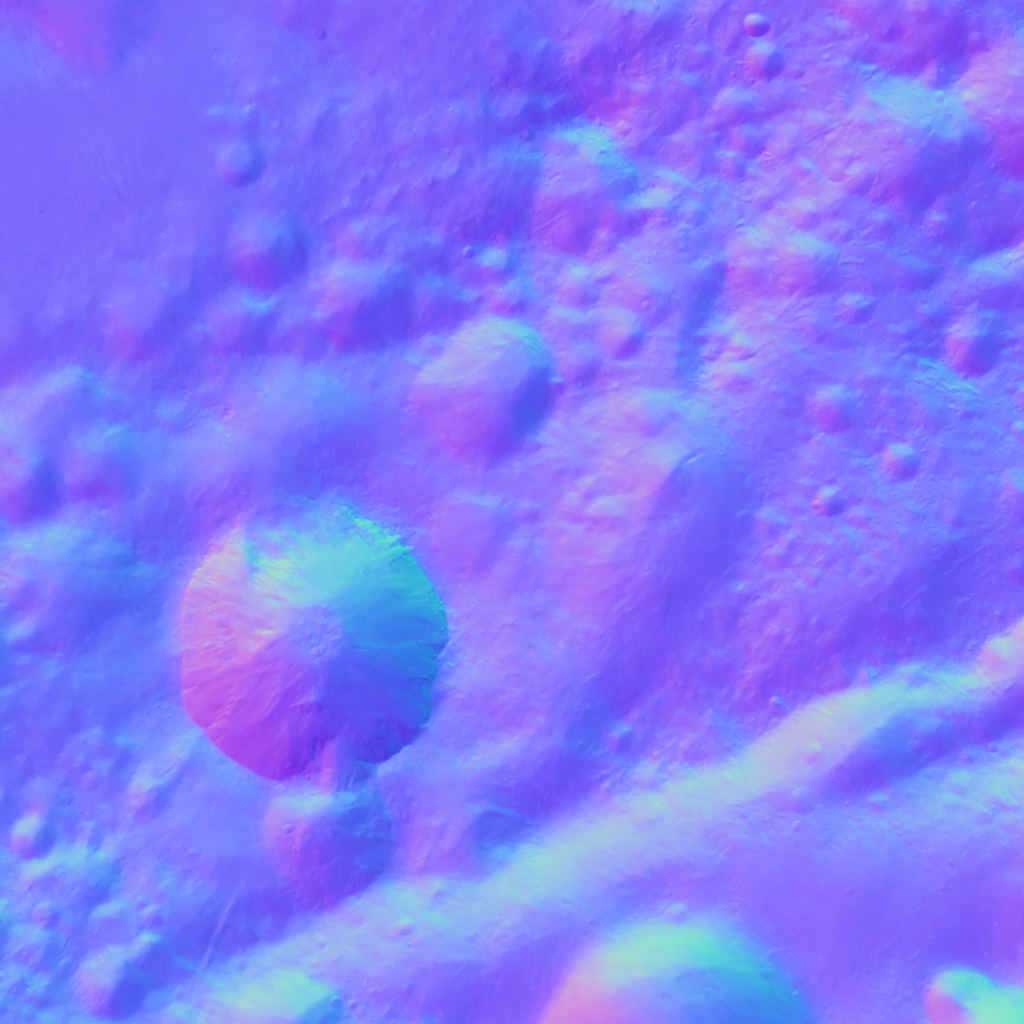} &
    \includegraphics[width=\linewidth]{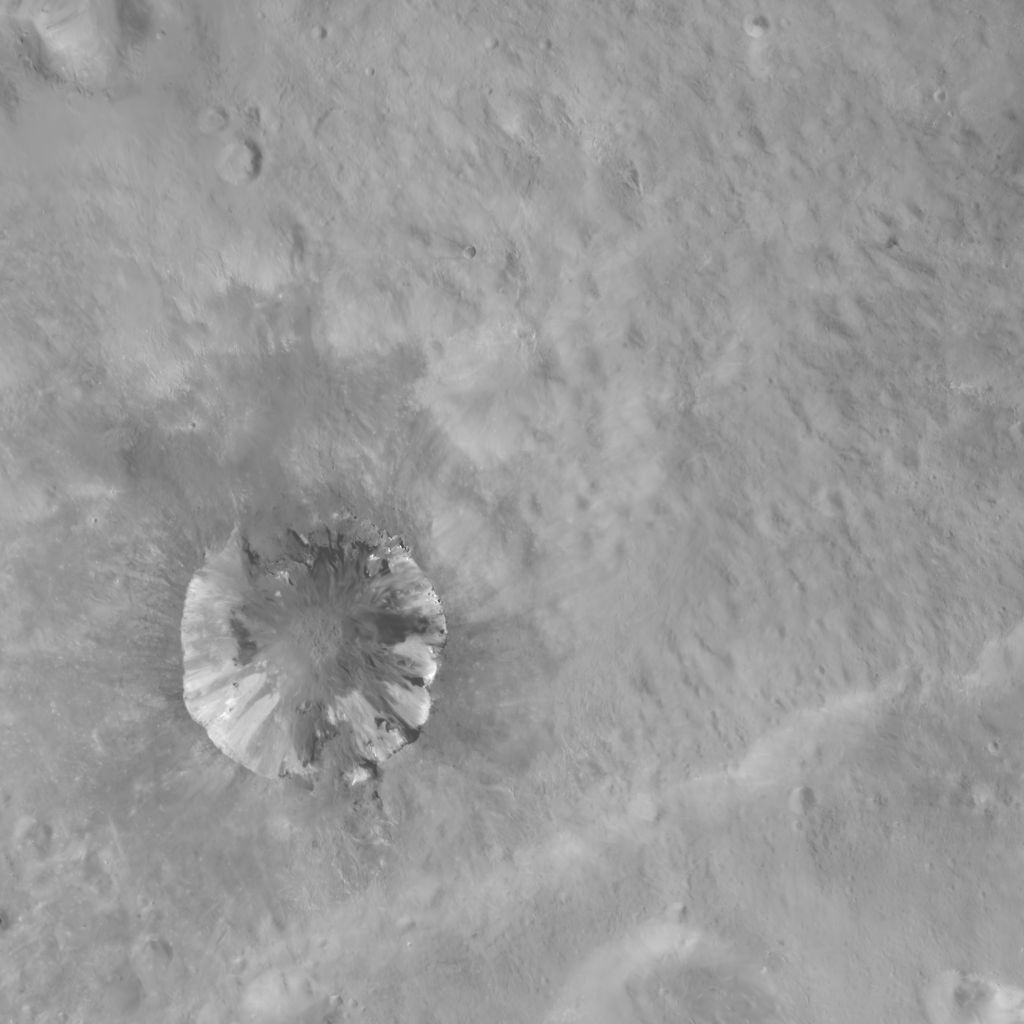} &
     \includegraphics[width=\linewidth]{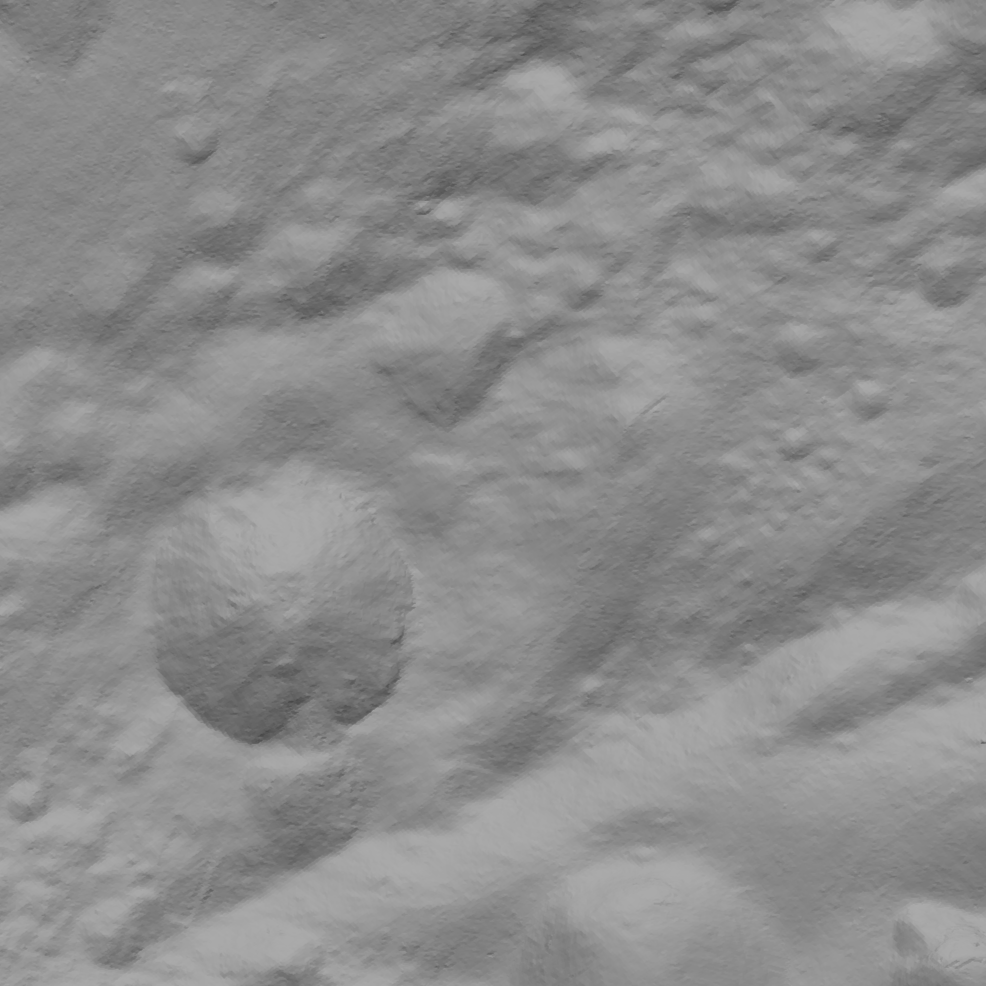}\\
    % & \multicolumn{1}{c}{\small{Ground Truth}} & \multicolumn{1}{c}{\small{Render}} & \multicolumn{1}{c}{\small{Normals}} & \multicolumn{1}{c}{\small{Albedos}} \\
    & \centering\small (a) Input Image
    & \centering\small (b) SH Normals
    & \centering\small (c) Proposed Normals
    & \centering\small (d) Proposed Albedos
    & \centering\small (e) Proposed Mesh \\
\end{tabular}

%% file: sec/0_abstract.tex
\begin{abstract}
% Why should the audience care?
Image-based surface reconstruction and characterization are crucial for missions to small celestial bodies (e.g., asteroids), as it informs mission planning, navigation, and scientific analysis. 
% Which problem do you address? 
Recent advances in Gaussian splatting enable high-fidelity neural scene representations but typically rely on a spherical harmonic intensity parameterization that is strictly appearance-based and does not explicitly model material properties or light-surface interactions.
% How do you address the problem?
We introduce AstroSplat, a physics-based Gaussian splatting framework that integrates planetary reflectance models to enable high-fidelity reconstruction and photometric characterization of small-body surfaces from \textit{in-situ} imagery. 
% What does your approach achieve?
The proposed framework is validated on \textit{real} imagery taken by NASA's Dawn mission, where we demonstrate superior rendering performance and surface reconstruction accuracy compared to the typical spherical harmonic parameterization.
\end{abstract}

%% file: sec/1_intro.tex
\section{Introduction} \label{sec:intro}

% Please follow the steps outlined below when submitting your manuscript to the IEEE Computer Society Press.
% This style guide now has several important modifications (for example, you are no longer warned against the use of sticky tape to attach your artwork to the paper), so all authors should read this new version.

Image-based surface reconstruction and photometric characterization are fundamental in the exploration of small airless bodies such as asteroids, comets, and moons. 
High-fidelity surface maps enable autonomous navigation, landing-site selection, hazard detection and avoidance, and scientific discovery. 
% Unlike terrestrial environments, small bodies exhibit extreme illumination conditions, highly non-Lambertian reflectance, irregular geometry, and sparse viewpoints.
Though critical to these operations, the extreme illumination variability and similar surface features encountered in the space environment make accurate geometric and appearance modeling challenging. 
% These characteristics make accurate geometry and appearance modeling particularly challenging in deep-space missions.

The traditional approach to reconstructing small body topography is stereophotoclinometry (SPC) \cite{gaskell2008,palmer2022practical}. 
The SPC process combines stereophotogrammetry (SPG) with photoclinometry to align captured images with an existing shape model and estimate\textit{ digital terrain maps} (DTMs), local topography and albedo maps. 
Specifically, SPC combines a reflectance model with accurate \textit{a priori} estimates of the camera pose, illumination direction, surface topography, and albedo to generate renders of a body's surface and employs correlation-based matching to register captured images to the DTM. 
This registration process is repeated for multiple images to generate dense correspondences, which is then followed by a photoclinometry step that refines surface topography and albedo estimates. 
Refinements are made to those \textit{a priori} values until convergence criteria is met between the ground truth and rendered images. 
%The surface topography and albedo information is then encoded into maplets, which are used to construct digital terrain maps (DTMs).
While SPC has been applied extensively to mapping various small bodies in the solar system, including Mars's moon Phobos \cite{ernst2023} and the asteroids 101955 Bennu \cite{barnouin2020}, 4 Vesta \cite{raymond2011dawn, mastrodemos2012}, and 25143 Itokawa \cite{gaskell2008}, it requires high-quality \textit{a priori} information and human operators to correct alignment errors. 
These limitations restrict scalability and constrain the level of autonomy achievable for future applications.
% which combines stereophotogrammetry (SPG) for global shape estimation with photoclinometry (shape-from-shading) for local refinement. 
% While SPC has enabled landmark planetary missions, it relies heavily on carefully curated tie points, manual intervention, and iterative operator supervision. 

Recent advances in neural scene representations, particularly Gaussian splatting \cite{kerbl2023gaussiansplatting}, have demonstrated impressive real-time rendering and reconstruction capabilities from multi-view imagery. 
Namely, the 2D Gaussian splatting (2DGS)~\cite{huang20242dgs} variant achieves high-fidelity view synthesis and topographic estimation by representing a scene with a set of Gaussian disks.
The reflectance properties of each Gaussian are typically represented by spherical harmonic (SH) coefficients. 
However, this parameterization is fundamentally appearance-based: it encodes view-dependent intensity without explicitly modeling the underlying physics of light–surface interactions. 
As a result, geometry and albedo effects are entangled, resulting in parameter estimates that may attribute illumination variance to albedo instead of topography.
To combat this effect, we propose \textit{AstroSplat}, a physics-based Gaussian splatting framework that applies photometric modeling in the intensity computation step.
Instead of the purely appearance-based SH parameterization, our method uses planetary reflectance models, which introduce a dependence on illumination direction and surface normal to determine per-Gaussian intensity. 

We validate AstroSplat on real imagery acquired by NASA’s Dawn mission \cite{russell2012dawn} to asteroid 4 Vesta and minor planet 1 Ceres. 
Experiments demonstrate that incorporating physics-based reflectance models yields superior rendering performance and improved topography accuracy compared to the standard SH parameterization (see Figure~\ref{fig:teaser}). 
% The result is an autonomous framework that eliminates the need for the human-intensive process of SPC while yielding superior reconstruction quality to traditional 2DGS. 
In summary, the main contributions of this paper are:
\begin{itemize}
    \item We integrate \textit{physics-based} planetary reflectance models into the 2DGS framework for topography estimation and rendering of small bodies.
    \item We analyze the effects of different reflectance models on reconstruction and rendering performance.
    \item We demonstrate that these reflectance models yield better performance compared to the typical spherical harmonic parameterization.
\end{itemize}

%% file: sec/2_related_work.tex
\section{Related Work} \label{sec:related-work}

To reduce reliance on the human-in-the-loop processes of SPC, pipelines that leverage deep learning have been developed to reconstruct small bodies from imagery. 
\citet{driver2025} proposed Photoclinometry-from-Motion (PhoMo), an autonomous framework which incorporates photoclinometry techniques into a keypoint-based structure-from-motion (SfM) system for normal and albedo estimation of small celestial bodies. 
PhoMo uses a deep learning feature detection and matching method~\cite{edstedt2024roma,driver2022astrovision} as well as the formalism of factor graphs~\cite{dellaert2017} to simultaneously estimate feature positions, camera poses, surface normals, and albedos.
This approach shows superior rendering and reconstruction quality over SPC while being fully autonomous and without requiring prior information. 

In the absence of explicit feature-matching algorithms, neural scene representations have gained popularity for use in space-based applications. 
Small body reconstructions have been achieved through the use of neural radiance fields (NeRFs) \cite {givens2024nerf,chen2024asteroidnerf} and their signed-distance function (SDF) \cite{hu2024neuralimplicit} extensions.
3D Gaussian Splatting (3DGS) has recently emerged as an efficient alternative to NeRFs~\cite{kerbl2023gaussiansplatting}. 
Although applications to asteroid imagery are nascent, early work such as \citet{prosvetov2025illuminating} applied 3DGS for detailed reconstruction of the lunar surface, and \citet{aira2025satellitegs} adapts Gaussian splatting to pushbroom satellite sensors for Earth--observation photogrammetry. 
These studies demonstrate the promise of neural implicit methods for reconstructing irregular small body geometries. 
However, existing methods remain predominantly \textit{appearance-based}, lacking explicit modeling of bidirectional reflectance and material properties. 
Our work builds on these trends by introducing a physics-aware Gaussian splatting framework that integrates planetary reflectance models, enabling autonomous, photometrically consistent reconstruction from \textit{in situ} imagery.

%% file: sec/3_splatting.tex
\section{Gaussian Splatting} \label{sec:splatting}

Conventional 3DGS \cite{kerbl2023gaussiansplatting} represents a scene with a set of ellipsoids containing learned parameters that are iteratively optimized to minimize the discrepancy between ground truth images and corresponding renders of the scene. 2DGS \cite{huang20242dgs} adjusts the 3DGS method by removing a dimension of the Gaussian representation, such that each ellipsoid flattens into a planar ellipse. In doing so, this representation allows for better geometric reconstructions by encouraging the Gaussians to align with surfaces in the scene. 
To accomplish this, the 2DGS method utilizes 5 types of learned parameters: each Gaussian's mean $\textbf{p}_k$, scaling $\textbf{s}_k$, orientation $\textbf{q}_{\mathcal{\mathcal{WS}}_{k}}$, opacity $\alpha_k$, and spherical harmonic coefficients that encode the view-dependent appearance $\textbf{c}_k$. The index subscript $k$ is hereon dropped for conciseness.

\begin{figure}
    \centering
    \includegraphics[width=0.9\linewidth]{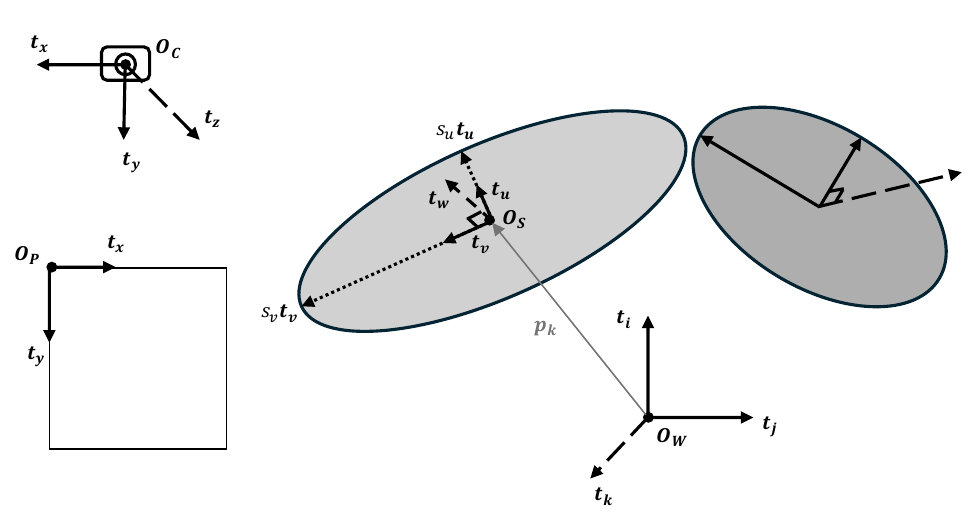}
    % \fbox{\rule{0pt}{3in} \rule{.9\linewidth}{0pt}}
    \caption{\textbf{2DGS frame definitions.} Relative orientation and positions of the local splat $\mathcal{S}$, world $\mathcal{W}$, camera $\mathcal{C}$, and pixel $\mathcal{P}$ frames. The origin of each frame is indicated by the point labeled $O$ and the basis directions are defined by $\textbf{t}$ vectors.}
    \label{fig:2dgs-frames}
\end{figure}

To begin, a set of $n$ Gaussians, derived from an initial sparse point cloud (e.g., from SfM) or simply randomized, are initialized in some world frame $\mathcal{W}$. 
Each Gaussian has a locally defined frame, referred to as the local splat frame $\mathcal{S}$. The origin is at the Gaussian's center, and the x- and y-axes of this local splat frame are defined by the principal directions of the ellipse, $\textbf{t}_u$ and $\textbf{t}_v$. 
The third direction completes the right-hand rule $\textbf{t}_w=\textbf{t}_u\times\textbf{t}_v$, and this direction is, by construction, normal to the Gaussian plane itself. 
The local splat frame is also scaled along its $\textbf{t}_u$ and $\textbf{t}_v$ directions by the scalars $s_u$ and $s_v$. 
The basis vectors of the local splat frame are expressed in the world frame such that they define the rotation from local splat to world coordinates by applying the rotation $\textbf{R}_{\mathcal{WS}}=\begin{bmatrix}\textbf{t}_u^\mathcal{W} & \textbf{t}_v^{\mathcal{W}} & \textbf{t}_w^\mathcal{W}\end{bmatrix}$. 
This work uses the passive representation of attitude \cite{christian2025,zanetti2019}. 
A quaternion encoding this rotation $\textbf{q}_{\mathcal{\mathcal{WS}}}$ represents the learned relative orientation of each Gaussian with respect to the world frame. The scaling $\textbf{s}$ is encoded in the matrix $\textbf{S}=\text{diag}([s_u,\ s_v,\ 0])$. 
Figure \ref{fig:2dgs-frames} provides a visual representation of these frames.

For all transformations, points in a given frame will be represented by homogeneous coordinates, in which a one is appended as weight to each 3D coordinate representation. Let $\textbf{x}^\mathcal{S}=\begin{bmatrix}x^\mathcal{S}&y^\mathcal{S} & 0 & 1\end{bmatrix}^\top$ be the homogeneous coordinate for a point on a Gaussian expressed in its local splat frame.
The third coordinate is zero due to the 2D nature of the Gaussian representation. In order to transform a point in this local splat frame to the 3D world frame, the following transformation is applied:
\begin{equation}
    \textbf{T}_{\mathcal{WS}}=\begin{bmatrix}
            \textbf{R}_{\mathcal{WS}}\textbf{S} & \textbf{p}^{\mathcal{W}}\\ 0 & 1
    \end{bmatrix}=
    \begin{bmatrix}
        s_u\textbf{t}_u^{\mathcal{W}} & s_v\textbf{t}_v^{\mathcal{W}} & \textbf{0}_{3\times1} & \textbf{p}^{\mathcal{W}}\\ 0 & 0 & 0 & 1
    \end{bmatrix}.
\end{equation}

To transform a point in $\mathcal{W}$ to the camera frame $\mathcal{C}$, whose $x$- and $y$-axes align with the image height and width directions and $z$-axis points boresight, the following transformation is applied:
\begin{equation}
    \textbf{T}_{\mathcal{CW}}=\begin{bmatrix}
            \textbf{R}_{\mathcal{CW}} & \textbf{r}_{WC}^{\mathcal{C}}\\ 0 & 1
    \end{bmatrix},
\end{equation}
where $\textbf{R}_{\mathcal{CW}}$ is the rotation matrix from $\mathcal{W}$ to $\mathcal{C}$, and $\textbf{r}_{WC}^{\mathcal{C}}$ is the position of the origin of $\mathcal{C}$ with respect to the origin of $\mathcal{W}$ expressed in $\mathcal{C}$. 
2DGS assumes these camera poses are known \textit{a priori}. 
After applying the transformation to the camera frame, homogeneous coordinates can be converted to pixel coordinates $\mathcal{P}$ by applying the conventional camera calibration matrix $\textbf{K}$ with an appended column of zeros to account for the projection from the $\mathbb{P}^3$ to the $\mathbb{P}^2$ projective spaces. Let $(d_x,d_y)$ denote the focal length in units of pixel length along both axes, and let $(u_p,v_p)$ denote the pixel coordinates of the image center, such that this transformation is defined by
\begin{equation}
\begin{split}
    \textbf{T}_{\mathcal{PC}}&=\begin{bmatrix}
            d_x & 0 & u_p & 0\\
            0 & d_y & v_p & 0\\
            0 & 0 & 1 & 0
    \end{bmatrix}=\mathbf{K}\begin{bmatrix}
        \mathbf{I}_{3\times3} & \mathbf{0}_{3\times1}
    \end{bmatrix}.
\end{split}
\end{equation}

The training process for 2DGS is broken into two stages: pre-processing and rendering. 
In the pre-processing stage, the transformation from each local splat frame to pixel frame is computed by applying the transformation defined above. 
This transformation is used to compute a bounding box around each Gaussian, which defines which pixels need to include contributions from this Gaussian in the rendering stage. 
Next, the view-dependent appearance of each Gaussian is determined using the reflectance model of choice. 
The original 2DGS implementation parameterizes the reflectance model using the coefficients of an $m$th degree SH expansion.

Once the parameters of each Gaussian are determined, the rendering stage begins, employing the same rasterization techniques used for 3DGS. 
This procedure requires ordering all Gaussians contributing to a given pixel by their depth. 
These depths are computed by finding the point where a ray emitted from the camera through the pixel coordinates, $\textbf{x}^{\mathcal{P}}=(x,y)$, intersects each Gaussian in its path, $\textbf{u}^\mathcal{S}=\textbf{u}^{\mathcal{S}}(\textbf{x}^{\mathcal{P}})=(u,v)$. Then, the Gaussian value at this intersection is computed according to $\mathcal{G}(\textbf{u}^{\mathcal{S}})=\exp\left(-(u^2+v^2)/2\right)$.
Sorting by the depths and starting from front to back, an iterative alpha-blending process is used to compute the intensity of each pixel $\bar{\textbf{c}}(\textbf{x}^{\mathcal{P}})$ by accumulating the view-dependent intensity $c_i$ of each Gaussian according to its opacity $\alpha_i$ and Gaussian spread $\hat{\mathcal{G}}_i$
\begin{equation}\label{alpha_blending}
\begin{split}
    \bar{\textbf{c}}(\textbf{x}^\mathcal{P})&=\sum_{i=1}c_i\omega_i,
\end{split}
\end{equation}

where $\omega_i=\alpha_i\hat{\mathcal{G}}_i(\textbf{u}^{\mathcal{S}})\prod_{j=1}^{i-1}(1-\alpha_j\hat{\mathcal{G}}(\textbf{u}^{\mathcal{S}}))$ and $\hat{\mathcal{G}}_i(\textbf{u}^{\mathcal{S}})=\max\left\{\mathcal{G}_{i}(\textbf{u}^{\mathcal{S}}),\mathcal{G}_{i}(\sqrt{2}(\textbf{u}^{\mathcal{P}}-\textbf{p}^{P})\right\}$.
Since small body imagery is monochrome, the computed intensity at each pixel is duplicated across the RGB channels to yield a grayscale image. This alpha-blending procedure is also used to accumulate a normal map and depth map by substituting the value $c_i$ with the third column of the $R_{\mathcal{WS}}$ matrix and the previously computed depth, respectively.

This rasterization step is conducted at each training iteration using a random camera pose from the set of training data, and the render, normal map, and depth map products are used to define the loss function that the optimization scheme is seeking to minimize. This loss function has two components in our implementation: an intensity loss $\mathcal{L}_{I}$ and a normal loss $\mathcal{L}_{n}$. The intensity loss is a weighted sum of the L1 difference and the Structural Similarity Index Measure (SSIM)\cite{zhou2004} between the RGB render and the ground truth image. The normal loss encourages the normal $\textbf{n}_i$ defined by the third axis of $\textbf{R}_{\mathcal{WS}}$ to align with the normal estimated from the gradient of the depth map $\textbf{n}_{d_{i}}$ for every $i$-th pixel in the normal and depth maps. The total loss $\mathcal{L}$ is defined as a weighted sum of these individual loss terms:
\begin{align}
     \mathcal{L} &= \mathcal{L}_{I}+\beta\mathcal{L}_{n},\ \beta\in\mathbb{R},\ \text{ where}\\
    \mathcal{L}_{I} &= (1-\lambda) \mathcal{L}_{\text{L1}} +\lambda\mathcal{L}_{\text{SSIM}},\ \lambda\in[0,1],\\
    \mathcal{L}_{n} &= \sum_i\omega_i(1-\textbf{n}_i^\top\textbf{n}_{d_{i}}).
\end{align}

%% file: sec/4_photometry.tex
\section{Small Body Photometry} \label{sec:photometry}

\subsection{Background}

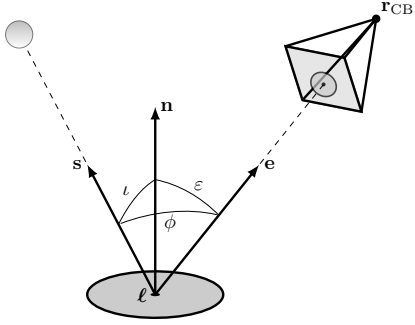
\begin{figure}
    \centering
    \resizebox{0.7\columnwidth}{!}{
        \input{fig/photometry-conventions}
    }
    % \par\vspace{15pt}
    % \vspace{-20pt}
    \caption{\textbf{Photometry conventions.} Photometric angles and their relationships to the Sun vector $\textbf{s}$, emission vector $\textbf{e}$, and normal $\textbf{n}$ with respect to a local patch centered at $\bm{\ell}$.}
    \label{fig:photometry-conventions}
\end{figure}

Photometry seeks to model the observed ``brightness'' of a scene point in an image as influenced by its surface topography and material properties.
While photometry is inherently complex, operation in space presents us with a number of advantages that simplify the photometric modeling process. 
First, we assume the Sun is a point source (the Sun subtends an angle of $0.5^\circ$ at Earth~\cite{shepard2017}) delivering approximately collimated light to the surface due to the large distance to the Sun and the lack of atmosphere of most small bodies to scatter the incoming light. 
Second, the direction of the incoming light can be precisely measured using typical onboard instrumentation (e.g., Sun sensors) or \textit{a priori} state knowledge. 
Third, previous photometric modeling of small bodies has demonstrated that \textit{global} reflectance functions, as opposed to \textit{spatially-varying} reflectance functions, can precisely estimate the observed brightness across the surface of a target small body.
Moreover, these global reflectance properties are similar across asteroids of the same taxonomic class~\cite{li2015}.

More formally, an image may be considered as a mapping $I \colon \Omega \rightarrow \mathbb{R}_+$ over the pixel domain $\Omega \subset \mathbb{R}^2$ that maps a point in the image, $\vvec{p} \in \Omega$, to its corresponding ``brightness" value $I(\vvec{p}) \in \mathbb{R}_+$. 
Here, ``brightness'' refers to the fact that the image values correspond to the amount of light falling on the photodetector inside the camera, referred to as the image \textit{irradiance} with units of power per unit area (W$\cdot$m$^{-2}$). 
Specifically, each brightness value $I(\vvec{p})$ is initially represented by a digital number (DN), which is computed by converting the charge accumulated by the photodetector over some exposure time $\Delta t$ to an integer DN using an analog-to-digital converter. 
For example, the Dawn mission to Asteroid 4 Vesta and Minor Planet 1 Ceres digitized the signal from the framing camera to a 14-bit integer DN~\cite{sierks2011dawn}. 
An important requirement for deep space imagers is response linearity, i.e., an almost perfectly linear relationship between the incident irradiance on the detector and the quantized charge rate $\mathrm{DN} / \Delta t$~\cite[Chapter 7]{christian2025}. 
This linear relationship is characterized by a rigorous \textit{radiometric calibration} process conducted both on the ground and during flight~\cite{sierks2011dawn,schroder2013cal,schroder2014cal}. 
It can be shown~\cite[Chapter~10.3]{horn1986} that the irradiance on the photodetector is proportional to the \textit{radiance}, with units of power per unit solid angle per unit area (W$\cdot$sr$^{-1}$$\cdot$m$^{-2}$), reflected towards the camera from the surface. 
Thus, each point $\vvec{p} \in \Omega$ corresponds to the radiance emitted from a point (or, more precisely, an infinitesimal patch) on the surface of the body. 

The emitted radiance from the surface, $L(\sang, \eang, \phi, a)$ and the incident (collimated) irradiance from the Sun, $F$, which is inversely related to the square of the distance to the Sun~\cite{christian2025,shepard2017}, are related by the \textit{bidirectional reflectance function}, $r$ (in units of $\text{sr}^{-1}$):
\begin{equation} \label{eq:brf}
    \frac{L(\sang, \eang, \phi, a)}{F} = r(\sang, \eang, \phi, a),
\end{equation}
where $a$ is the surface albedo, $\sang$ is the angle between the incoming light and the surface normal, or the \textit{incidence angle}, $\eang$ is the angle between the emitted light and the surface normal, or the \textit{emission angle}, and $\phi$ is the angle between the emitted light and the incoming light, or the \textit{phase angle} (see Fig. \ref{fig:photometry-conventions}). 
A similar measure of reflectance, which is very popular in the context of planetary photometry~\cite{shepard2017, li2015}, is the \textit{bidirectional radiance factor}, $r_\mathrm{F}$, which is the ratio of the bidirectional reflectance of a surface to that of a perfect Lambertian surface illuminated and viewed from overhead (i.e., $\sang = \eang = \phi = 0$):
\begin{equation} \label{eq:rad-factor}
    r_\mathrm{F}(\sang, \eang, \phi, a) = \pi r(\sang, \eang, \phi, a) = \frac{\pi L(\sang, \eang, \phi, a)}{F}. 
\end{equation}
While the radiance factor is dimensionless, it is common to refer to its value as being in units of reflectance or $L/F$ (or, more commonly, $I/F$ when $I$ is used to denote the radiance). 
Henceforth, we will refer to the radiance factor when mentioning the reflectance function. 

If we normalize the radiance factor by the radiance factor when observed and illuminated from overhead ($\iota = \varepsilon = \phi = 0^\circ$), we get the \textit{photometric function}, $\rho(\sang, \eang, \phi)$:
\begin{equation}
    \rho(\sang, \eang, \phi) = \frac{r_\mathrm{F}(\sang, \eang, \phi)}{r_\mathrm{F}(0, 0, 0)} \Leftrightarrow r_\mathrm{F}(\sang, \eang, \phi) = a_\mathrm{n} \rho(\sang, \eang, \phi), 
\end{equation}
where $a_\mathrm{n} := r_\mathrm{F}(0, 0, 0)$ is the \textit{normal albedo}. 
Finally, the photometric function may be factorized into the \textit{phase function}, $\Lambda(\phi)$, and the disk function, $d(\sang, \eang, \phi)$~\cite{shkuratov2011optical}:
\begin{equation}
    r_\mathrm{F}(\sang, \eang, \phi) = a_\mathrm{n} \Lambda(\phi) d(\sang, \eang, \phi). 
\end{equation}
The phase function, normalized to unity at $\phi = 0^\circ$, models the phase-dependent brightness variations that are independent of the incident and emission angles, referred to as the \textit{opposition effect}~\cite[Chapter 9]{hapke2012theory}, where $\Lambda(\phi)$ is often represented as an $n$th-order polynomial with parameters fit to imagery of the specific target body~\cite{schroder2013resolved,schroder2014cal}.
The disk function models brightness variations due to the underlying topography (which may also be a function of the phase angle). 
Henceforth, the term albedo will refer to the normal albedo and be denoted by $a$. 

We consider the case where the image values have not been radiometrically calibrated to units of reflectance. 
Recalling the response linearity of the photodetector, we may rewrite Equation \eqref{eq:rad-factor} as
\begin{equation}\label{eq:refl_uncalib}
    \frac{\pi L(\sang, \eang, \phi)}{F} \propto \frac{\mathrm{DN}(\sang, \eang, \phi)}{\Delta t}  - \xi = \Tilde{a}_\mathrm{n} \lambda d(\sang, \eang, \phi). 
\end{equation}
The scale term $\lambda$ accounts for the phase dependent brightness in the absence of an explicit phase function, the incident solar flux, and the lack of radiometric calibration to radiance. 
The bias term $\xi$ may be included to account for possible background noise~\cite{gaskell2008,park2019spc}. 
The \textit{relative} albedo $\Tilde{a}_\mathrm{n}$ is only proportional to the absolute albedo since compensation for the lack of a radiometric calibration may also be expressed through scaling of the albedos.
%We refer to this case as \textit{uncalibrated}. 

\subsection{Photometry Implementation in 2DGS}

The 2DGS SH reflectance model uses only the viewing direction to fit the coefficients of an SH expansion for intensity computation. 
Physics-based reflectance models can also use light direction (e.g., a sun sensor measurement) and surface normal in this computation step to estimate pixel intensity. 
Therefore, replacing the SH approach with a physics-based photometric function adds more geometric information to the estimation of image intensity, intrinsically tying the render to the underlying surface topography.

To exemplify the benefit of photometric modeling in this framework, we employ three reflectance models: Lambert, Lommel-Seeliger, and Lunar-Lambert. 
The \textit{Lambert} model assumes that a surface is perfectly diffuse, with light scattering equally in all directions. 
As such, its disk function is only dependent on the illumination direction and surface normal,
\begin{equation}\label{lam_refl}
    d_{\text{L}}(\iota)=\cos(\iota).
\end{equation}
The \textit{Lommel-Seeliger} (L-S) model considers that the light reflected by the surface experiences single-scattering along a specified viewing direction, yielding a dependence on illumination direction, viewing direction, and surface normal,
\begin{equation}\label{ls_refl}
    d_{\text{L-S}}(\iota,\varepsilon)=\frac{2\cos(\iota)}{\cos(\iota)+\cos(\varepsilon)}.
\end{equation}
The \textit{Lunar-Lambert} model\footnote{The Lunar-Lambert model was referred to as the McEwen model in the PhoMo study \cite{driver2025} used as the ground truth in this work.} is a weighted sum of the Lambert and Lommel-Seeliger models \cite{mcewen1986,mcewen1991}, which uses a phase weighting function $g(\phi)$, where $\phi$ is the phase angle in degrees, to adjust  their relative contributions,
\begin{equation}\label{ll_refl}
    d_{\text{L-L}}(\iota,\varepsilon,\phi)=(1-g(\phi))\cos(\iota)+g(\phi)\frac{2\cos(\iota)}{\cos(\iota)+\cos(\varepsilon)},
\end{equation}
where we use the exponential form of the phase weighting function $g$ proposed by Gaskell et al.~\cite{gaskell2008}, i.e., $g(\phi)=\exp\left(-\phi/60^\circ\right)$.
Derivations of these functions and a discussion of their properties can be found in \cite{christian2025}.

In order to apply these reflectance functions to the view-dependent intensity computation step, we add a per-Gaussian relative albedo $\tilde{a}_i$, which captures material property variation across the surface, and a per-image scale and bias correction factor (similar to the learned exposure compensation in hierarchical 3DGS \cite{hierarchicalgaussians24}), which accounts for the use of uncalibrated PNG image data. Gaussians with negative albedos or angle configurations that cause $\cos(\iota)+\cos(\varepsilon)$ to approach zero are not considered during rasterization, though no special consideration is given to shadowed and occluded regions.

%% file: fig/photometry-conventions.tex
\tdplotsetmaincoords{70}{120}
\begin{tikzpicture}[tdplot_main_coords, scale=1.2, every node/.style={scale=1.0}]

    \coordinate (oo) at (0,0,0);
    \coordinate (bl) at (-1/2,-1/2,0);
    \coordinate (br) at (1/2,-1/2,0);
    \coordinate (tl) at (-1/2,1/2,0);
    \coordinate (tr) at (1/2,1/2,0);
    %\filldraw[very thick, fill=gray!40] (tl) -- (tr) -- (br) -- (bl) -- cycle;
    \filldraw[very thick, fill=gray!40] (oo) circle[radius=1.0];
    \node[label=west:{$\bm{\ell}\,$}] (a) at (0,0,0) {};

    \coordinate (ss) at (3/0.75, 0, 4/0.75);
    \coordinate (ss2) at (3/0.8, 0, 4/0.8);
    
    \coordinate (cc) at (0, 3/0.80, 4/0.80);
    \node[label=north east:{$\vvec{r}_{\oC\oB}$}] () at (cc) {};
    
    \coordinate (cp) at (0, 3/1.05, 4/1.05);
    \coordinate (ctl) at (0.5, 2.5, 4.5);
    \coordinate (ctr) at (0.5, 3.5, 4.5);
    \coordinate (cbl) at (0, 2.5, 3.5);
    \coordinate (cbr) at (0, 3.5, 3.5);
    \draw[very thick,gray] (ctl) -- (ctr);
    \draw[very thick,gray] (ctl) -- (cbl);
    \draw[very thick,gray] (cbl) -- (cbr);
    \draw[very thick,gray] (ctr) -- (cbr);
    
    %% camera plane
    \filldraw[very thick, fill=gray!20] (ctl) -- (ctr) -- (cbr) -- (cbl) -- cycle;
    \draw[very thick] (cc) -- (ctl);
    \draw[very thick] (cc) -- (ctr);
    \draw[very thick] (cc) -- (cbr);
    \draw[very thick] (cc) -- (cbl);

    %\coordinate (mtl) at ($(cc)!0.27!(tl)$);
    %\coordinate (mtr) at ($(cc)!0.27!(tr)$);
    %\coordinate (mbl) at ($(cc)!0.27!(bl)$);
    %\coordinate (mbr) at ($(cc)!0.27!(br)$);
    %\filldraw[very thick, fill=gray!20] (mtl) -- (mtr) -- (mbr) -- (mbl) -- cycle;
    %\draw[thick, black] (cc) -- (mid);
    %\draw[thick, gray!50] (cc) -- (tr);
    %\draw[thick, gray!50] (cc) -- (br);
    %\draw[thick, gray!50] (cc) -- (bl);

    % camera and sun
    \shadedraw[opacity = .5] (3/0.75, 0, 4/0.75) circle (.2cm);
    \fill[black] (oo) circle (0.07);
    %\filldraw[draw=none, fill=gray!90] (cp) circle (.05cm);
    \filldraw[thick, fill=black] (cc) circle (.05cm);
    \node[inner sep=0pt] (camera) at (0, 3/1.1, 4/1.1) {};

    % draw an oriented circle in rotated coordinates
    \def\radius{.2}
    \def\thetavec{60}  % rotation about z
    \def\phivec{45}    % rotation from z toward xy-plane
    \tdplotsetrotatedcoords{\thetavec}{\phivec}{0}
    \tdplotdrawarc[tdplot_rotated_coords, thick, fill=gray!40, opacity=0.7]
      {(cp)}{\radius}{0}{360}{}{}
    \filldraw[draw=none, fill=black!80] (cp) circle (.03cm);
    
    \draw[dashed] (ss2) -- (oo);
    \draw[dashed] (cp) -- (oo);
    
    %\node[inner sep=0pt] (tmp) at (cp) {$\,\quad\vvec{p}$};
    
    % draw axes
    \draw[very thick,->] (oo) -- (0,0,5/1.7) coordinate (nhat) node [black,right] {$\,\mathbf{n}$};
    \draw[very thick,->] (oo) -- (3/1.5,0,4/1.5) coordinate (ihat) node [black,left] {$\mathbf{s}\,$};
    \draw[very thick,->] (oo) -- (0,3/1.7,4/1.7) coordinate (ehat) node [black,right] {$\,\mathbf{e}$};
    
    % arcs
    \tdplotsetrotatedcoords{0}{-90}{0}
    \tdplotdrawarc[tdplot_rotated_coords]{(0,0,0)}{1.8}{0}{90-53.13}{anchor=south}{$\quad\varepsilon$}
    \tdplotsetrotatedcoords{90}{-90}{0}
    \tdplotdrawarc[tdplot_rotated_coords]{(0,0,0)}{1.8}{0}{-36.87}{anchor=south}{$\iota\quad$}
    \tdplotsetrotatedcoords{45}{-62.0616}{0}
    \tdplotdrawarc[tdplot_rotated_coords]{(0,0,0)}{1.8}{24}{-24}{anchor=north}{$\phi$}
\end{tikzpicture}

%% file: sec/5_experiments.tex
\section{Experimental Setup} \label{sec:experiments}

To demonstrate the impact of implementing different photometric functions, we utilize three datasets, which were captured by NASA's Dawn mission during its observation of the Cornelia crater on 4 Vesta, as well as the Ikapati crater and Ahuna Mons on 1 Ceres~\cite{sierks2011dawn, russell2012dawn, pds, driver2025}. 
The train/test split is indicated in Table \ref{tab:rendering-metrics} and Table \ref{tab:normal-albedo-metrics}.
We trained for 30,000 iterations using the default parameters.
Each experiment was initialized with camera poses estimated by SPC, and an initial point cloud estimated by Georgia Tech’s Structure-from-Motion(GTSfM) library \cite{gtsfmcvpr} using these poses and keypoints derived by RoMa \cite{edstedt2024roma}.

To quantify the performance of each reflectance model, metrics for both the quality of the render and of the surface reconstruction were used. 
To evaluate render quality, we compare the peak signal-to-noise ratio (PSNR), the structural similarity index measure (SSIM) \cite{zhou2004}, and the learned perceptual image patch similarity (LPIPS) \cite{zhang2018}. 
To evaluate reconstruction quality, we compare our normal and albedo maps to values derived from PhoMo. 
Although a single point of comparison, the PhoMo outputs are well aligned with SPC, SPG, and dense SfM solutions, and they have shown superior rendering quality to other deep learning methods, like NeRFs \cite{driver2025}.
The normal error is defined as the total angular difference between the ground truth $\mathbf{n}_{ij}$ and rendered normals $\mathbf{\tilde{n}}_{ij}$ at each corresponding pixel in the normal maps: $\delta\theta_{ij}=\cos^{-1}\left(\mathbf{\tilde{n}}_{ij}^\top\mathbf{n}_{ij}\right)$. 
The albedo error is defined as the relative difference between the ground truth albedo $a_{ij}$ and rendered albedo $\tilde{a}_{ij}$, after being linearly fit to the ground truth, at each corresponding pixel in the albedo maps: $\delta a_{ij}=|a_{ij}-\tilde{a}_{ij}|/a_{ij}$. 
The average normal error $\delta\theta$ and average albedo error $\delta a$ across the overlapping regions of the PhoMo and AstroSplat results are used to quantify reconstruction quality.
Finally, the learned products from 2DGS can be exported as triangular meshes using a truncated signed-distance fusion formulation. These meshes are aligned to the PhoMo ground truth point cloud using an iterative closest point algorithm, and the Hausdorff distance \cite{huttenlocher1993} normalized by the size of the mesh , $d_H$, is calculated.

%% file: sec/6_results.tex
\section{Results} \label{sec:results}

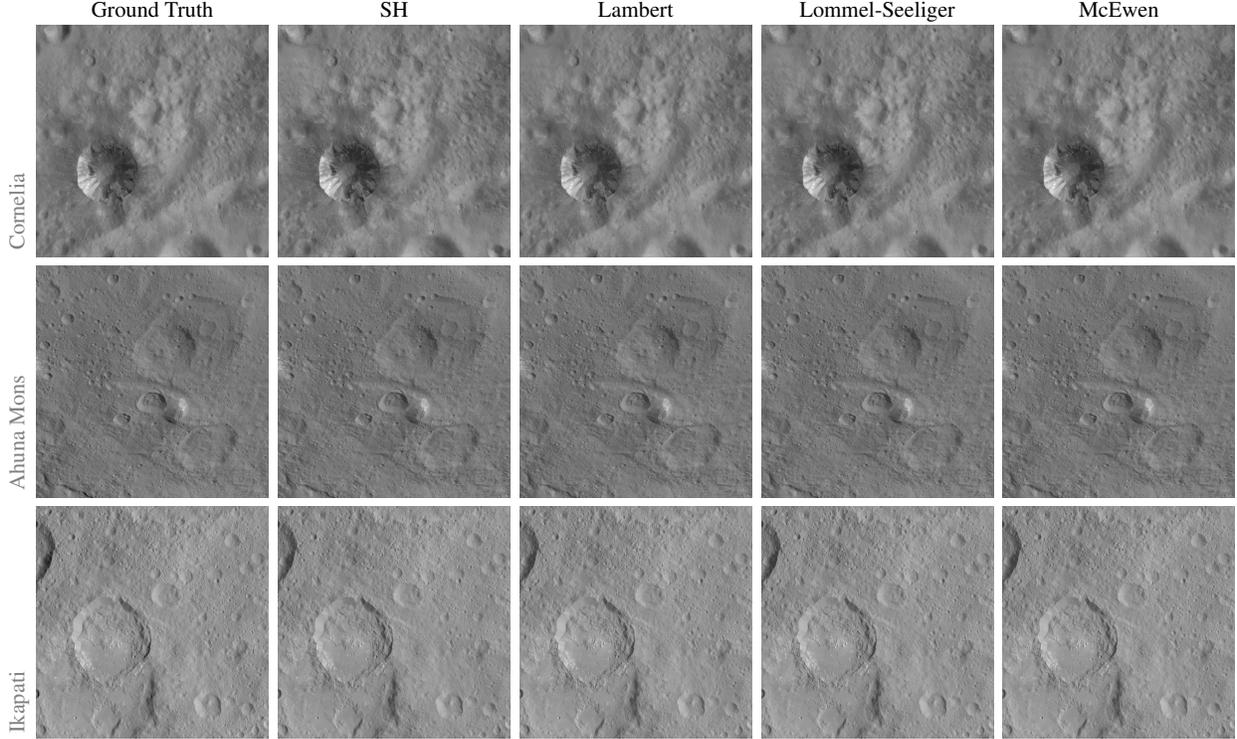
\begin{figure*}
    \centering
    \begin{adjustbox}{width=0.95\linewidth}
    \input{fig/full-qual-compare}
    \end{adjustbox}
    \caption{Qualitative comparison of test image renderings for each reflectance model.}
    \label{fig:rendering-qual-compare}
\end{figure*}

A set of ground truth test images as well as the associated test renders for each reflectance model are shown in Figure \ref{fig:rendering-qual-compare}. 
The overall PSNR, SSIM, and LPIPS values for both train and test image sets are provided in Table \ref{tab:rendering-metrics}. 
Across all datasets and all render metrics, the physics-based reflectance functions consistently outperformed the SH approach. 
For nearly all test images, the Lambert reflectance model yielded the best results, with Lunar-Lambert yielding the next best performance.

\begin{table*}[tb!]
    \centering
    \ra{1.2}
    \caption{\textbf{Rendering metrics for each reflectance model.} Each entry is of the form [train\_value]/[test\_value].  The \textbf{first} and \underline{second} best values are bold and underlined, respectively. The number of train and test images used are shown next to the body name.}
    \begin{adjustbox}{width=0.95\linewidth}
    \input{fig/rendering-metrics}
    \end{adjustbox}
    \label{tab:rendering-metrics}
\end{table*}

\begin{figure*}
    \centering
    \begin{adjustbox}{width=0.95\linewidth}
    \input{fig/normal-qual-compare}
    \end{adjustbox}
    \caption{Qualitative comparison of test image normals for each reflectance model.}
    \label{fig:normal-qual-compare}
\end{figure*}
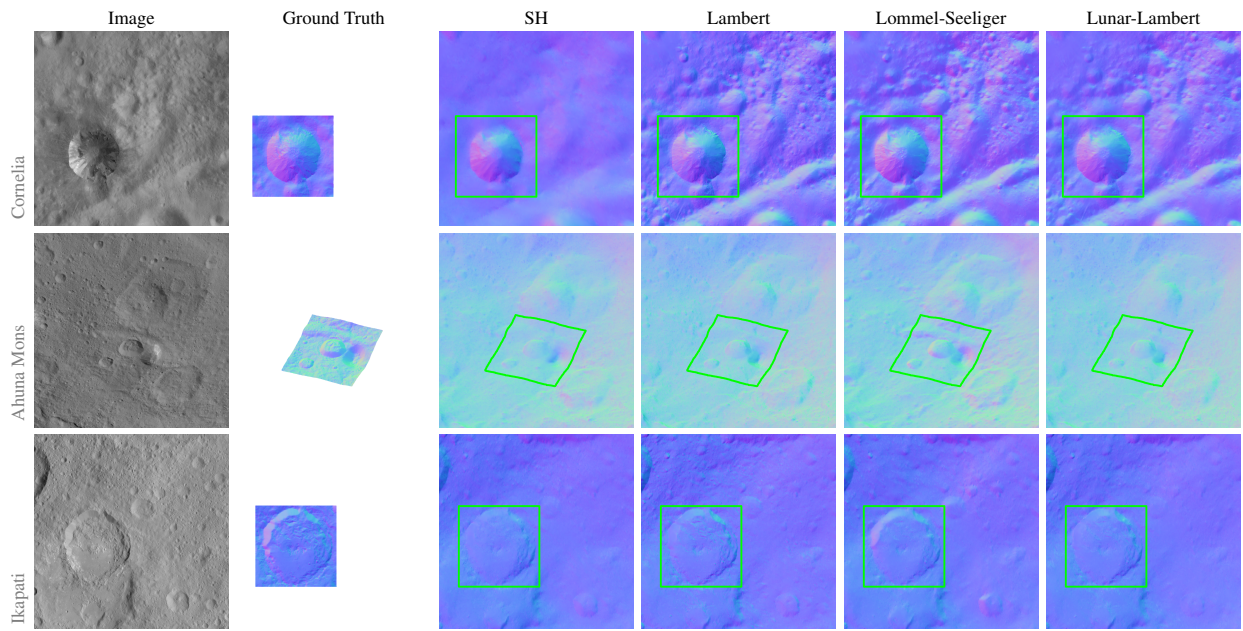

Beyond rendering quality, the normal maps and meshes created from the 2DGS products are used to determine the surface reconstruction quality. 
The ground truth images, PhoMo normal maps, and associated AstroSplat normal maps for a test set of each model are shown in Figure \ref{fig:normal-qual-compare}. 
The mean normal errors and symmetric Hausdorff distances over the ground truth region are shown in Table \ref{tab:normal-albedo-metrics} for all reflectance models. 
All physics-based reflectance models yielded lower mean normal errors compared to the original SH method, with the Lommel-Seeliger model minimizing the error for all datasets. Looking beyond the cropped region in each normal map, it is clear that several minor topographical features, such as smaller craters and ridges, appear in the physics-based approaches whereas those same areas are smoothed out by the SH approach. Comparing the ground truth PhoMo point cloud to the cropped mesh product for each approach, the symmetric Hausdorff distance was also consistently minimized by a physics-based approach. 

\begin{table*}[tb!]
    \centering
    \ra{1.2}
    \caption{\textbf{Reconstruction metrics for each reflectance model.} Each entry of the form [train\_value]/[test\_value]. The \textbf{first} and \underline{second} best values are bold and underlined, respectively. See Section \ref{sec:experiments} for metric definitions. The number of train and test images used are shown next to the body name.}
    \begin{adjustbox}{width=0.95\linewidth}
    \input{fig/normal-albedo-metrics}

    \end{adjustbox}
    \label{tab:normal-albedo-metrics}
\end{table*}

\begin{figure*}
    \centering
    \begin{adjustbox}{width=0.95\linewidth}
    \input{fig/albedo-qual-compare}
    \end{adjustbox}
    \caption{Qualitative comparison of test image albedos for each reflectance model (except spherical harmonics).}
    \label{fig:albedo-qual-compare}
\end{figure*}
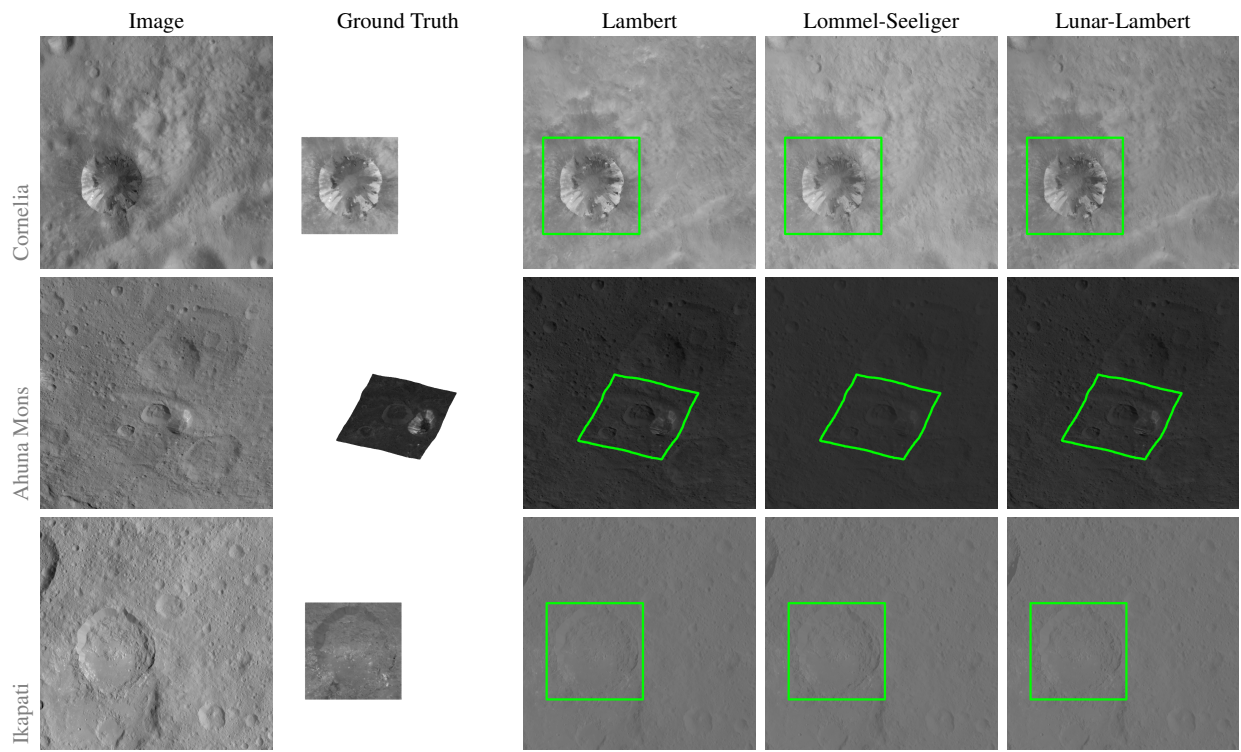

The use of physics-based reflectance models also adds the benefit estimating albedo, which can be used to assess the quality of the surface reconstruction. 
The ground truth images, PhoMo albedo maps, and associated AstroSplat albedo maps for a test set of each model are shown in Figure \ref{fig:albedo-qual-compare}. 
The mean relative albedo error for each physics-based approach for all datasets is shown in Table \ref{tab:normal-albedo-metrics}. 
The Lambert model provides the closest albedo to the ground truth for Cornelia, yielding about a 20\% and 30\% reduction in error compared to Lunar-Lambert and Lommel-Seeliger, respectively. The difference in relative errors between the models for Ikapati and Ahuna Mons are on the order of thousandths.
The underestimation of higher albedo values on the Ceres features yields less variation in albedo value compared to the PhoMo results, resulting in a consistent error across models, especially compared to Cornelia. 

With superior rendering quality, more accurate surface normal estimation, and the capability to estimate albedo, physics-based reflectance models outperform the traditional 2DGS approach on all metrics. Furthermore, these models require fewer parameters than the SH method, which learns 16 parameters for the standard third-degree expansion. The Lambert, Lommel-Seeliger, and Lunar-Lambert models require only one intensity-specific parameter, namely albedo, and leverage the available illumination information and the normal estimate from the Gaussian orientation to accomplish the same task with better results.
 
This experiment relies on SPC poses for initialization, but camera poses estimated from an autonomous pipeline would yield an end-to-end autonomous reconstruction. 
To this end, the supplementary materials contain an initialization sensitivity analysis using an autonomous approach.

%% file: fig/full-qual-compare.tex
\centering
\setlength{\tabcolsep}{2pt} % Default is 6pt
\begin{tabular}{cp{3.3cm}p{3.3cm}p{3.3cm}p{3.3cm}p{3.3cm}}
    & \multicolumn{1}{c}{\small{Ground Truth}} & \multicolumn{1}{c}{\small{SH}} & \multicolumn{1}{c}{\small{Lambert}} & \multicolumn{1}{c}{\small{Lommel-Seeliger}}  & \multicolumn{1}{c}{\small{Lunar-Lambert}} \\
    \parbox[t]{2.5mm}{\rotatebox[origin=l]{90}{\textcolor{gray}{\small{Cornelia}}}} & 
    %FC21B0008195_11275021648F1G
    \includegraphics[width=\linewidth]{fig/render/cornelia/GT/00000.png} &
    \includegraphics[width=\linewidth]{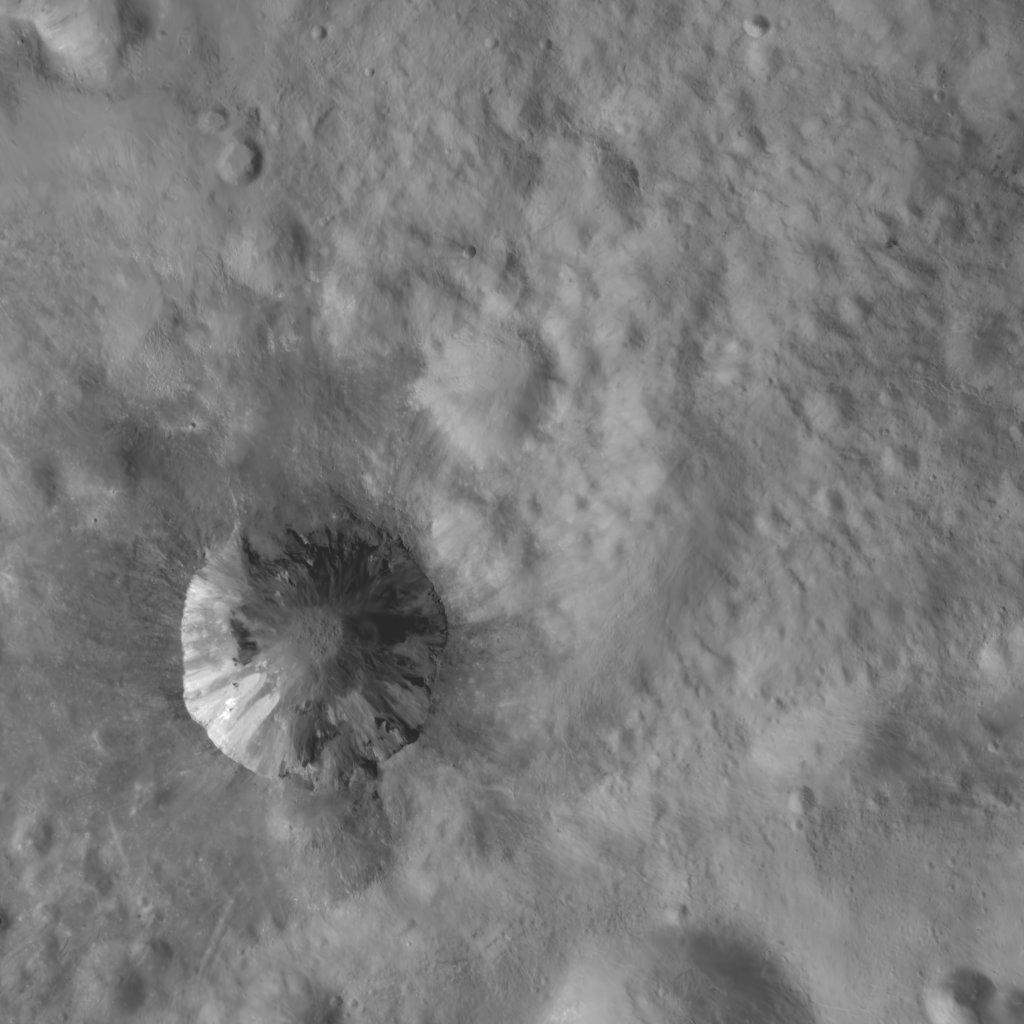} &
    \includegraphics[width=\linewidth]{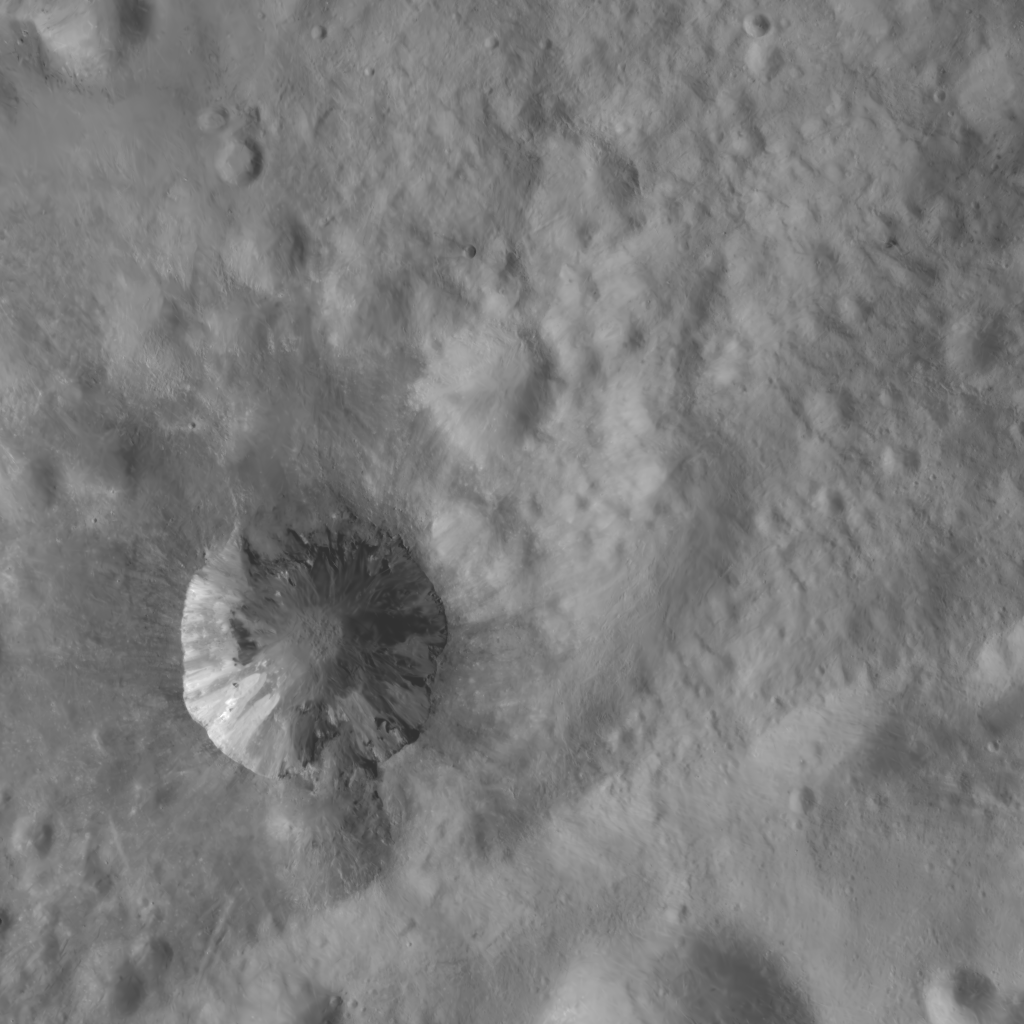} &
    \includegraphics[width=\linewidth]{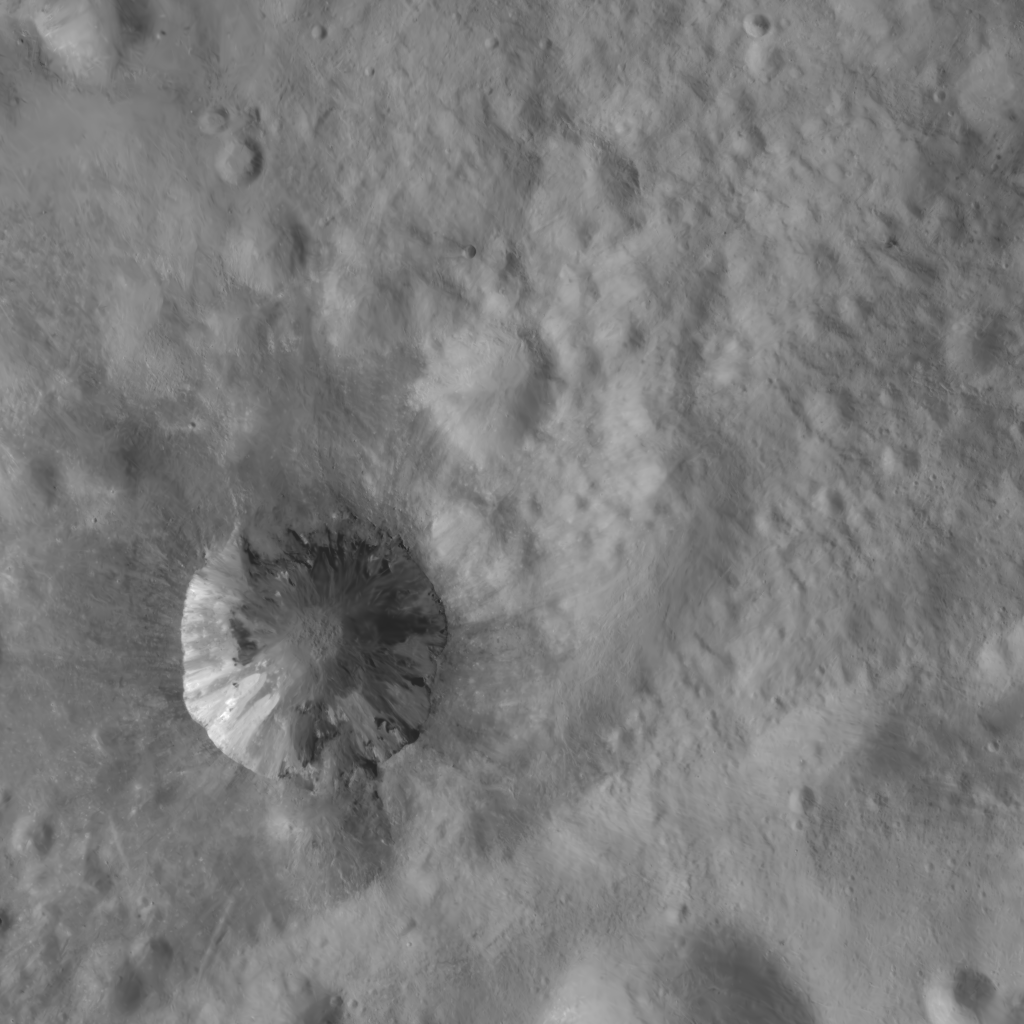} &
    \includegraphics[width=\linewidth]{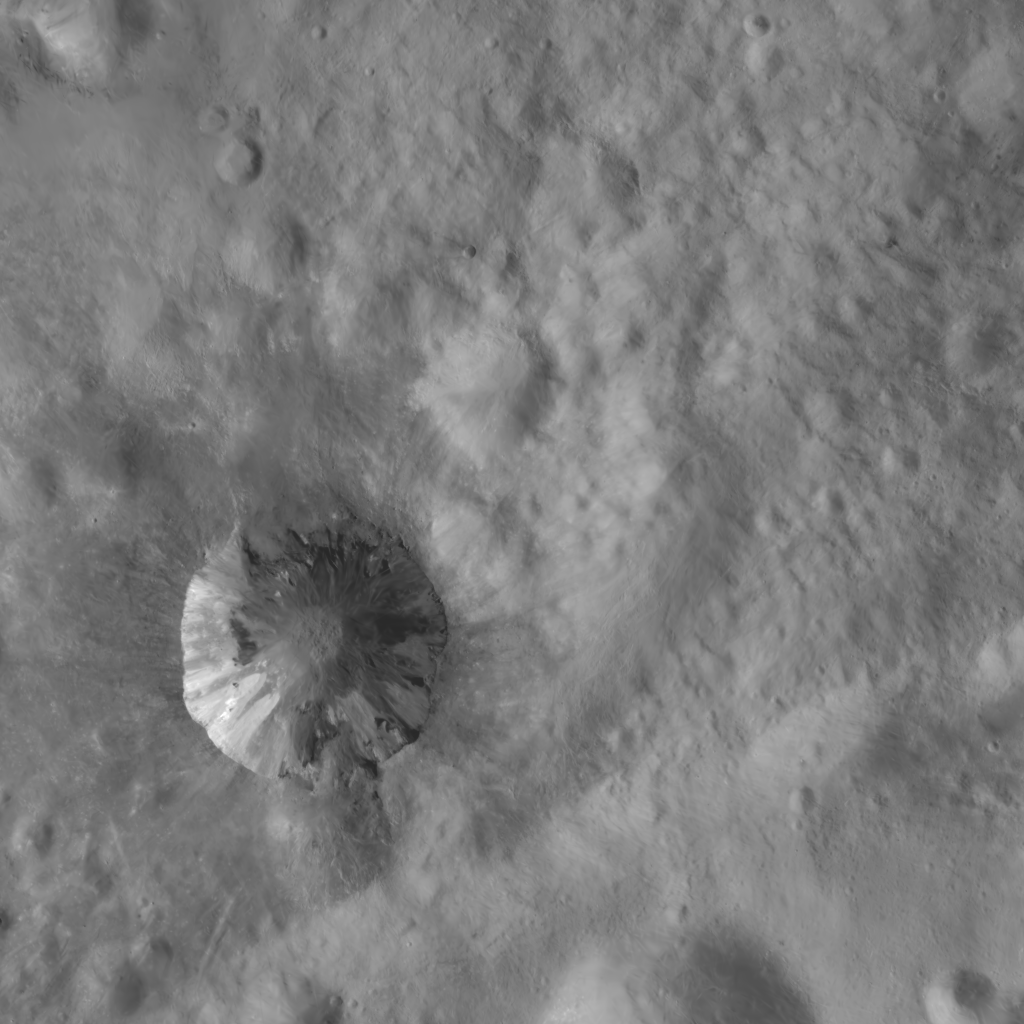} \\
    \parbox[t]{2.5mm}{\rotatebox[origin=l]{90}{\textcolor{gray}{\small{Ahuna Mons}}}} &
    \includegraphics[width=\linewidth]{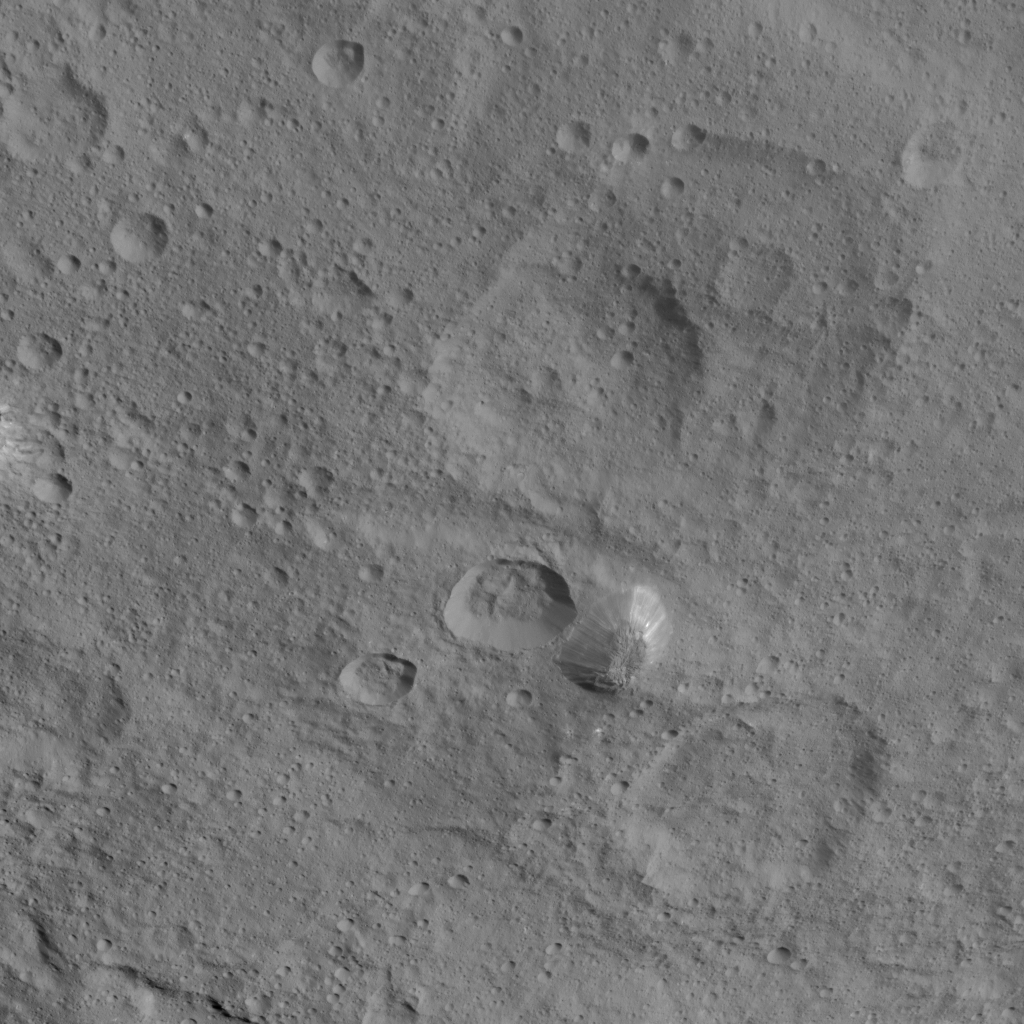} &
    \includegraphics[width=\linewidth]{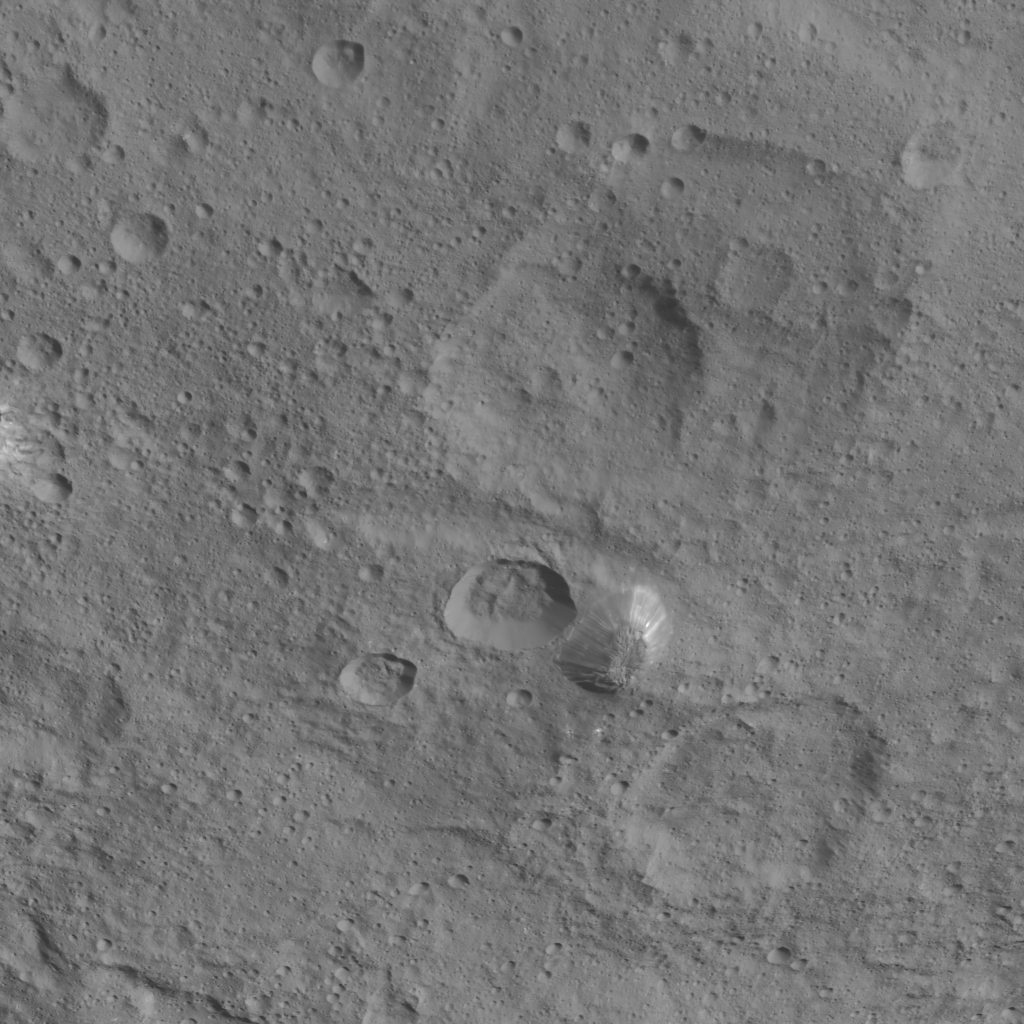} &
    \includegraphics[width=\linewidth]{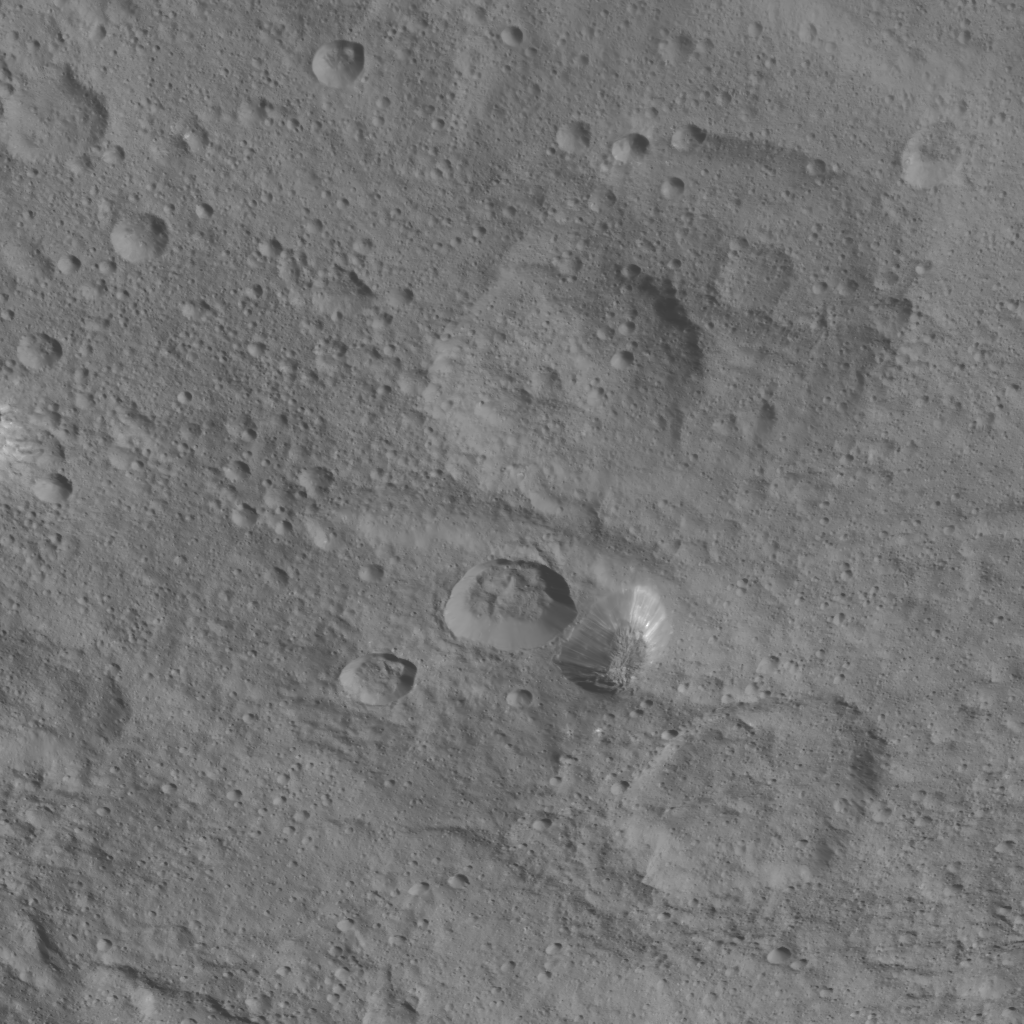} &
    \includegraphics[width=\linewidth]{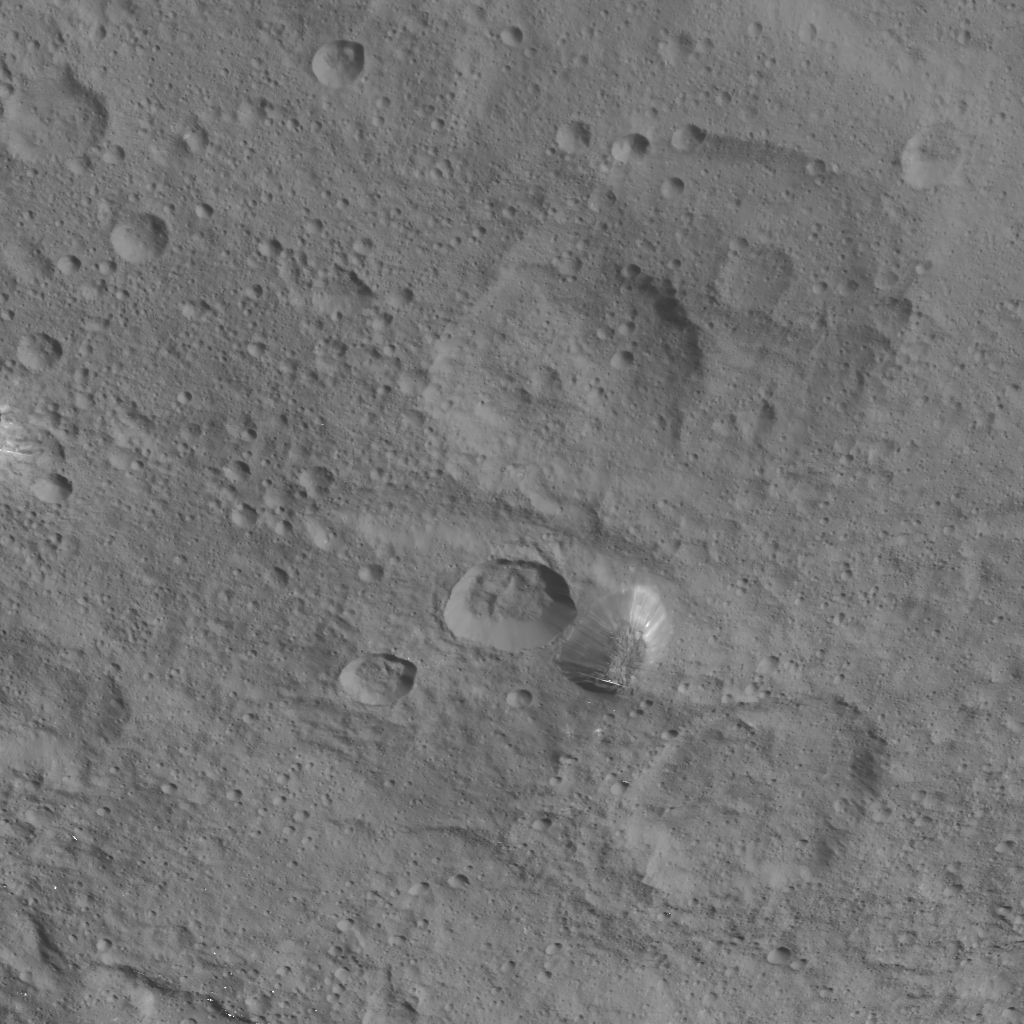} &
    \includegraphics[width=\linewidth]{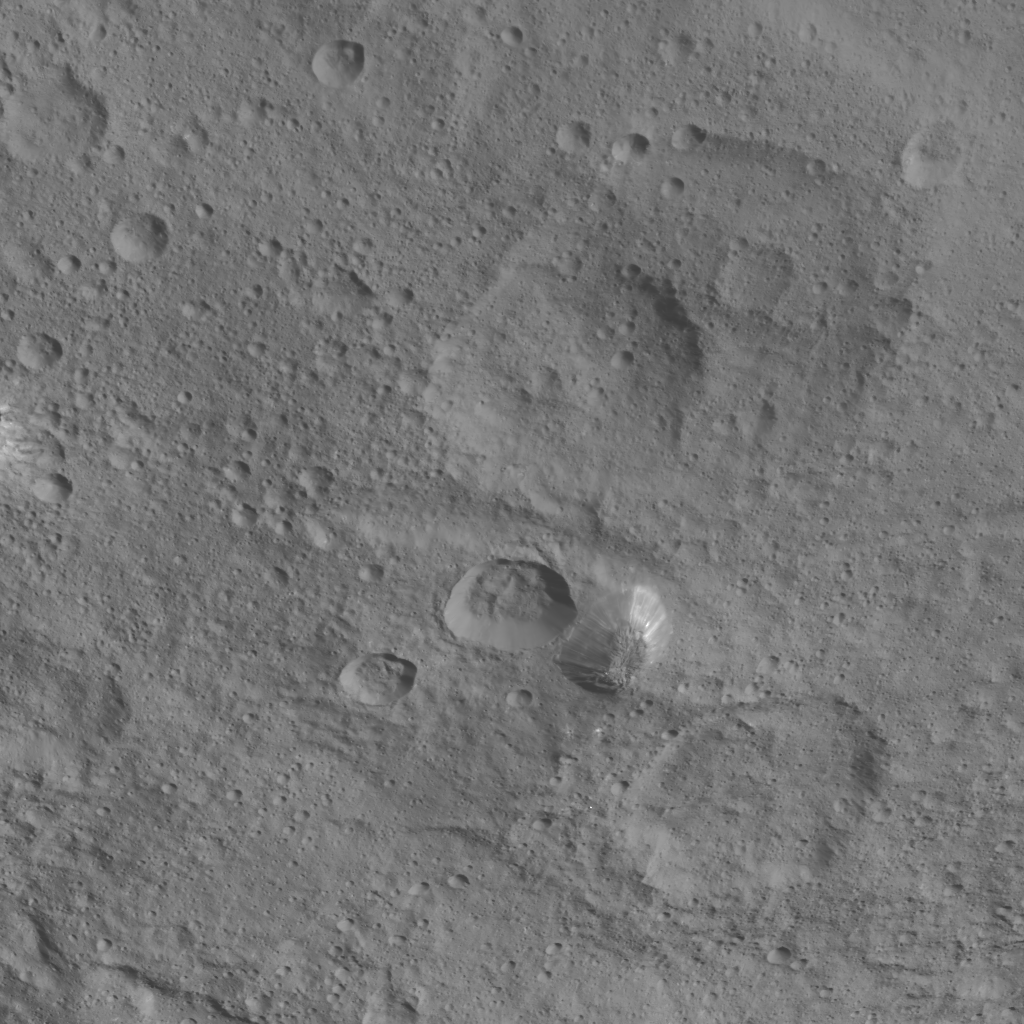} \\
    \parbox[t]{2.5mm}{\rotatebox[origin=l]{90}{\textcolor{gray}{\small{Ikapati}}}} & 
    \includegraphics[width=\linewidth]{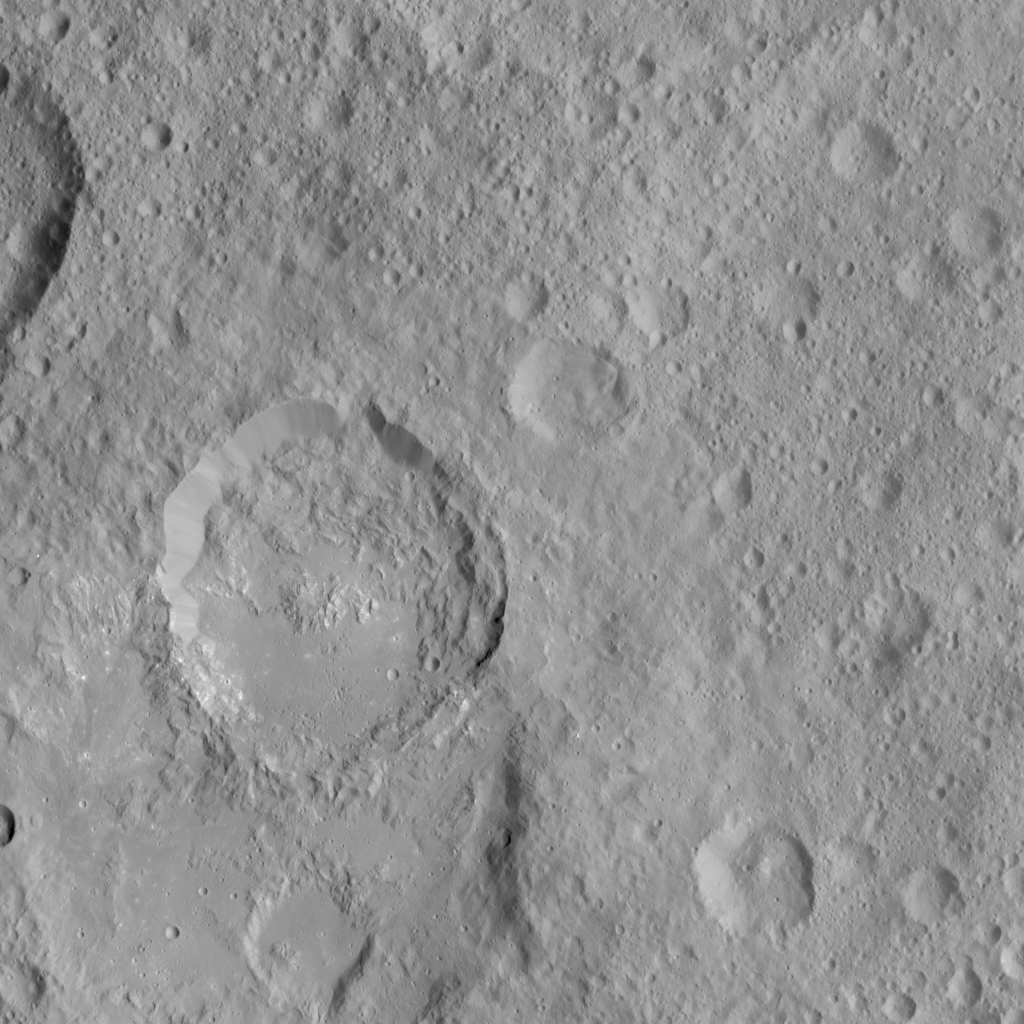} &
    \includegraphics[width=\linewidth]{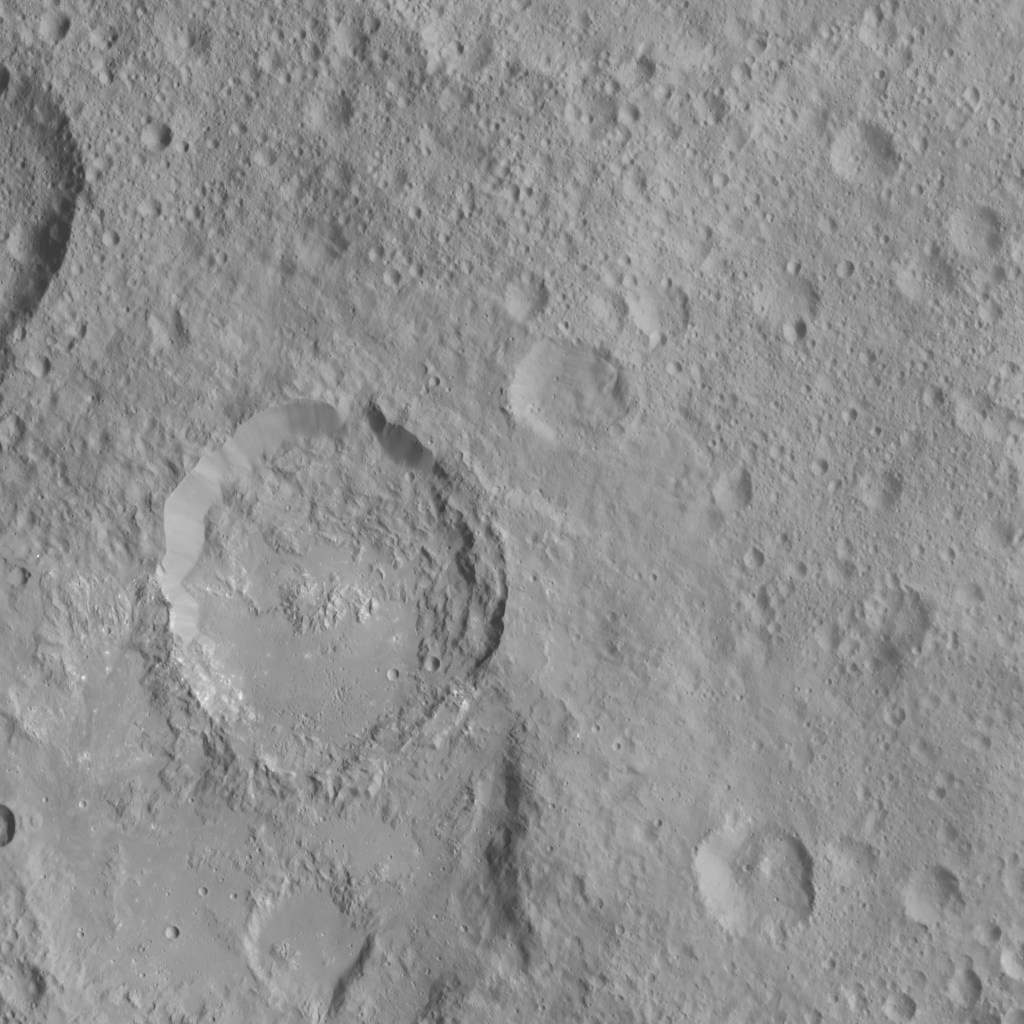} &
    \includegraphics[width=\linewidth]{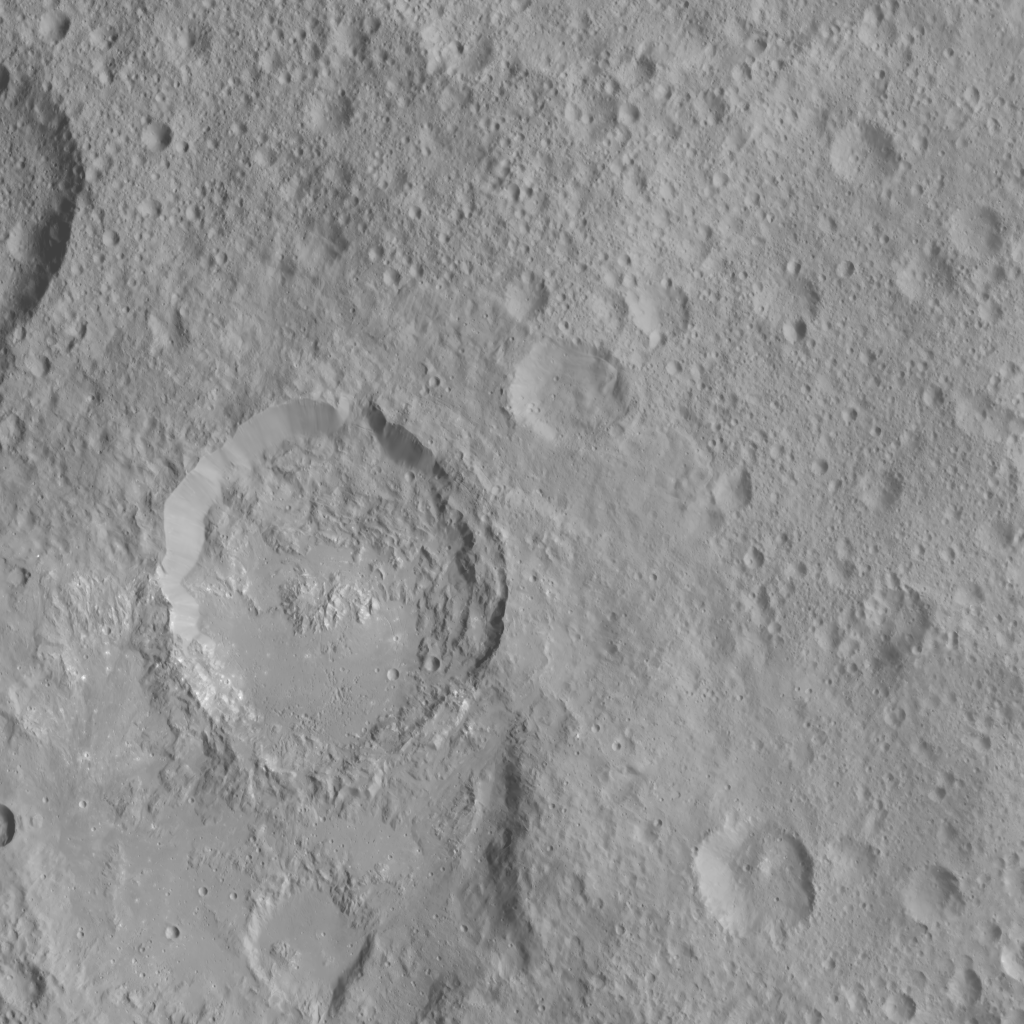} &
    \includegraphics[width=\linewidth]{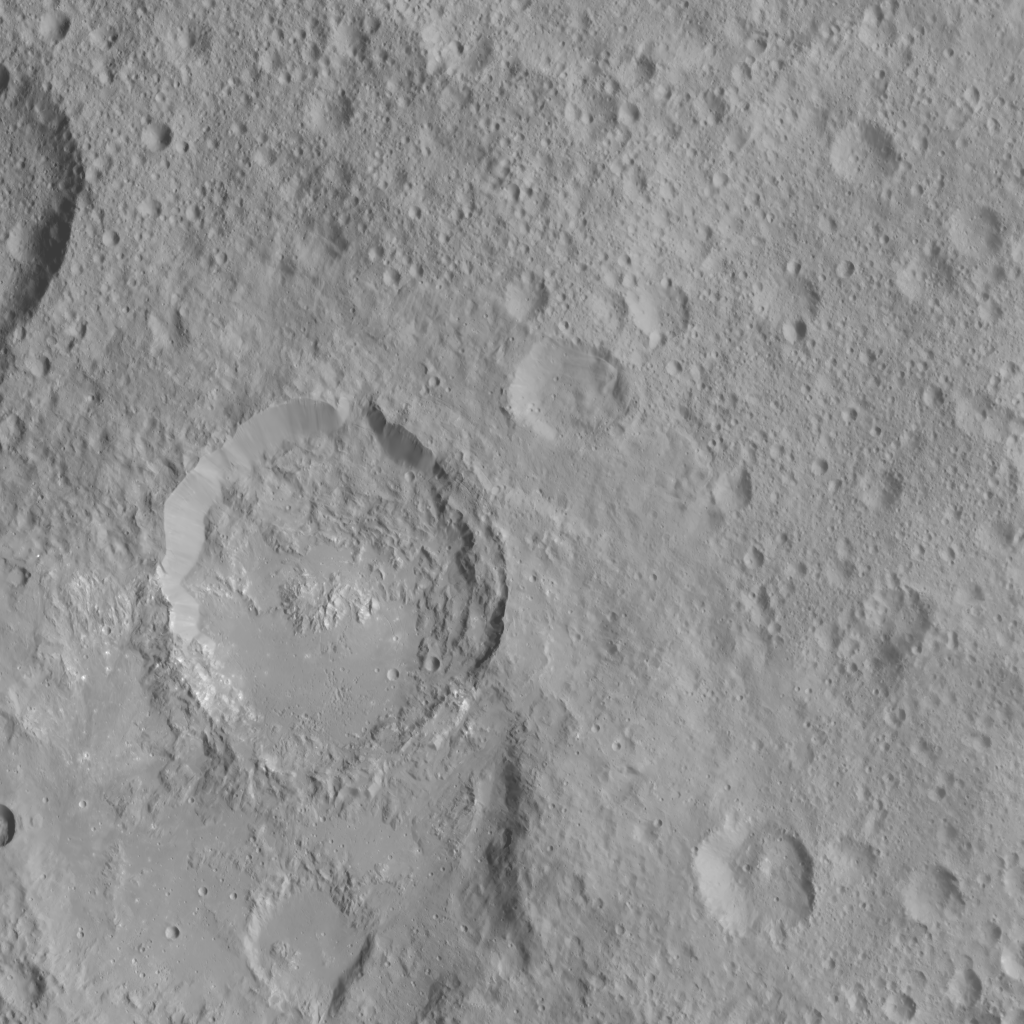} &
    \includegraphics[width=\linewidth]{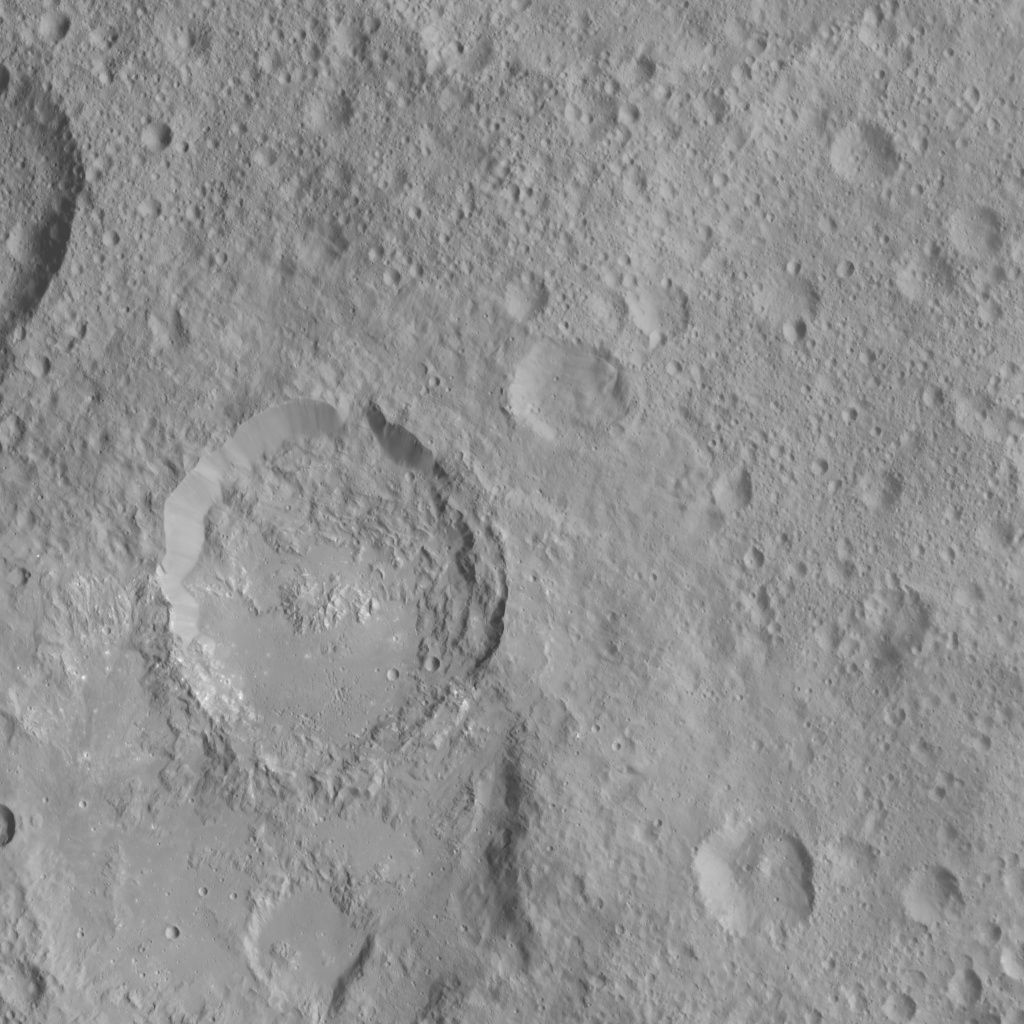} \\
\end{tabular}

%% file: fig/rendering-metrics.tex
\begin{tabular}{lccccccccc}
\toprule
& \multicolumn{3}{c}{Cornelia (27/2)} & \multicolumn{3}{c}{Ahuna Mons (30/2)} & \multicolumn{3}{c}{Ikapati (27/2)} \\
\cmidrule(lr){2-4} \cmidrule(lr){5-7} \cmidrule{8-10}
                 & PSNR $\uparrow$ & SSIM $\uparrow$ & LPIPS $\downarrow$ & PSNR $\uparrow$ & SSIM $\uparrow$ & LPIPS $\downarrow$ & PSNR $\uparrow$ & SSIM $\uparrow$ & LPIPS $\downarrow$ \\ 
\midrule
SH            &  36.59/35.17 & 0.9630/0.9614 & 0.1371/0.1280
           &  \underline{37.83}/36.73 & 0.9629/0.9561 & 0.0685/0.0738
           &  38.46/36.00 & 0.9805/0.9656 & 0.0416/0.0651
         \\
Lambert            &  \underline{39.23}/\textbf{39.32} & \textbf{0.9702}/\textbf{0.9689} & \textbf{0.1091}/\textbf{0.1017}
           &  36.58/\textbf{37.93} & \textbf{0.9653}/\textbf{0.9587} & \textbf{0.0630}/\textbf{0.0687}
           &  \underline{39.08}/\underline{36.93} & \textbf{0.9837}/\textbf{0.9683} & \textbf{0.0348}/\textbf{0.0597}
         \\
L-S            &  38.86/35.68 & 0.9687/0.9608 & 0.1174/0.1189
           &  \textbf{38.10}/37.10 & \underline{0.9642}/0.9551 & 0.0660/0.0761
           &  38.68/36.62 & 0.9795/0.9649 & 0.0427/0.0656
         \\
McEwen            &  \textbf{40.15}/\underline{38.35} & \underline{0.9695}/\underline{0.9651} & \underline{0.1149}/\underline{0.1101}
           &  37.21/\underline{37.88} & 0.9634/\underline{0.9569} & \underline{0.0654}/\underline{0.0718}
           &  \textbf{39.36}/\textbf{37.06} & \underline{0.9823}/\underline{0.9673} & \underline{0.0389}/\underline{0.0636}
         \\
\bottomrule
\end{tabular}

%% file: fig/normal-qual-compare.tex
\centering
\setlength{\tabcolsep}{2pt} % Default is 6pt
\begin{tabular}{cp{3.3cm}p{3.3cm}p{3.3cm}p{3.3cm}p{3.3cm}p{3.3cm}}
    & \multicolumn{1}{c}{\small{Image}} & \multicolumn{1}{c}{\small{Ground Truth}} & \multicolumn{1}{c}{\small{SH}} & \multicolumn{1}{c}{\small{Lambert}} & \multicolumn{1}{c}{\small{Lommel-Seeliger}}  & \multicolumn{1}{c}{\small{Lunar-Lambert}} \\
    \parbox[t]{2.5mm}{\rotatebox[origin=l]{90}{\textcolor{gray}{\small{Cornelia}}}} &
    \includegraphics[width=\linewidth]{fig/render/cornelia/GT/00000.png} & %FC21B0008195_11275021648F1G.png
    \includegraphics[width=\linewidth]{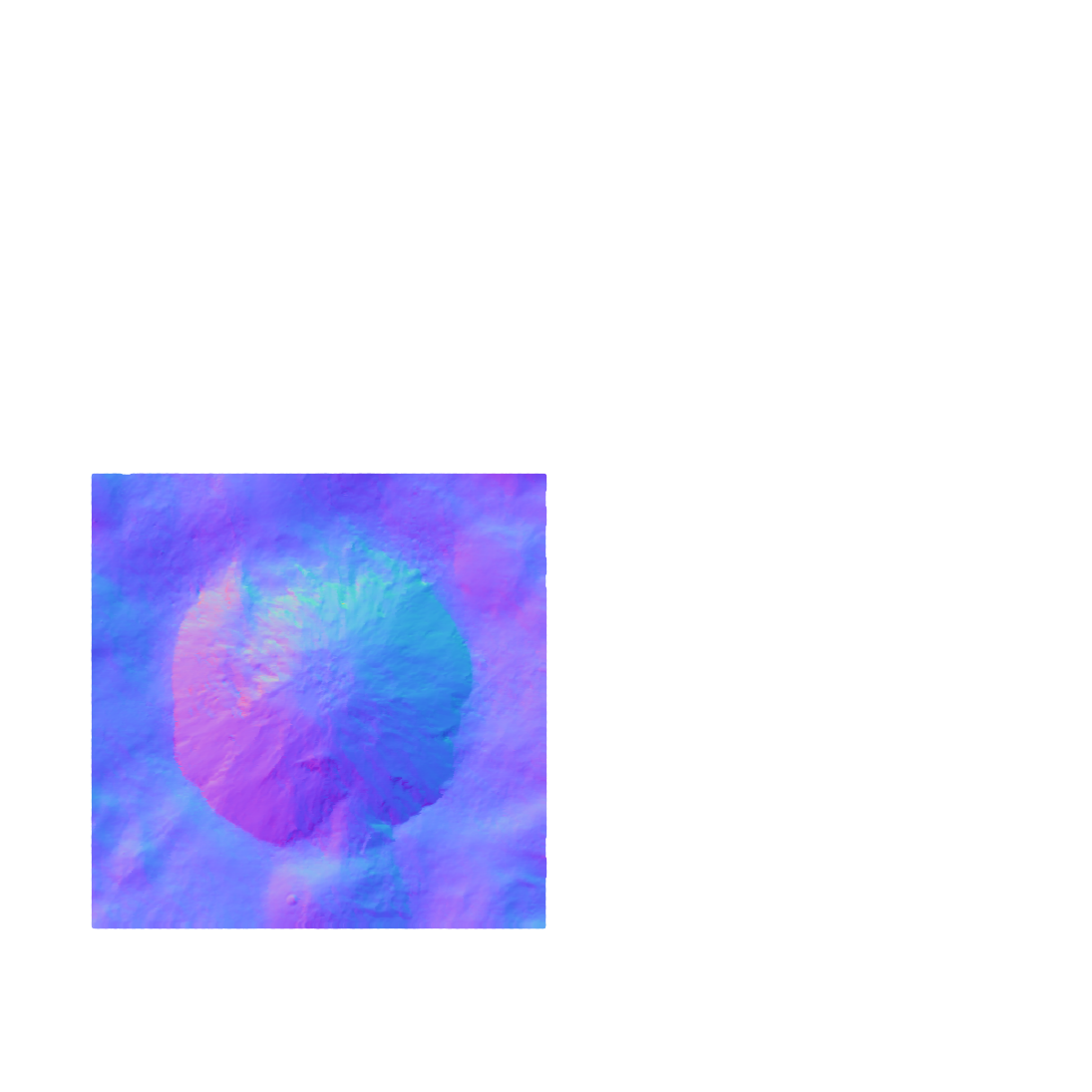} &
    \includegraphics[width=\linewidth]{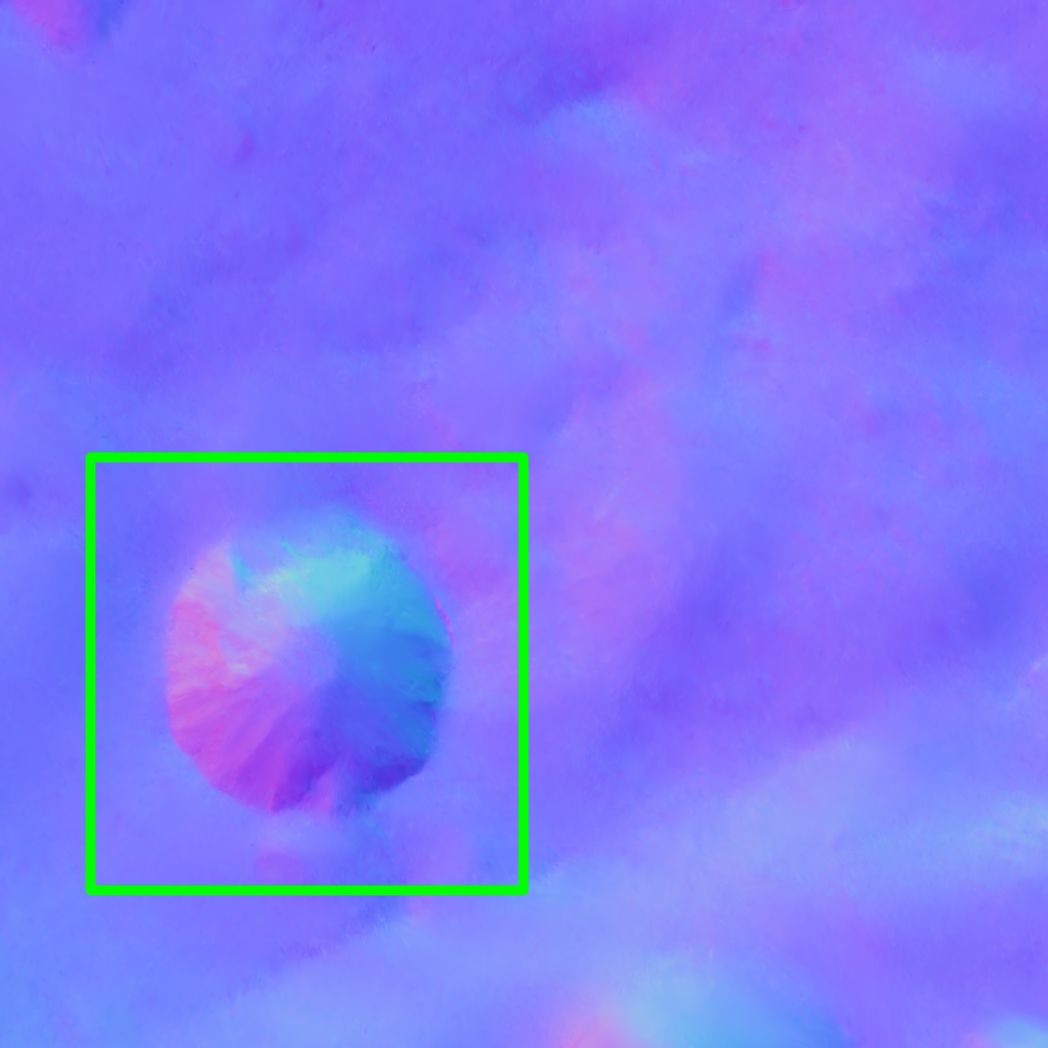} &
    \includegraphics[width=\linewidth]{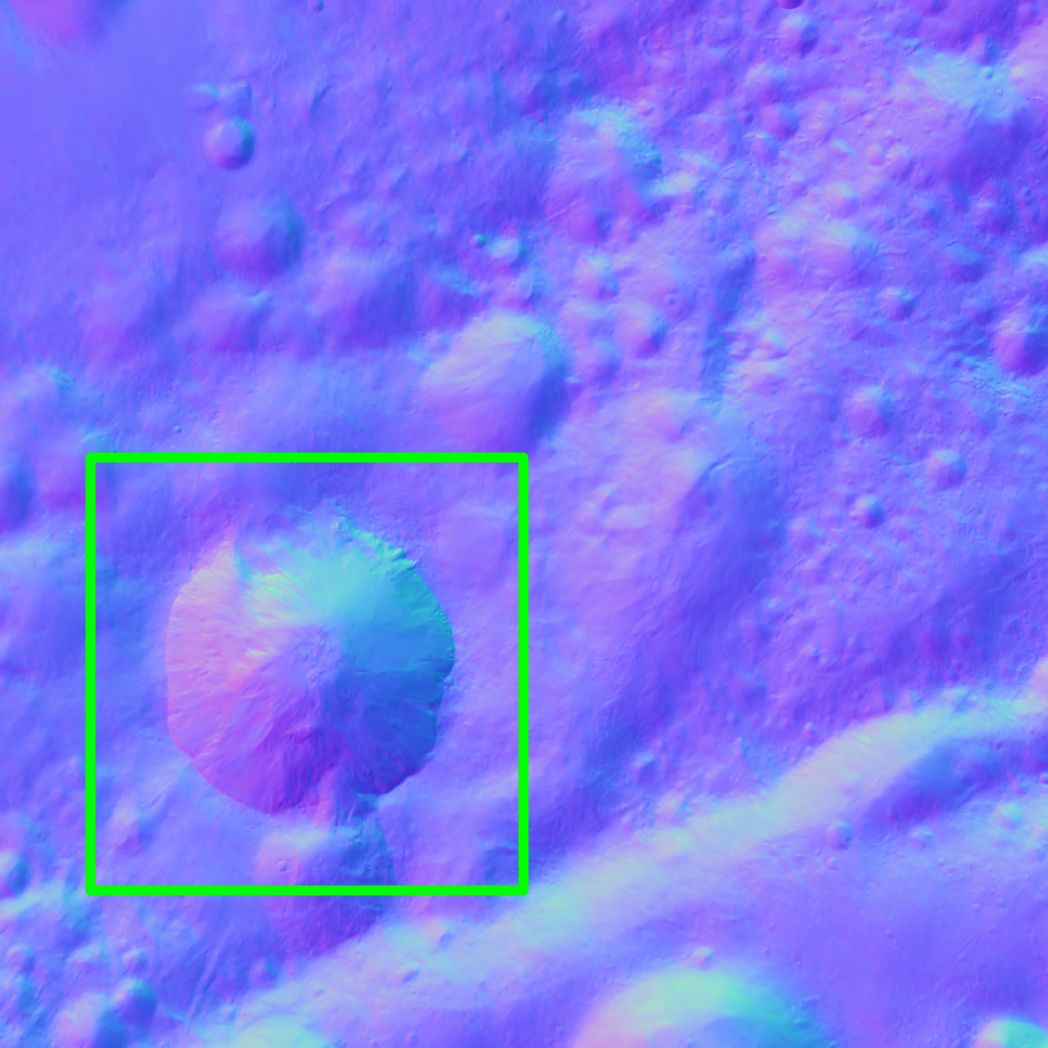} &
    \includegraphics[width=\linewidth]{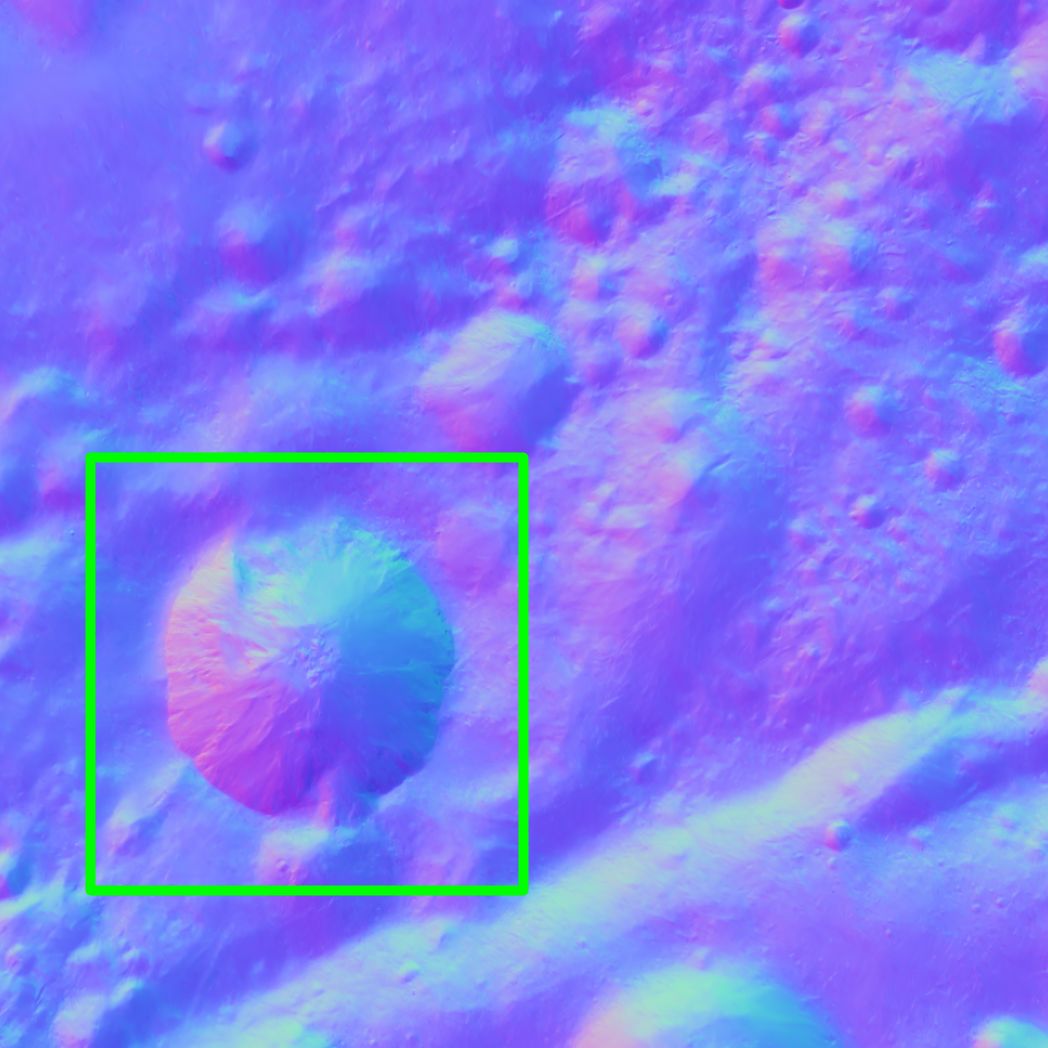} &
    \includegraphics[width=\linewidth]{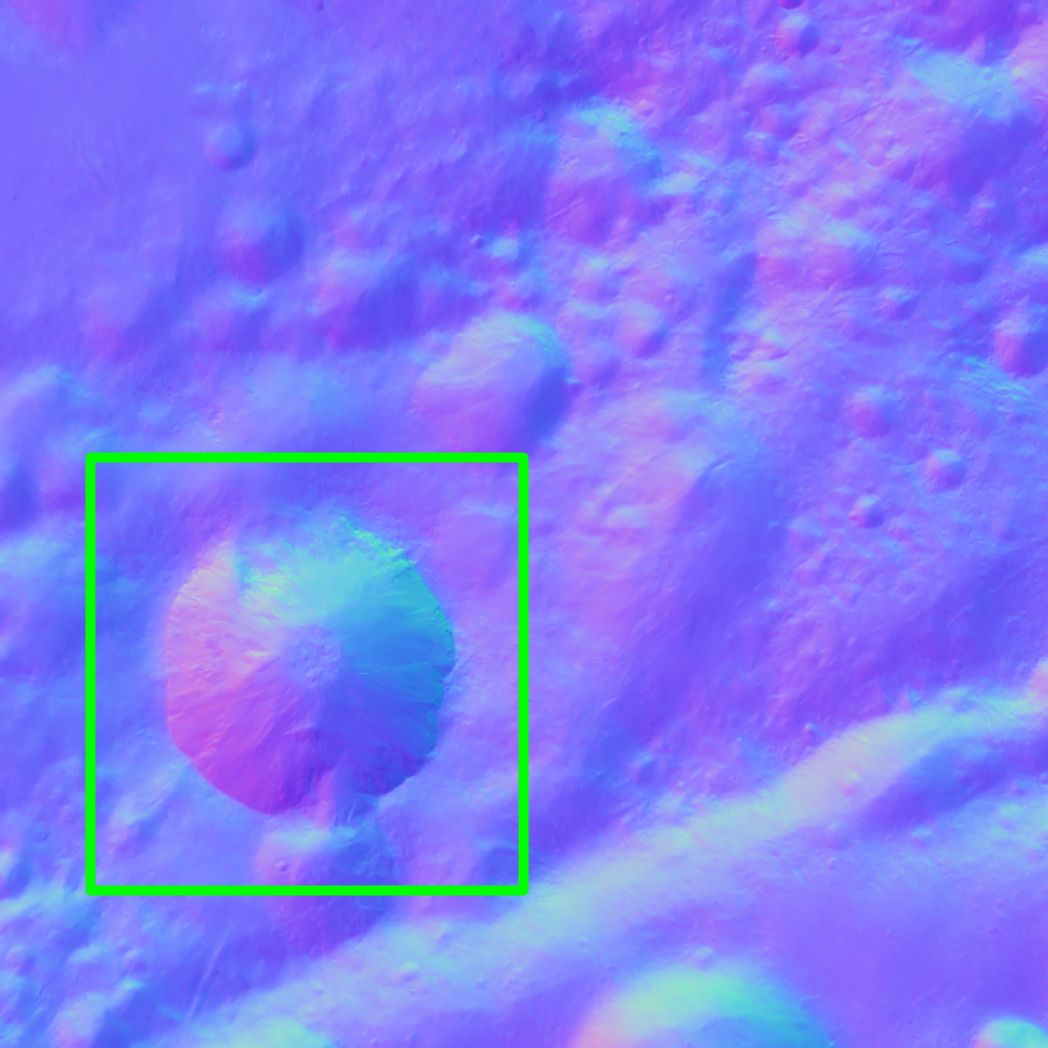} \\
    \parbox[t]{2.5mm}{\rotatebox[origin=l]{90}{\textcolor{gray}{\small{Ahuna Mons}}}} &
    \includegraphics[width=\linewidth]{fig/render/ahunamons/GT/00001.png} & %FC21B0008195_11275021648F1G.png
    \includegraphics[width=\linewidth]{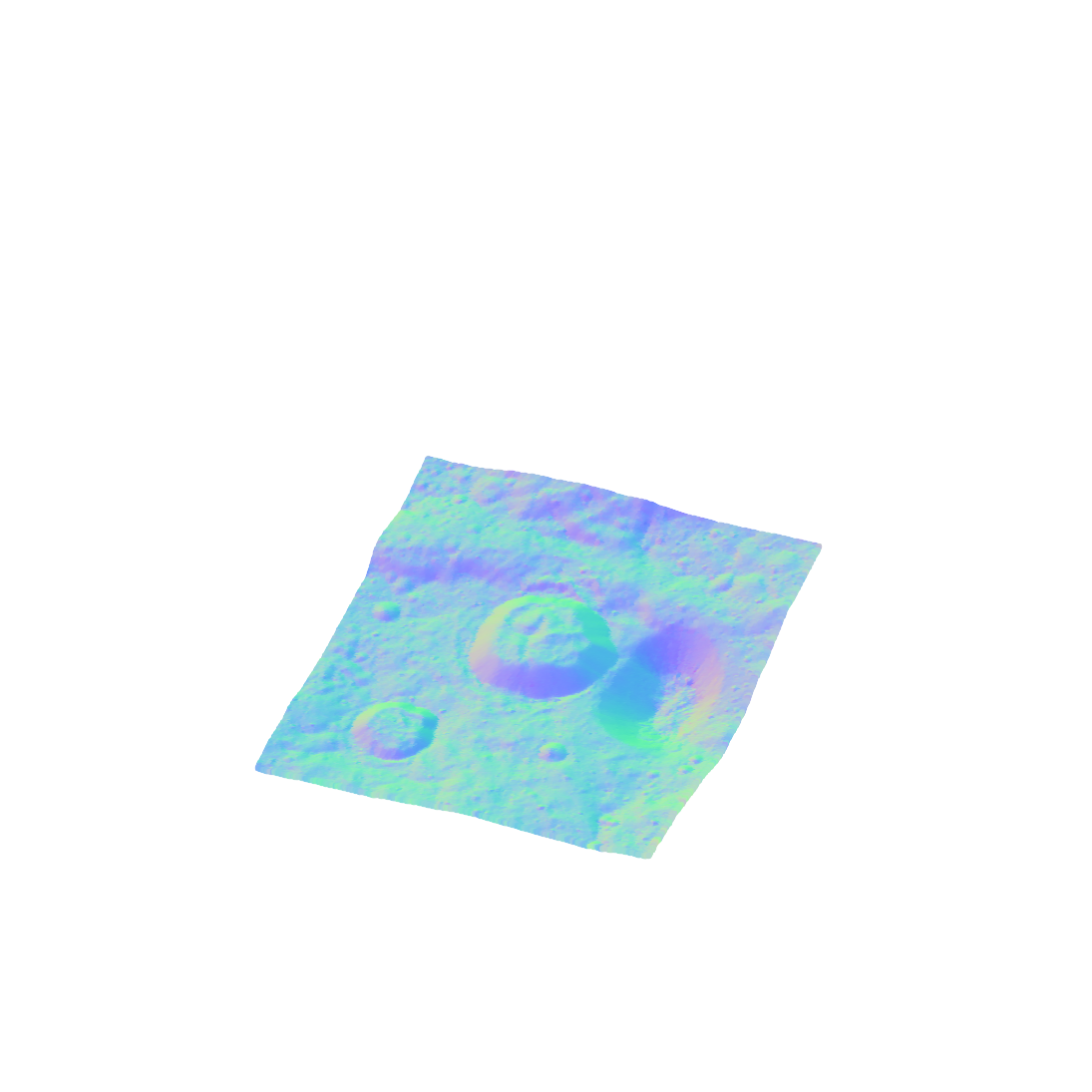} &
    \includegraphics[width=\linewidth]{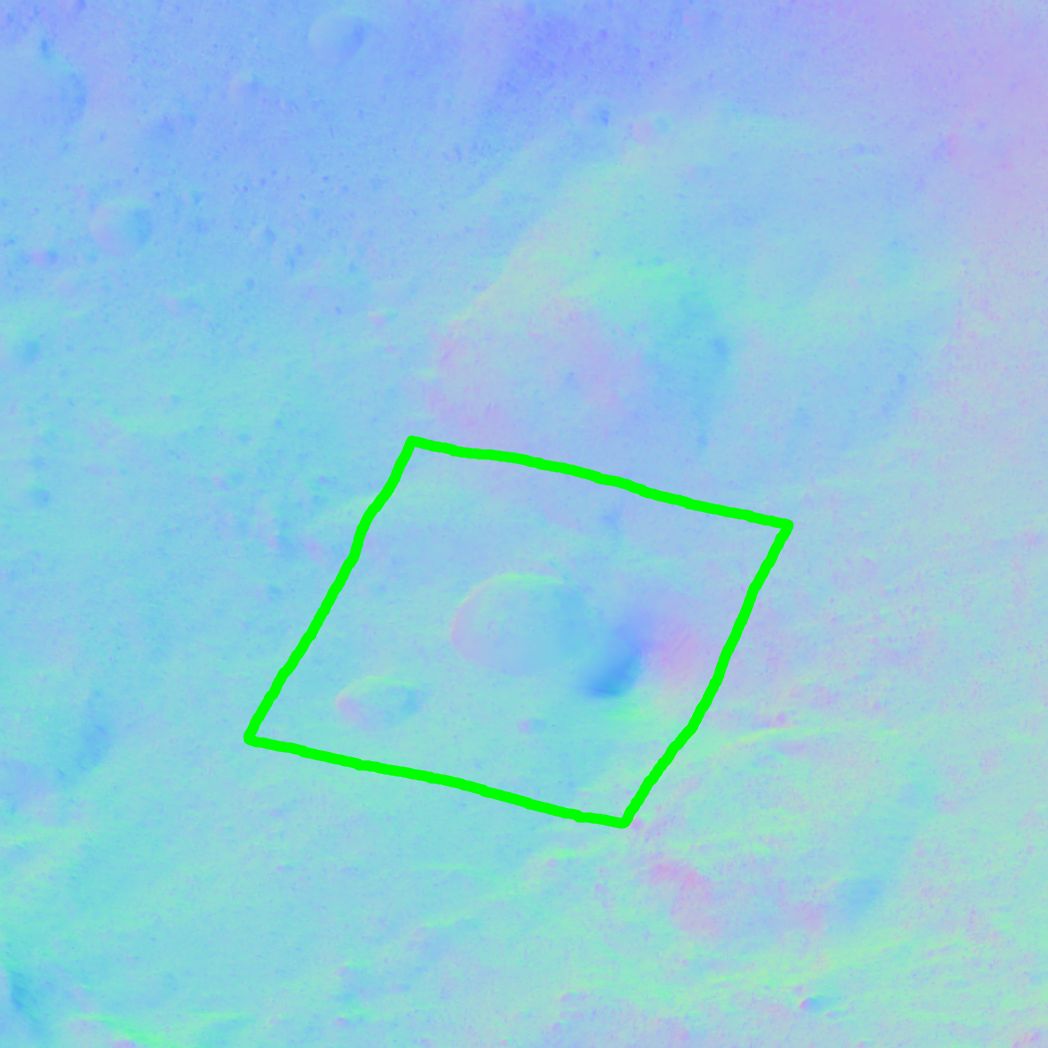} &
    \includegraphics[width=\linewidth]{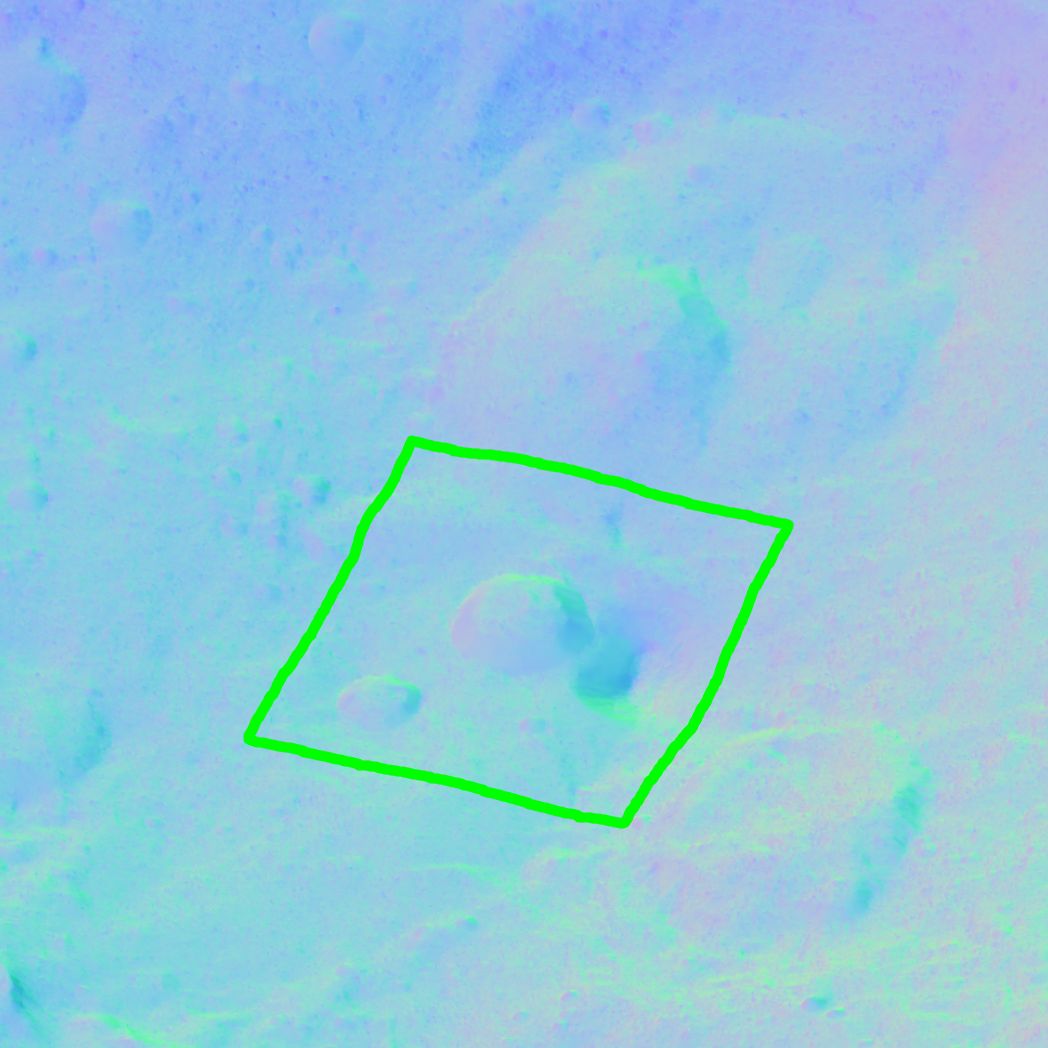} &
    \includegraphics[width=\linewidth]{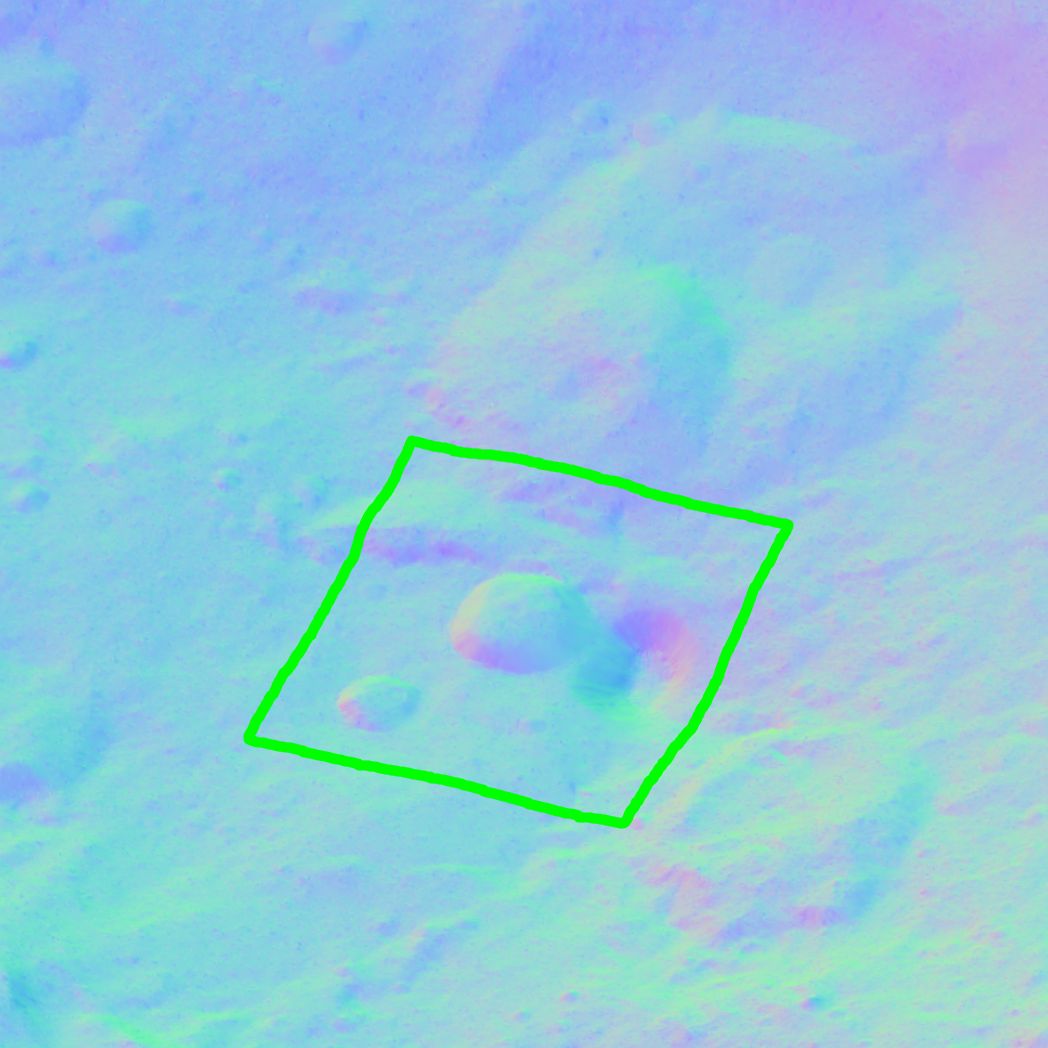} &
    \includegraphics[width=\linewidth]{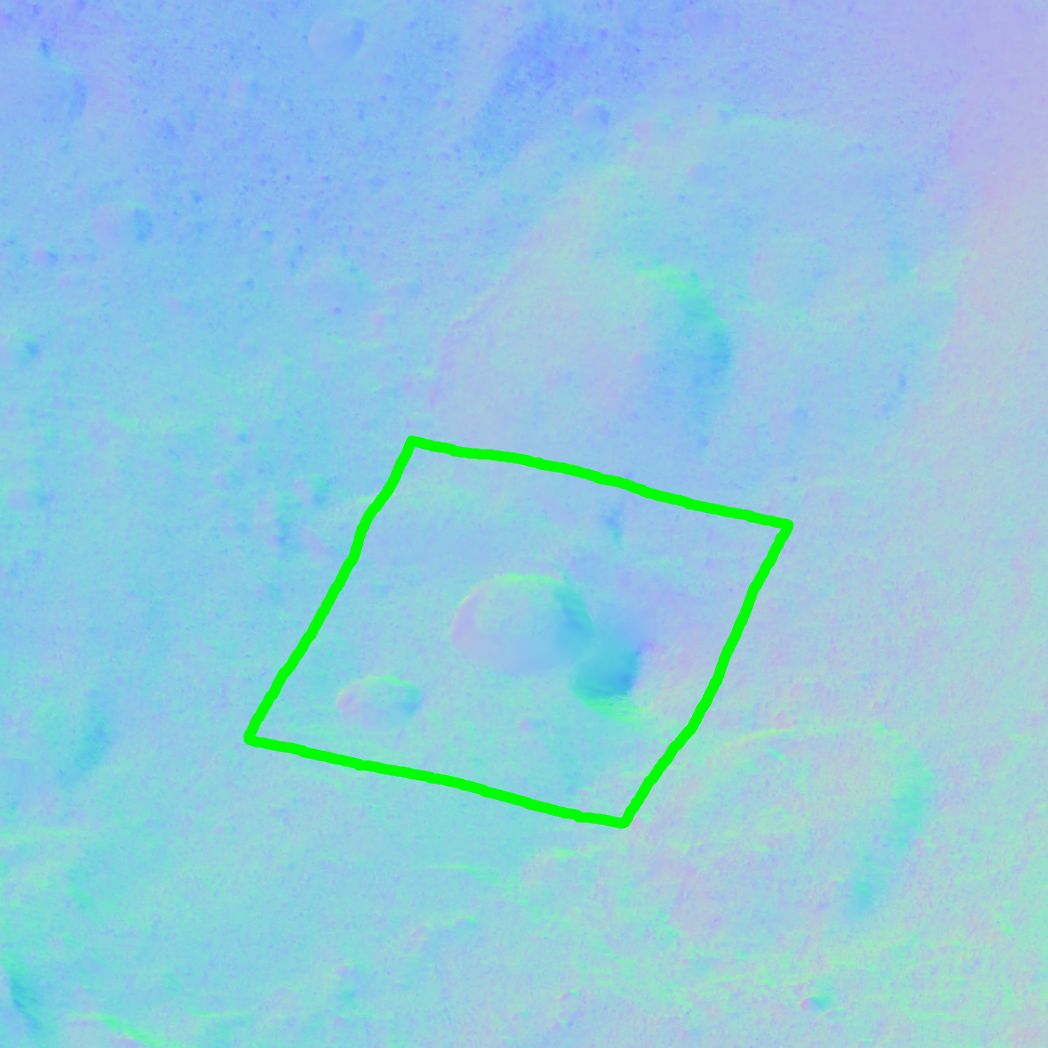} \\
    \parbox[t]{2.5mm}{\rotatebox[origin=l]{90}{\textcolor{gray}{\small{Ikapati}}}} &
    \includegraphics[width=\linewidth]{fig/render/ikapati/GT/00000.png} & %FC21B0008195_11275021648F1G.png
    \includegraphics[width=\linewidth]{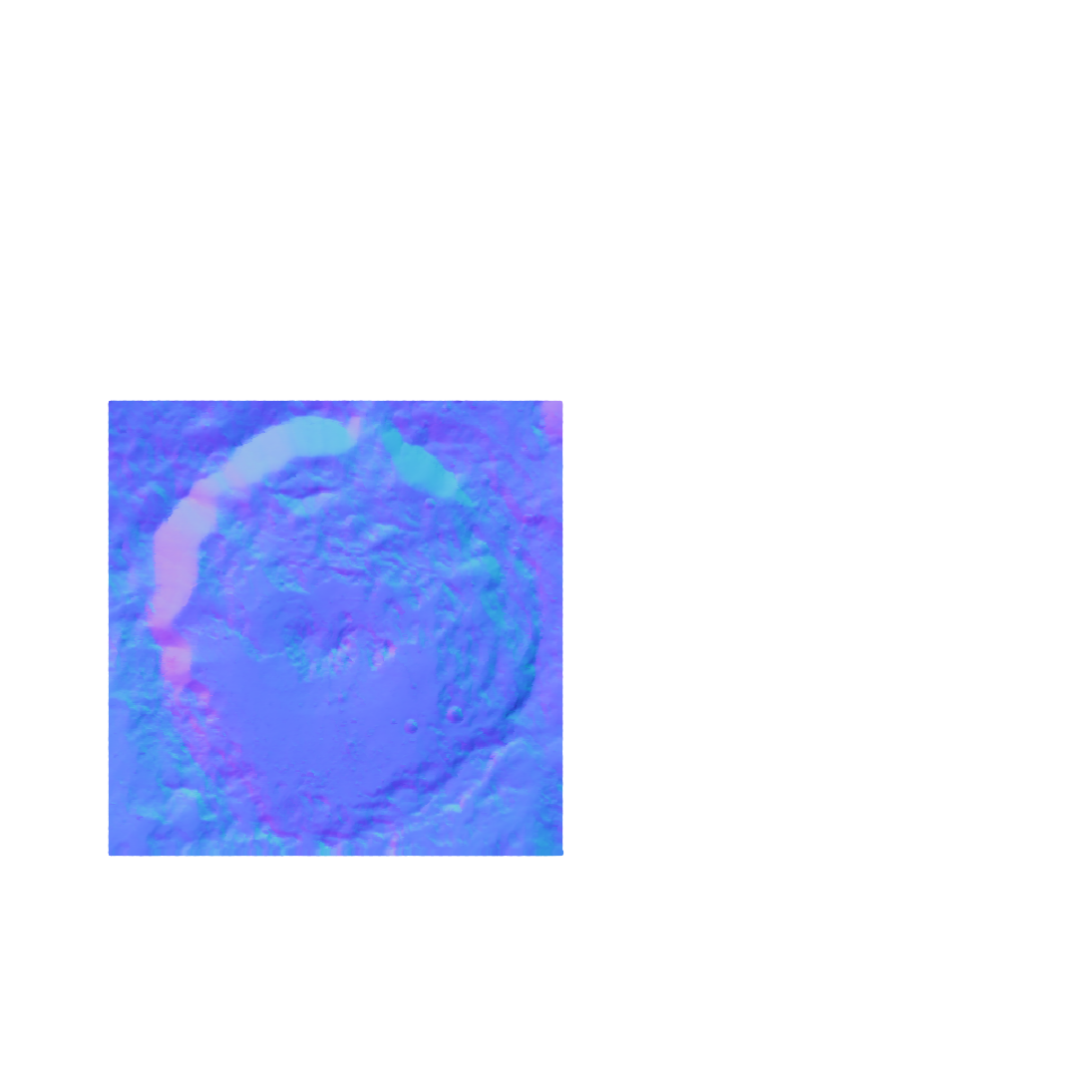} &
    \includegraphics[width=\linewidth]{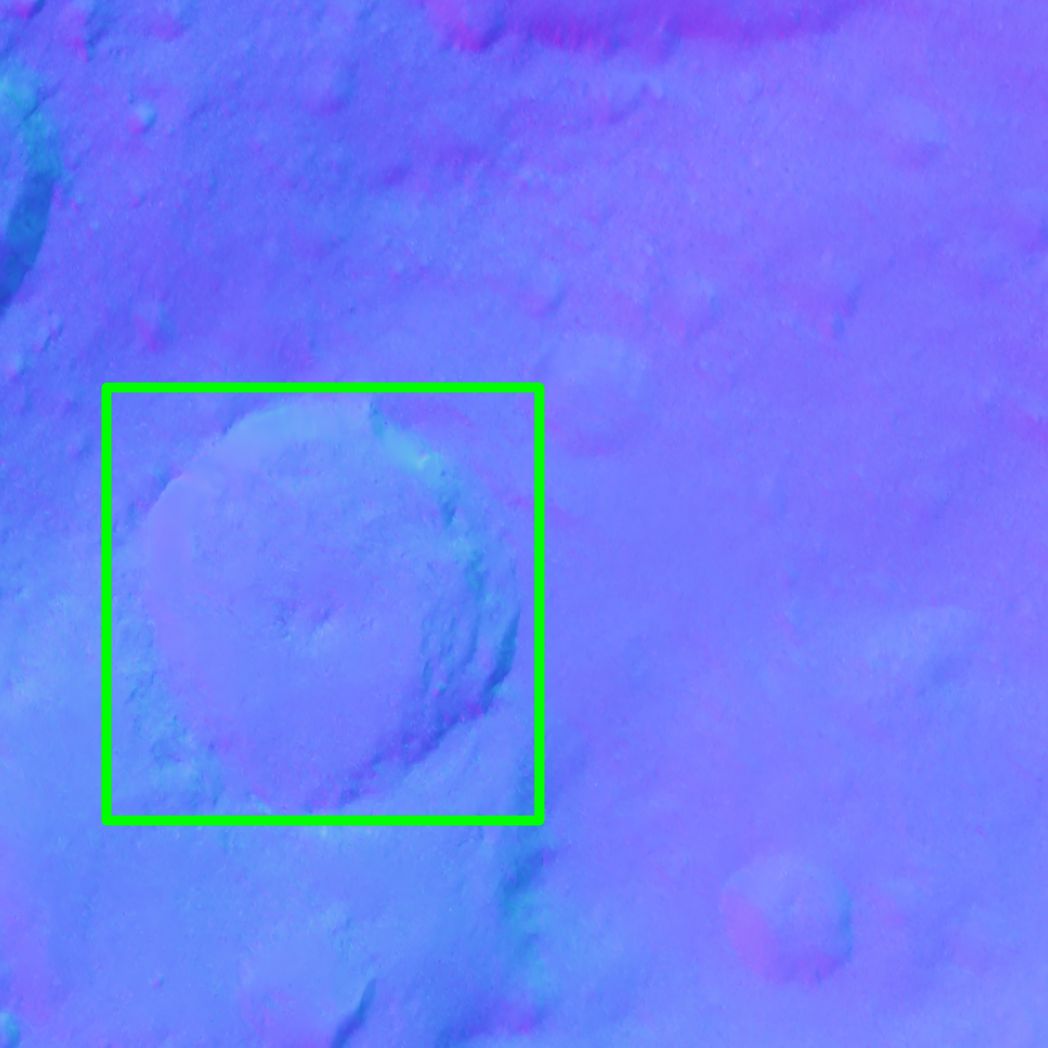} &
    \includegraphics[width=\linewidth]{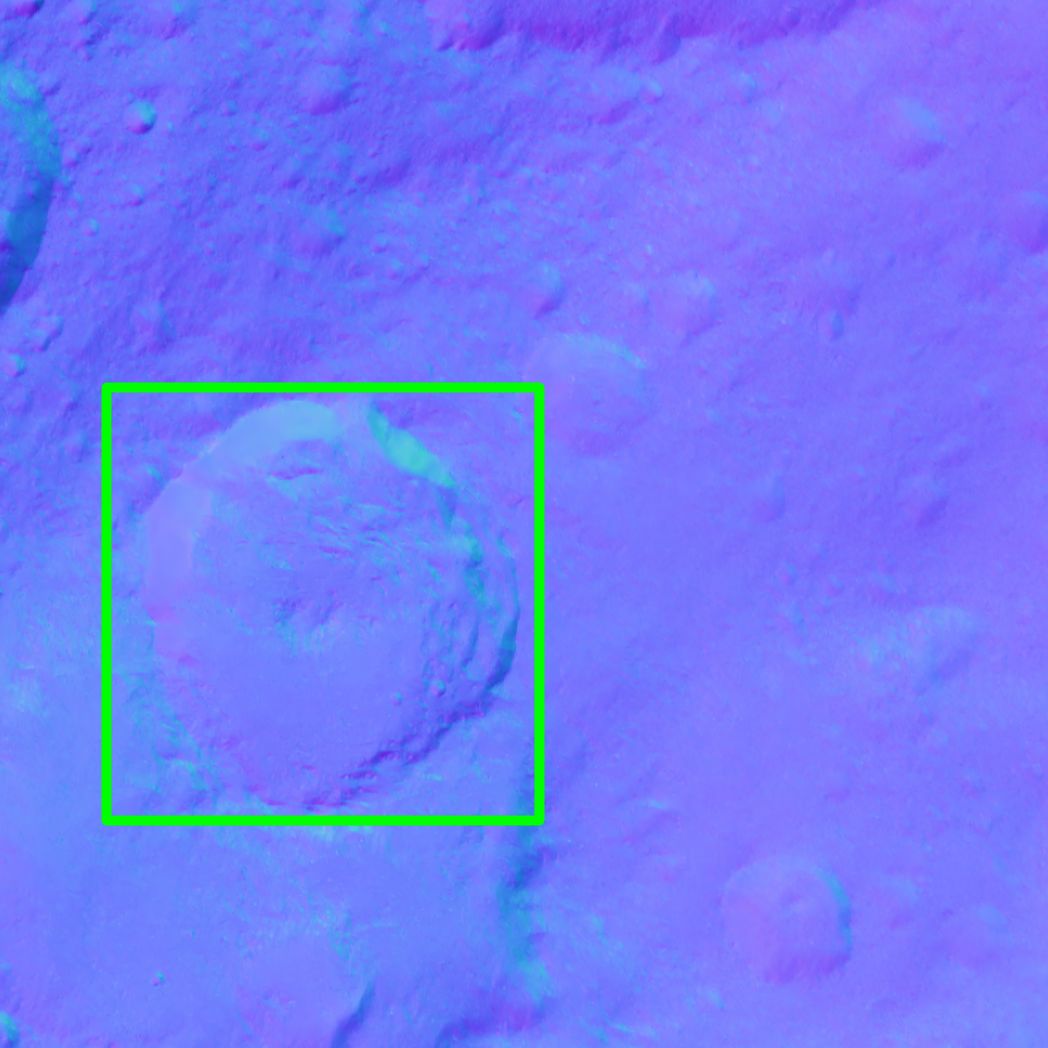} &
    \includegraphics[width=\linewidth]{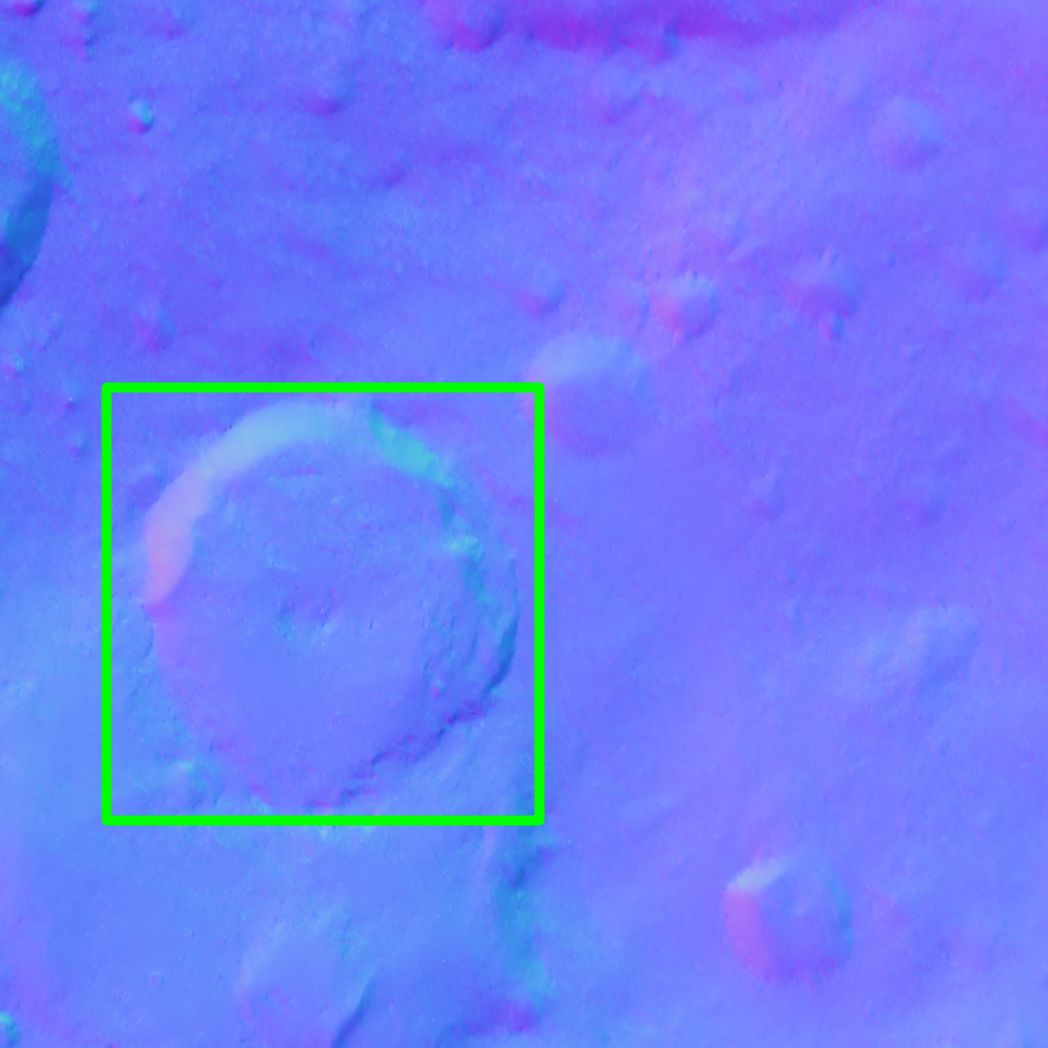} &
    \includegraphics[width=\linewidth]{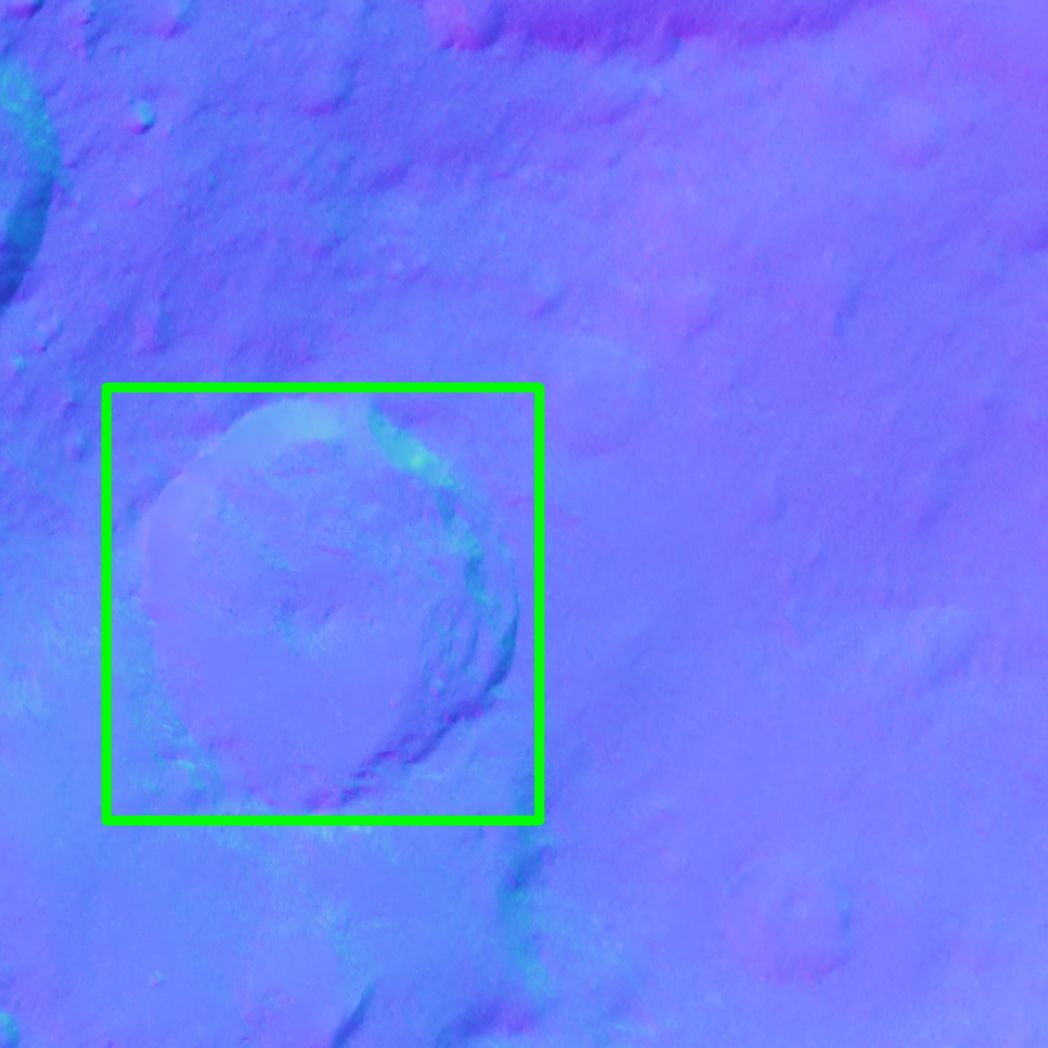} \\
\end{tabular}

%% file: fig/normal-albedo-metrics.tex
\begin{tabular}{lccccccccc}
\toprule
& \multicolumn{3}{c}{Cornelia (27/2)} & \multicolumn{3}{c}{Ahuna Mons (30/2)} & \multicolumn{3}{c}{Ikapati (27/2)} \\
\cmidrule(lr){2-4} \cmidrule(lr){5-7} \cmidrule{8-10}
                 & $\delta\theta\ ({}^\circ)$ $\downarrow$ & $\delta a$ $\downarrow$ & $d_{H}$ $\downarrow$ & $\delta\theta\ ({}^\circ)$ $\downarrow$ & $\delta a$ $\downarrow$ & $d_{H}$ $\downarrow$ & $\delta\theta\ ({}^\circ)$ $\downarrow$ & $\delta a$ $\downarrow$ & $d_{H}$ $\downarrow$ \\ 
\midrule
SH            &  7.4510/7.4873 & --/-- & 0.0032
           &  10.3486/10.4027 & --/-- & 0.0061
           &  8.4515/8.2710 & --/-- & 0.0033
         \\
Lambert            &  \underline{6.4468}/6.2974 & \textbf{0.0278}/\textbf{0.0474} & \textbf{0.0015}
           &  9.3419/9.5286 & \underline{0.0334}/\textbf{0.0310} & \underline{0.0053}
           &  \underline{7.4277}/\underline{7.3510} & \underline{0.0331}/0.0295 & \textbf{0.0021}
         \\
L-S            &  \textbf{6.3027}/\textbf{6.0265} & \underline{0.0409}/0.0672 & 0.0021
           &  \textbf{8.3245}/\textbf{8.2817} & 0.0341/\underline{0.0312} & \textbf{0.0021}
           &  \textbf{7.0273}/\textbf{6.8529} & 0.0332/\textbf{0.0292} & \underline{0.0026}
         \\
McEwen            &  6.5053/\underline{6.0975} & 0.0594/\underline{0.0614} & \underline{0.0016}
           &  \underline{9.3279}/\underline{9.4170} & \textbf{0.0316}/0.0313 & 0.0057
           &  7.9741/7.7830 & \textbf{0.0282}/\underline{0.0293} & 0.0040
         \\
\bottomrule
\end{tabular}

%% file: fig/albedo-qual-compare.tex
\centering
\setlength{\tabcolsep}{2pt} % Default is 6pt
\begin{tabular}{cp{3.3cm}p{3.3cm}p{3.3cm}p{3.3cm}p{3.3cm}p{3.3cm}}
    & \multicolumn{1}{c}{\small{Image}} & \multicolumn{1}{c}{\small{Ground Truth}} & \multicolumn{1}{c}{\small{Lambert}} & \multicolumn{1}{c}{\small{Lommel-Seeliger}}  & \multicolumn{1}{c}{\small{Lunar-Lambert}} \\
    \parbox[t]{2.5mm}{\rotatebox[origin=l]{90}{\textcolor{gray}{\small{Cornelia}}}} &
    \includegraphics[width=\linewidth]{fig/render/cornelia/GT/00000.png} & %FC21B0008195_11275021648F1G.png
    \includegraphics[width=\linewidth]{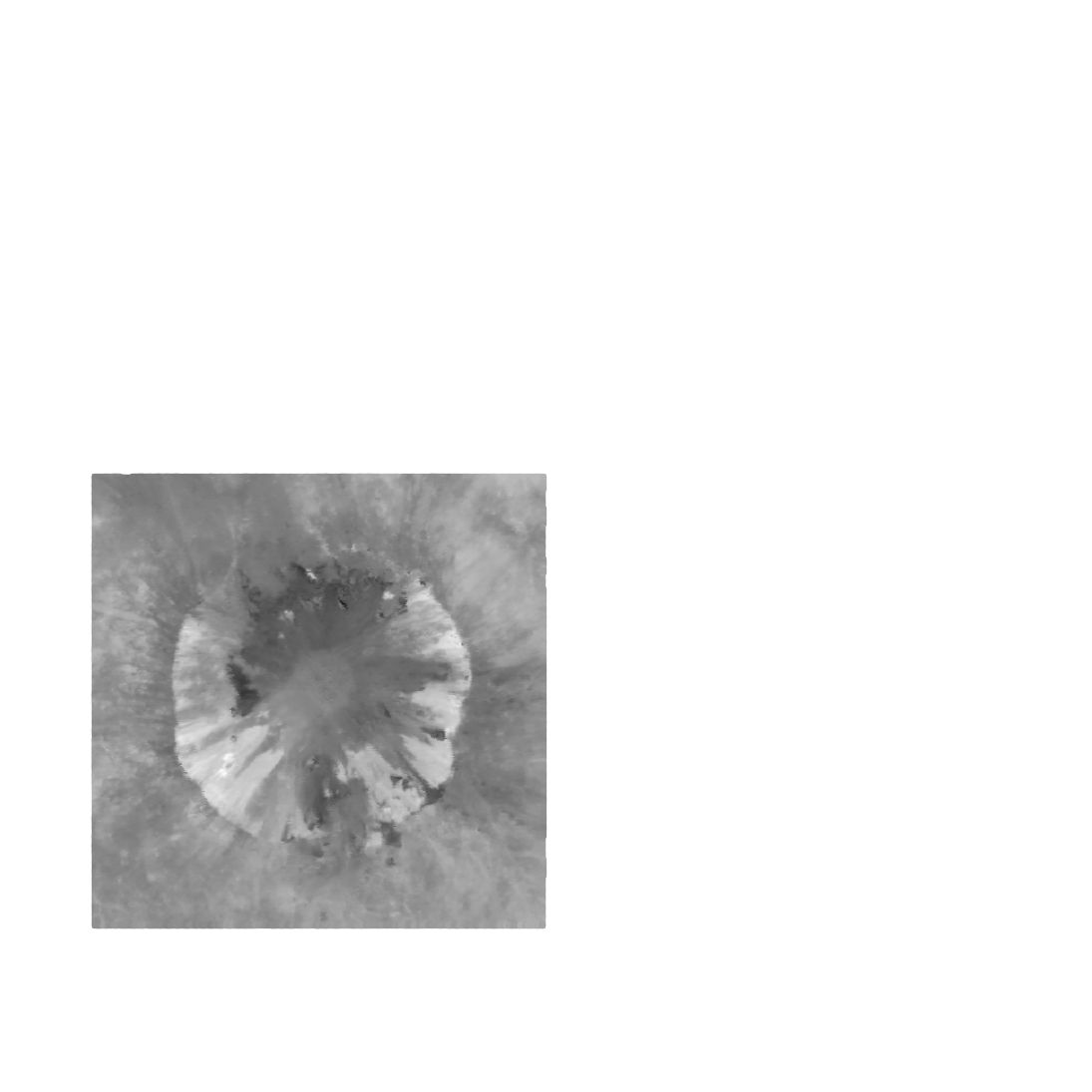} &
    \includegraphics[width=\linewidth]{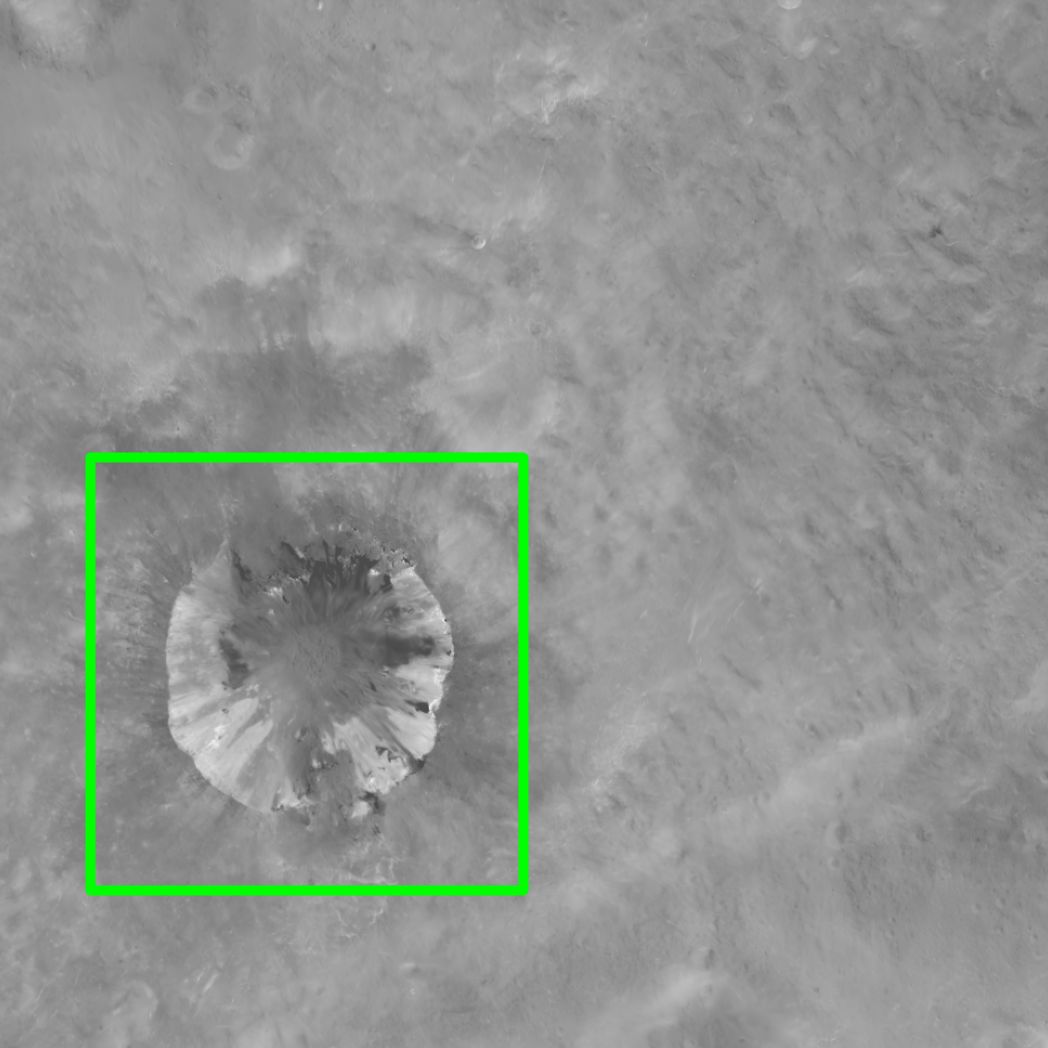} &
    \includegraphics[width=\linewidth]{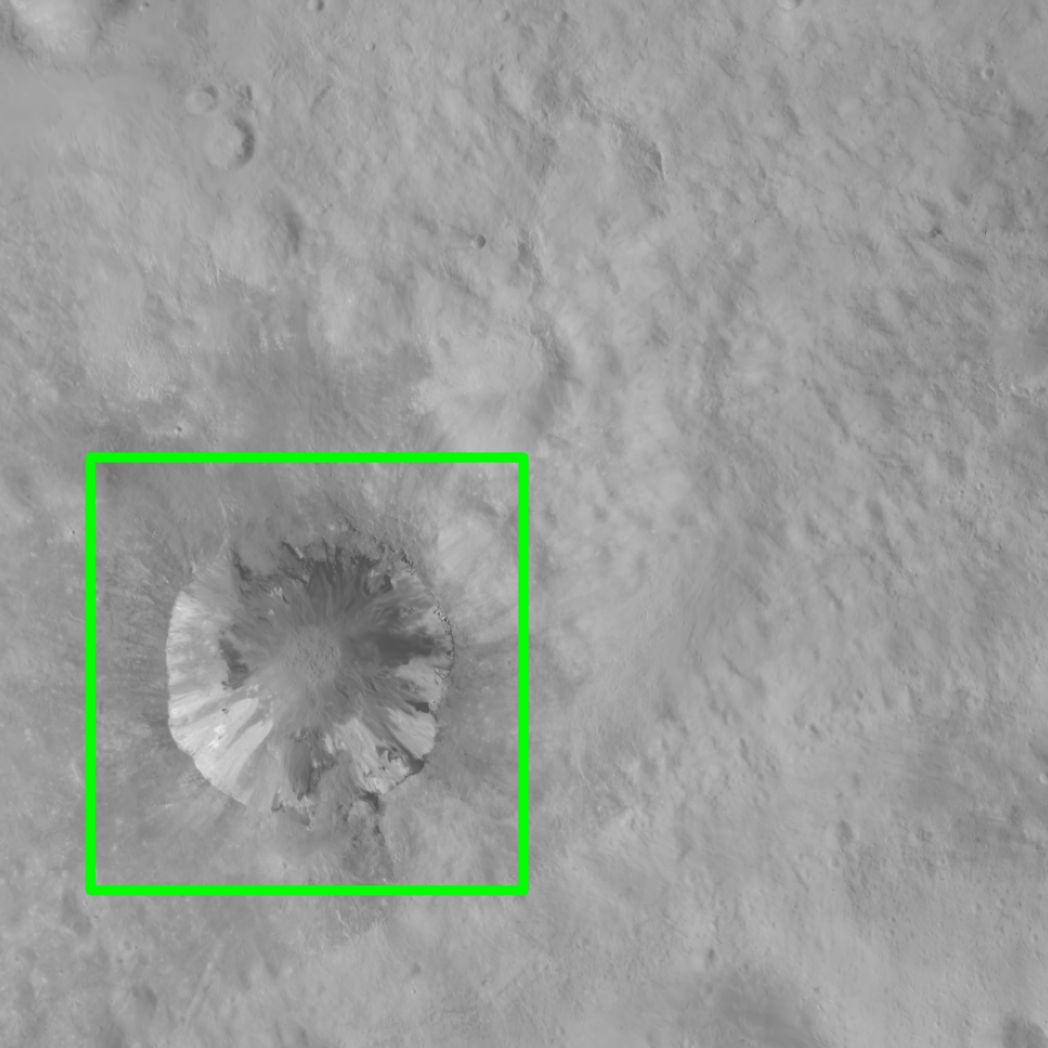} &
    \includegraphics[width=\linewidth]{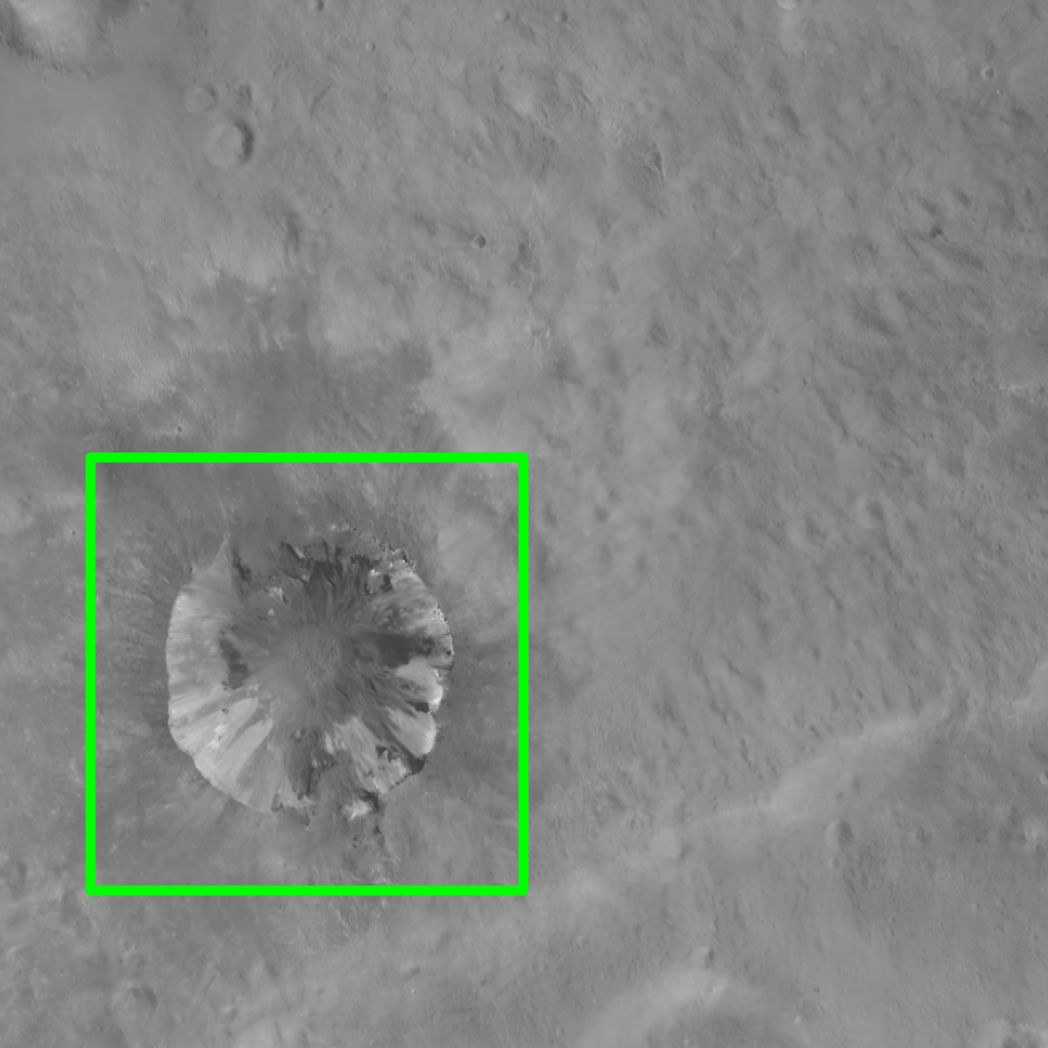} \\
    \parbox[t]{2.5mm}{\rotatebox[origin=l]{90}{\textcolor{gray}{\small{Ahuna Mons}}}} &
    \includegraphics[width=\linewidth]{fig/render/ahunamons/GT/00001.png} & %FC21B0008195_11275021648F1G.png
    \includegraphics[width=\linewidth]{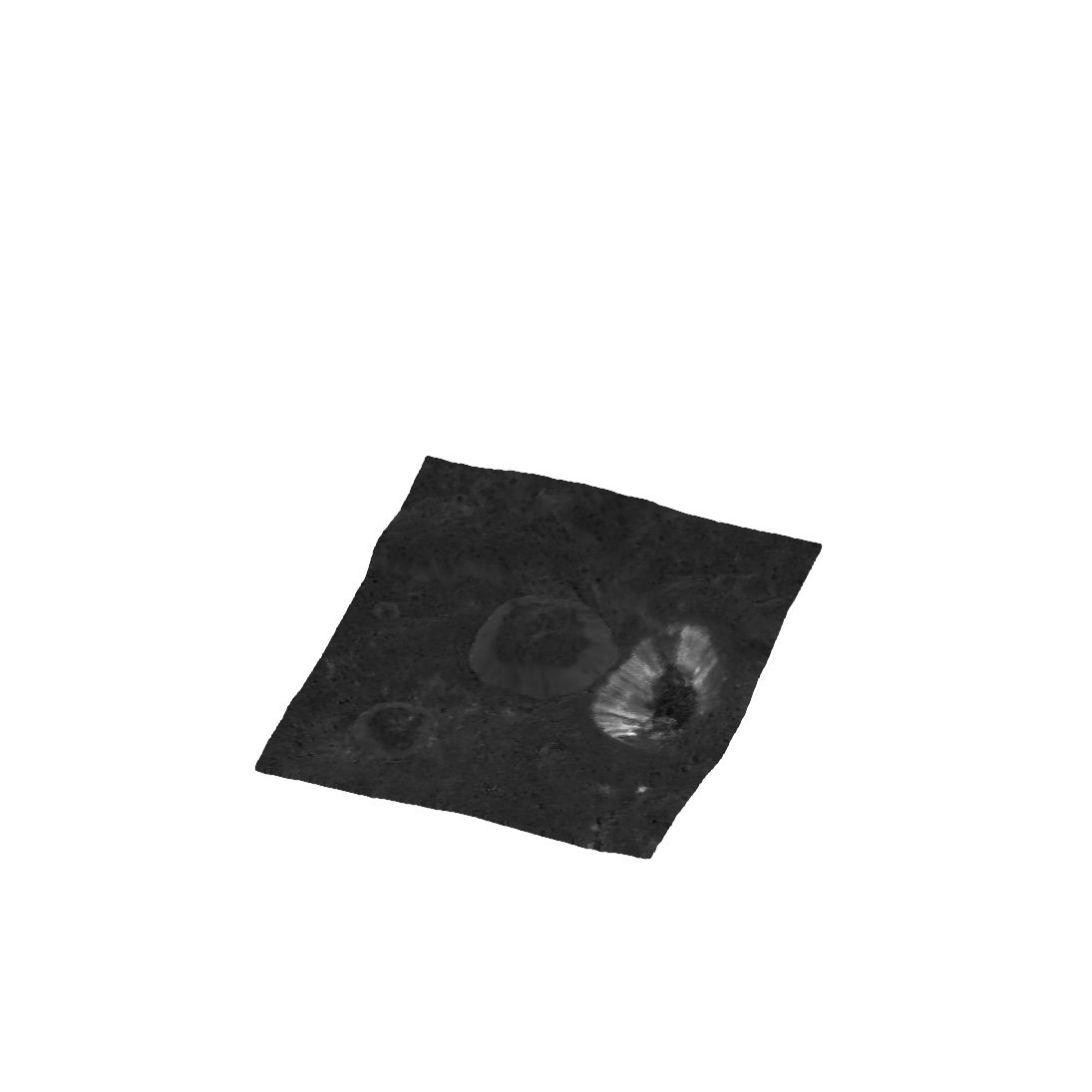} &
    \includegraphics[width=\linewidth]{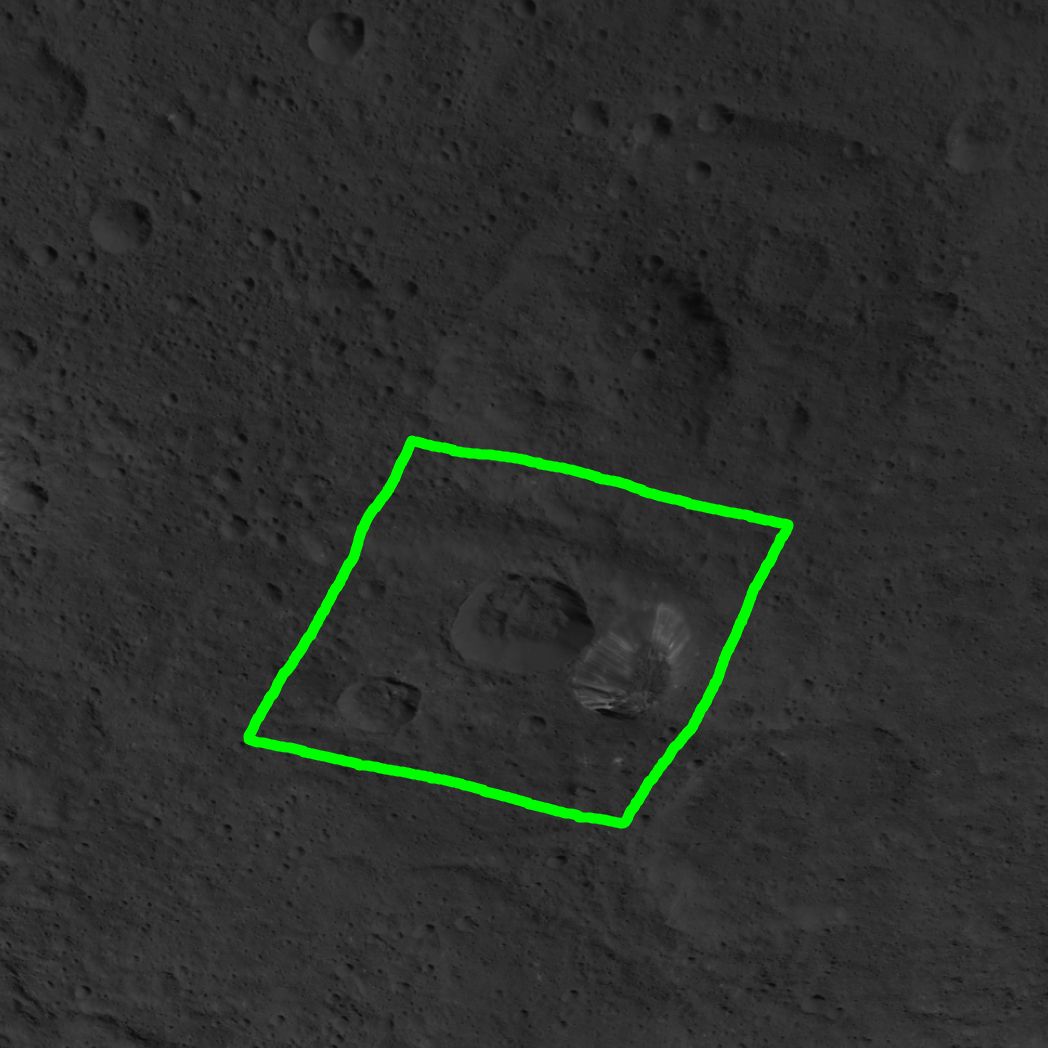} &
    \includegraphics[width=\linewidth]{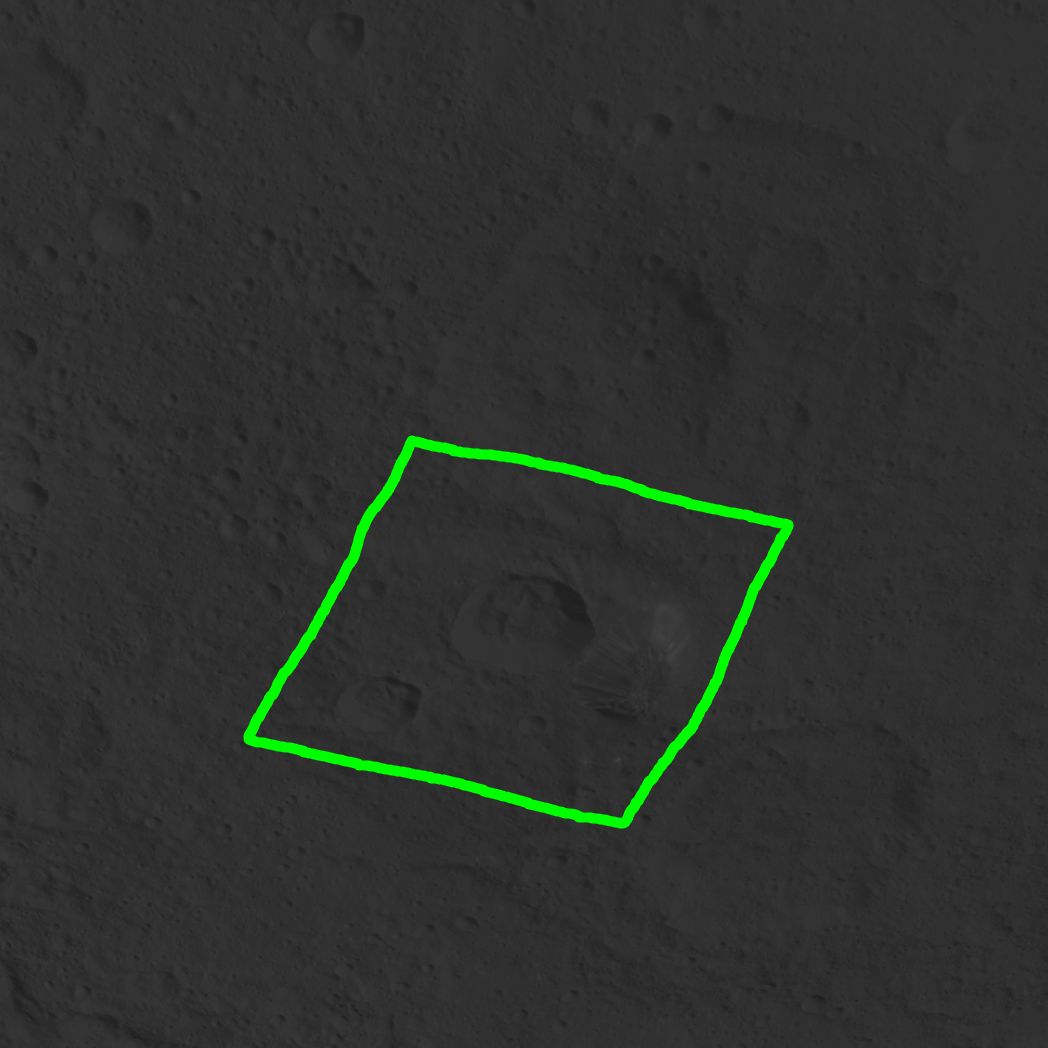} &
    \includegraphics[width=\linewidth]{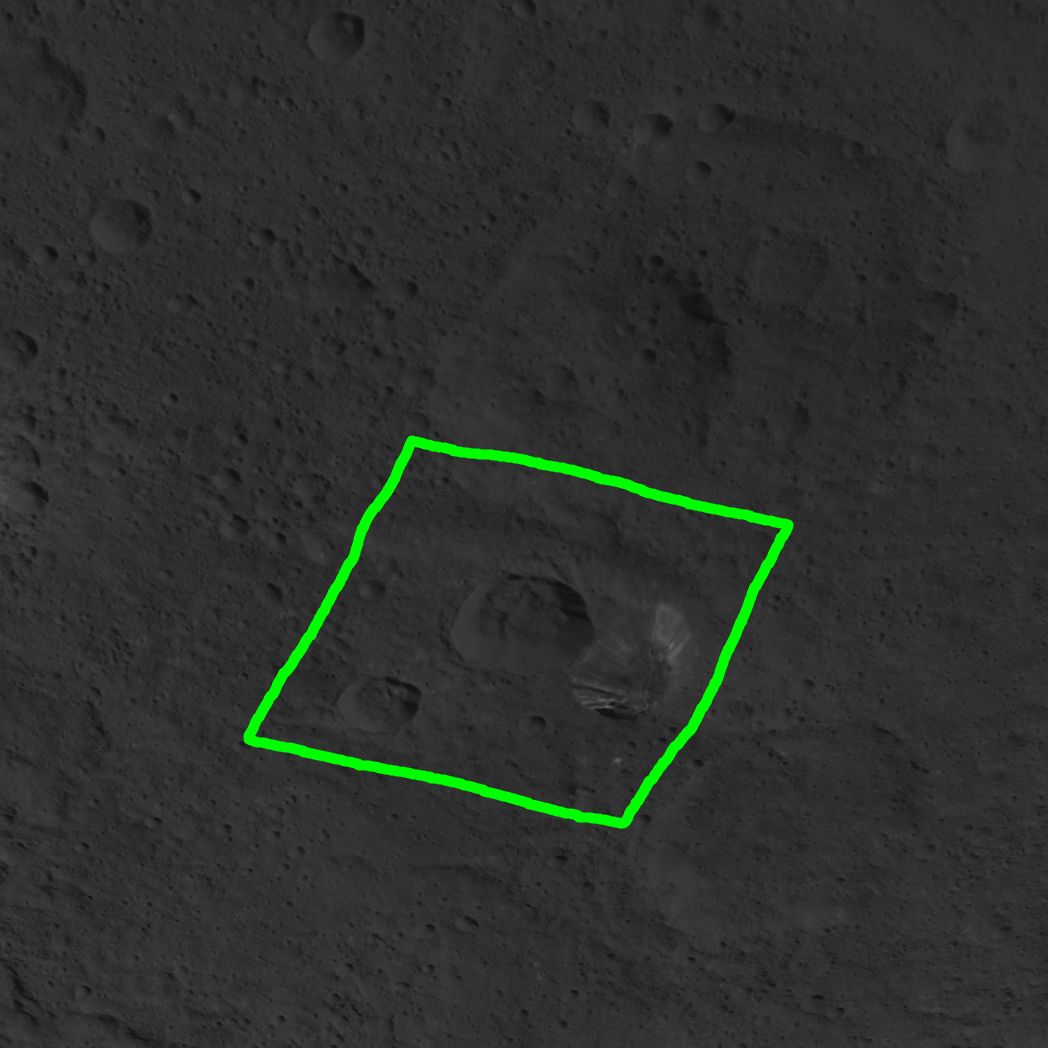} \\
    \parbox[t]{2.5mm}{\rotatebox[origin=l]{90}{\textcolor{gray}{\small{Ikapati}}}} &
    \includegraphics[width=\linewidth]{fig/render/ikapati/GT/00000.png} & %FC21B0008195_11275021648F1G.png
    \includegraphics[width=\linewidth]{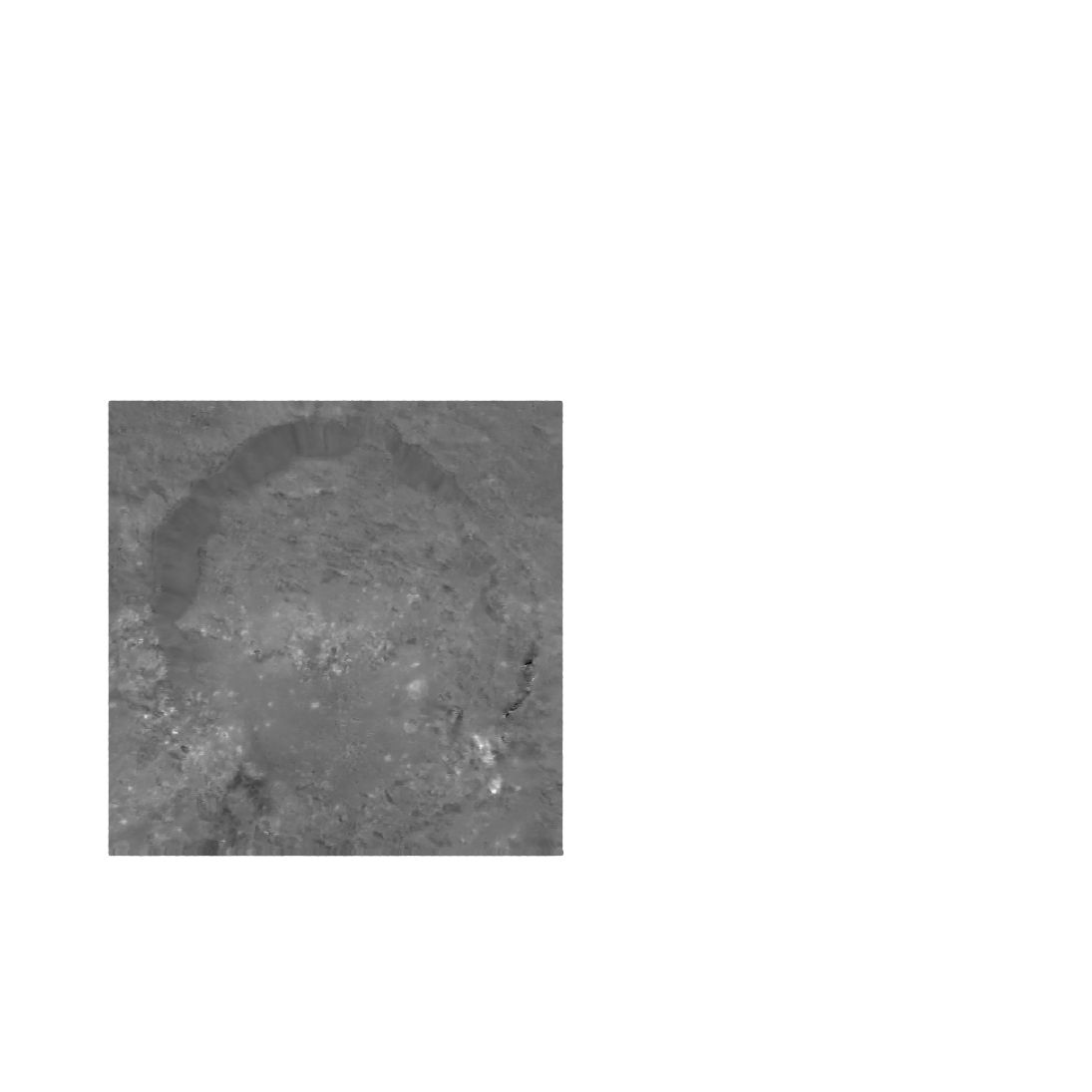} &
    \includegraphics[width=\linewidth]{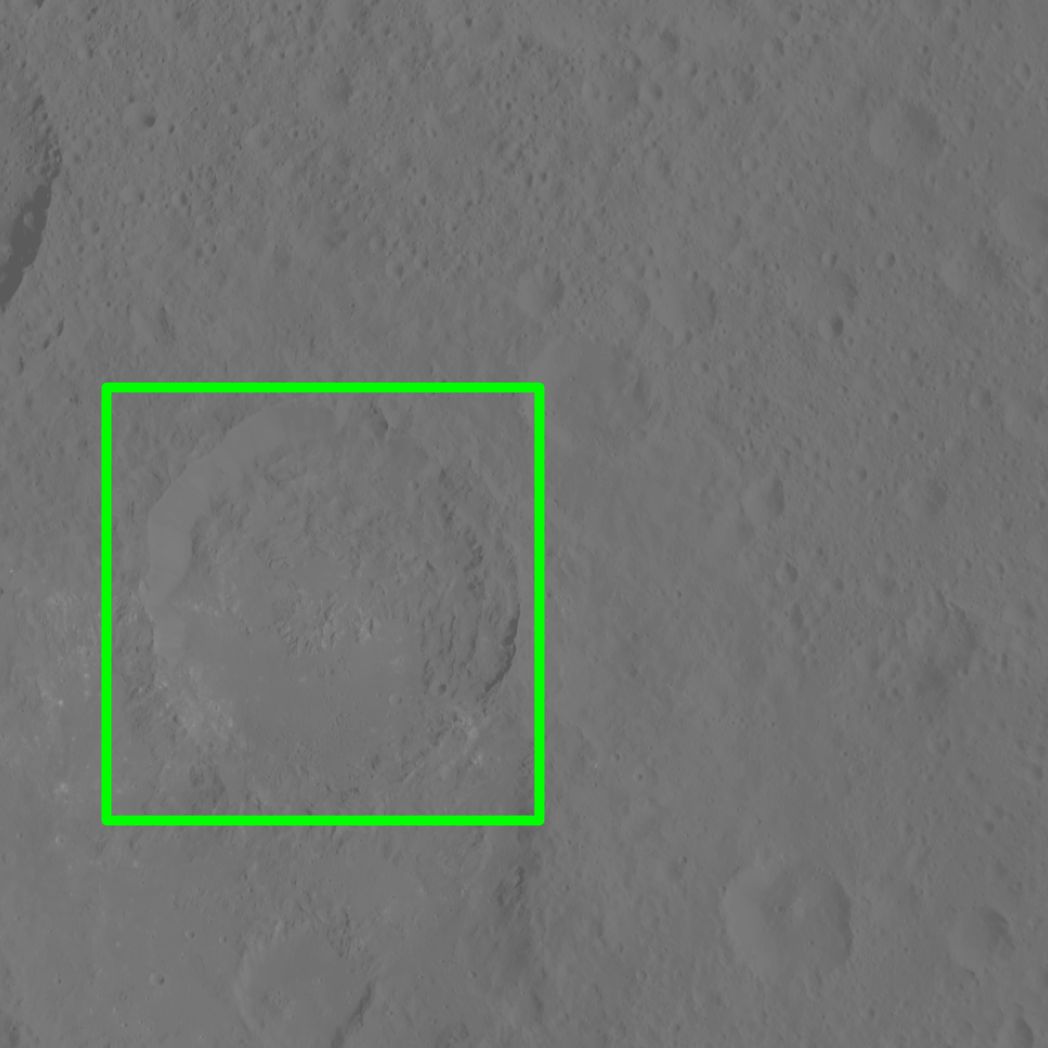} &
    \includegraphics[width=\linewidth]{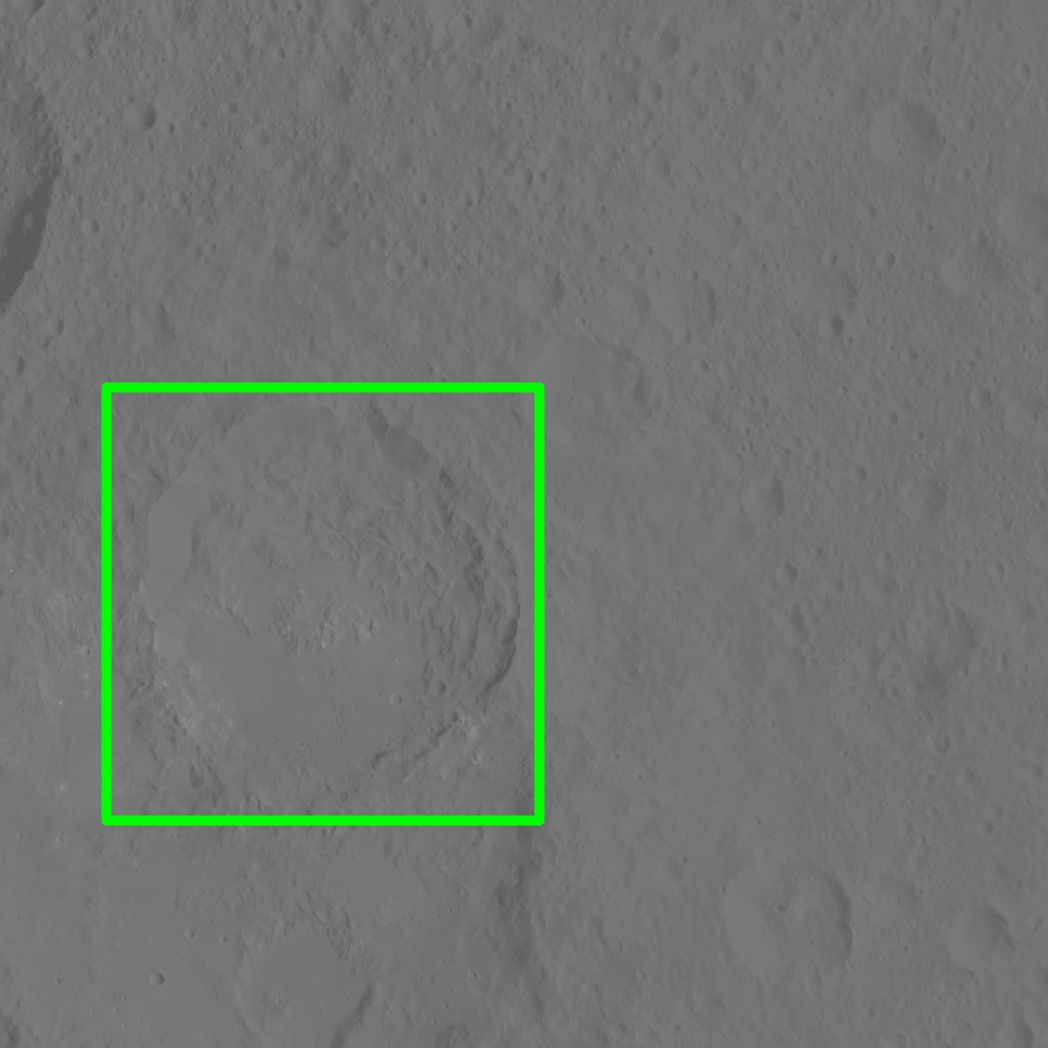} &
    \includegraphics[width=\linewidth]{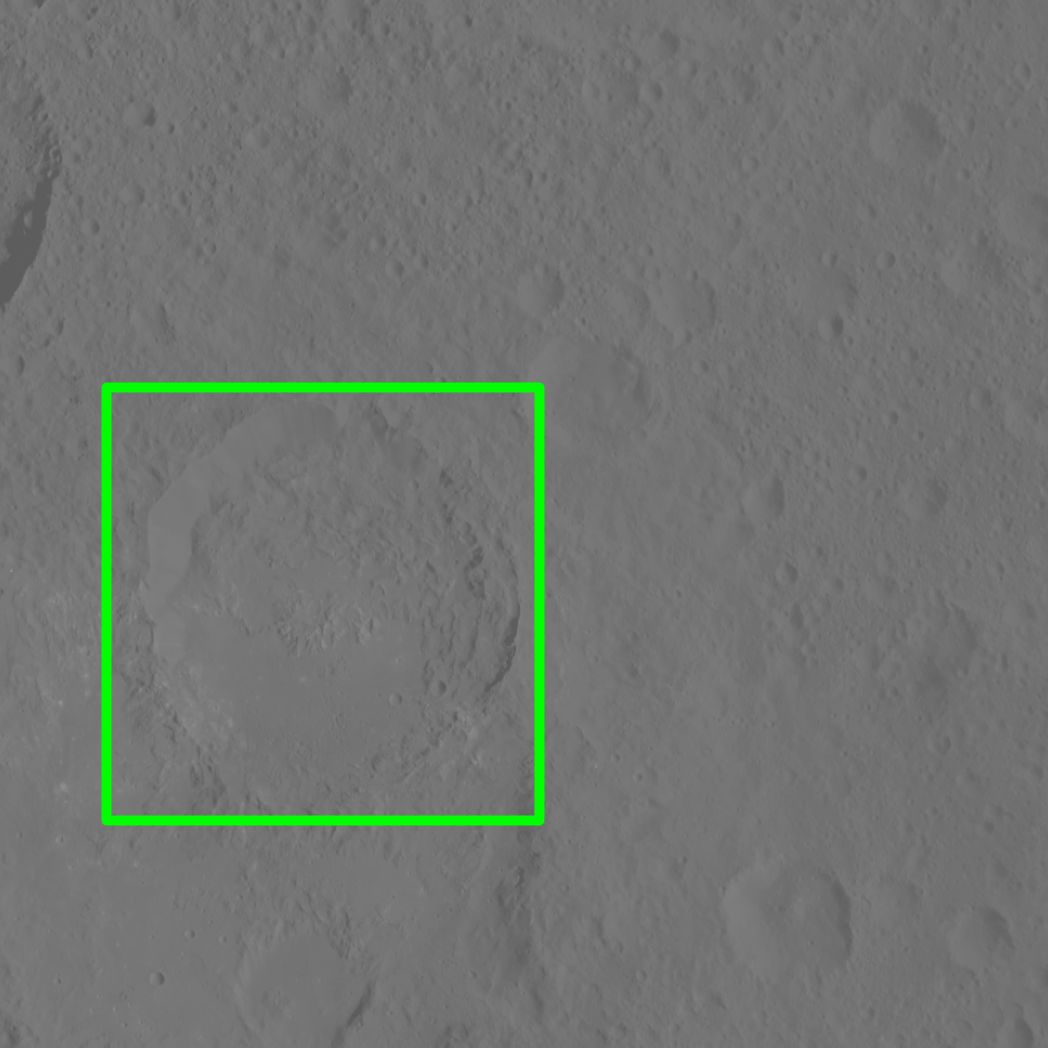} \\
\end{tabular}

%% file: sec/7_conclusion.tex
\section{Conclusion}

Deep learning strategies, like Gaussian splatting, have demonstrated promising results for surface reconstructions. However, in the space environment, physics-based methods have mission-proven success and yield reliable solutions, which make them desirable when mapping asteroids and minor planets. This paper introduces AstroSplat, a 2D Gaussian splatting framework that incorporates physics-based small body photometry into the intensity computation step to improve normal estimation of craters and other surface features. Instead of using the traditional Gaussian splatting technique of SH-based appearance-encoding, AstroSplat leverages available illumination information, viewing geometry, and surface normal estimates as well as a new per-Gaussian learned albedo to enable photometric modeling. This approach allows reflectance functions---like the Lambert, Lommel-Seeliger, and Lunar-Lambert models---to prescribe each Gaussian's intensity based on the behavior of light interacting with the topography of the surface, inherently linking the observed brightness in a render to the underlying normal estimates. This adjusted color computation step is validated on real imagery from NASA's Dawn mission and compared to a state-of-the-art surface reconstruction method. The results demonstrated both improved rendering quality and normal estimation accuracy when using physics-based approaches compared to the SH-based technique. Furthermore, the use of reflectance functions allowed for the introduction of albedo estimation, which is an important quantity for small body photometry. Future work may include exploring other reflectance models, such as empirical, body-specific methods, and introducing larger data sets to work toward global terrain mapping of a body.